\documentclass[11pt]{article}

\usepackage[final]{acl}

\usepackage{times}
\usepackage{latexsym}

\usepackage[T1]{fontenc}

\usepackage[utf8]{inputenc}

\usepackage{microtype}

\usepackage{inconsolata}

\usepackage{graphicx}

\newcommand{\ours}{\text{MA-Mesh}}
\usepackage{amsmath}
\usepackage{adjustbox} 
\usepackage{booktabs} 
\usepackage{multirow}
\usepackage{subcaption}
\usepackage{listings}
\usepackage{tcolorbox}
\usepackage{float}
\usepackage{algorithm}
\usepackage{algorithmic}
\usepackage{enumitem}
\usepackage{amssymb}
\usepackage{xcolor,colortbl}

\tcbuselibrary{listings, breakable,skins}
\newtcolorbox{promptbox}[1]{%
  enhanced jigsaw,
  breakable,
  width=\linewidth,
  colback=gray!10,
  colframe=black,
  colbacktitle=black,
  coltitle=white,
  title={\textbf{#1}},
  fonttitle=\bfseries,
  fontupper=\footnotesize,
  boxrule=0.5pt,
  arc=2pt,
  left=4pt,
  right=4pt,
  top=4pt,
  bottom=4pt,
  before skip=4pt,
  after skip=6pt
}

\tcbset{
  campPromptBox/.style={
    enhanced,
    colback=gray!10,
    colframe=black,
    fonttitle=\bfseries,
    boxrule=0.5pt,
    arc=1mm,
    left=4pt,
    right=4pt,
    top=4pt,
    bottom=4pt,
    listing only,
    listing options={
      basicstyle=\ttfamily\scriptsize,
      breaklines=true,
      breakatwhitespace=false,
      columns=fullflexible,
      keepspaces=true,
      showstringspaces=false,
      aboveskip=0pt,
      belowskip=0pt
    }
  }
}

\tcbset{
  campPlainListingBox/.style={
    enhanced,
    colback=gray!10,
    colframe=black,
    colbacktitle=black,
    coltitle=white,
    fonttitle=\bfseries,
    boxrule=0.5pt,
    arc=1mm,
    left=4pt,
    right=4pt,
    top=4pt,
    bottom=4pt,
    listing only,
    listing options={
      language={},
      basicstyle=\ttfamily\scriptsize,
      keywordstyle=\color{black},
      identifierstyle=\color{black},
      commentstyle=\color{black},
      stringstyle=\color{black},
      emphstyle=\color{black},
      frame=none,
      framerule=0pt,
      framesep=0pt,
      backgroundcolor=\color{gray!10},
      breaklines=true,
      breakatwhitespace=false,
      columns=fullflexible,
      keepspaces=true,
      showstringspaces=false,
      tabsize=2,
      aboveskip=0pt,
      belowskip=0pt,
      moredelim={[is][\bfseries]{@@}{@@}}
    }
  }
}
\DeclareMathOperator*{\argmax}{argmax}

\title{MA-Mesh: Resilient Multi-Agent Collaboration via Role-based Self-Healing}

\author{
JunYong Lee\thanks{Equal contribution.},
Woo Kyung Kim\footnotemark[1],
Honguk Woo\thanks{Corresponding author.} \\
{\normalfont Department of Computer Science and Engineering, Sungkyunkwan University} \\
{\normalfont \texttt{\{ljy7488,kwk2696,hwoo\}@skku.edu}}
}

\begin{document}
\maketitle
\begin{abstract}
Effective collaboration in multi-agent systems requires mechanisms for coordination and adaptability in dynamic and uncertain environments. 
However, existing multi-agent systems rarely model resilience explicitly, instead relying on ad-hoc output correction that limits systematic, runtime management of robustness.
In this work, we propose Multi-Agent Mesh (MA-Mesh), a framework that constructs a role-based self-healing layer on top of a collaboration pattern.
Specifically, MA-Mesh generates resilience-aware system specifications and synthesizes them into executable multi-agent policy code.
To enhance resilience, we introduce two self-healing mechanisms: Cross-agent reflection, a mechanism that allows agents to evaluate not only their own outcomes but also the performance of their peers, enabling informed behavioral adjustments and collective policy refinement; Role-based failover, a strategy which dynamically reassigns roles and enables agents to substitute for one another when sub-tasks fail or agents underperform, ensuring continuity of task execution. 
Experimental results across diverse tasks (e.g., software development, tool use, scientific reasoning) demonstrate that MA-Mesh significantly outperforms existing multi-agent systems in both task performance and robustness under system uncertainty. 
\end{abstract}

\section{Introduction}
Building on the success of single LLMs for individual tasks~\cite{coding:codellama,mm_reasoning:multiCoT,planning:saycan}, compound agentic systems composed of multiple LLM-based agents have emerged as a promising approach to scale these capabilities toward more complex, long-horizon tasks that demand collaboration and coordination~\cite{mas:mas_autogen}.
%
%
Early multi-agent systems were developed in a domain-specialized manner~\cite{mas:mr:llm_debate,mas:sw:metagpt}, with each system hand-crafted for a specific application. 
In parallel, a line of work has explored automating multi-agent orchestration, either by leveraging the parametric knowledge of LLMs ~\cite{mas:auto:agentverse,mas:auto:autoagents} or by learning to configure the coordination structures~\cite{mas:auto:maas}. 
However, these multi-agent systems are typically designed under nominal system conditions, which limits their \emph{resilience} to partial system failures and internal execution uncertainty.

To address this, we draw inspiration from the cloud service mesh~\cite{cloud:servicemesh} to automate the construction of resilient multi-agent systems that can be instantiated dynamically on top of a collaboration pattern.
Specifically, we explicitly decouple the collaboration logic, which governs task execution, from the resilience policy, which materializes a role-based self-healing layer.
%
%
The self-healing layer operates through two mechanisms: (\romannumeral 1) cross-agent reflection, which evaluates an agent's own outcomes or those of its peers against its role-specific requirements to detect performance degradation; (\romannumeral 2) role-based failover, which continuously monitors agent execution and substitutes an affected agent with a suitable alternative when reflection signals performance degradation or a response timeout occurs.
Such a design directly parallels a service mesh, where sidecar proxies transparently provide a resilience layer to microservices. 


To this end, we present the Multi-Agent Mesh ($\ours$) framework, which overlays a role-based self-healing layer on top of a collaboration pattern, yielding a resilient multi-agent system.
%
Given a task, an LLM orchestrator generates a resilience-aware system specification in a structured format. The specification comprises collaboration logic, which is either a user-supplied or an orchestrator-generated workflow, and a cross-agent reflection mechanism, which specifies the peers to evaluate and the role-specific criteria for assessing their execution outcomes.
The orchestrator then synthesizes the executable code by instantiating pre-defined agent and resilient proxy classes from this specification. A resilient proxy is attached to each agent; collectively, these proxies realize the self-healing layer.
%
%
The role-based failover mechanism is built into the resilient proxy class, which monitors agent execution and triggers recovery upon performance degradation (detected by cross-agent reflection) or response timeout. Once triggered, a dedicated meta-agent replaces the affected agent and reassigns its role, ensuring execution continuity without global re-orchestration.

In this way, the resulting resilient multi-agent system is constructed analogously to a service mesh architecture, comprising a control plane and a data plane.
%
%
The resilience-aware system specification, together with 
the meta-agent responsible for runtime recovery upon agent failures, constitutes the control plane; the agent instances and their attached sidecar-like resilient proxies form the data plane. 
%

Through evaluations on software development for GUI and web applications~\cite{mas:sw:rtadev, benchmark:rsde}, tool-use tasks~\cite{benchmark:gaia}, and scientific reasoning~\cite{benchmark:gpqa}, we demonstrate that $\ours$ substantially improves the resilience of multi-agent systems.
%
%
We evaluate under four system uncertainty scenarios~\cite{chaos:monkey} that model partial and behavioral uncertainty: agent chaos, prompt chaos, I/O chaos, and compound chaos.
Across these settings, $\ours$ consistently outperforms the strongest baseline C+I~\cite{mas:res:cni}, with an average accuracy gain of 32.51 percentage points (pp).
We further evaluate detection recall (DR), the fraction of affected agents correctly identified by resilient proxies, and observe an average improvement of $0.36$ in DR over C+I.
Moreover, to validate $\ours$'s adaptability to arbitrary collaboration patterns, we overlay it on top of MetaGPT~\cite{mas:sw:metagpt} and RTADev~\cite{mas:sw:rtadev}, 
observing an average accuracy gain of $43.74$ pp and $29.42$ pp, respectively (see Table~\ref{tab:analysis:plug_in_resilience}).
%
%
Lastly, we validate $\ours$ on AIOpsLab~\cite{cloud:aiopslab}, a cloud-native testbed, demonstrating its practical applicability beyond synthetic chaos scenarios.
The contributions of our work are summarized as follows:
\begin{itemize}[leftmargin=*]
    \item We present the $\ours$ framework to construct multi-agent systems resilient to partial system failures and behavioral uncertainty.
    \item We explicitly decouple collaboration logic from resilience policy in a structured specification, from which we synthesize executable code, enabling a role-based self-healing layer to be overlaid on top of any collaboration pattern.
    \item We introduce resilient proxies to realize the self-healing layer, comprising cross-agent reflection for role-specific degradation detection and role-based failover for recovery.
    \item Through experiments on benchmarks including software development, tool use, and scientific reasoning, we demonstrate that $\ours$ outperforms existing multi-agent systems under diverse system uncertainty scenarios. 
\end{itemize}
\begin{figure*}[t]
  \begin{center}
    \includegraphics[width=0.96\textwidth]{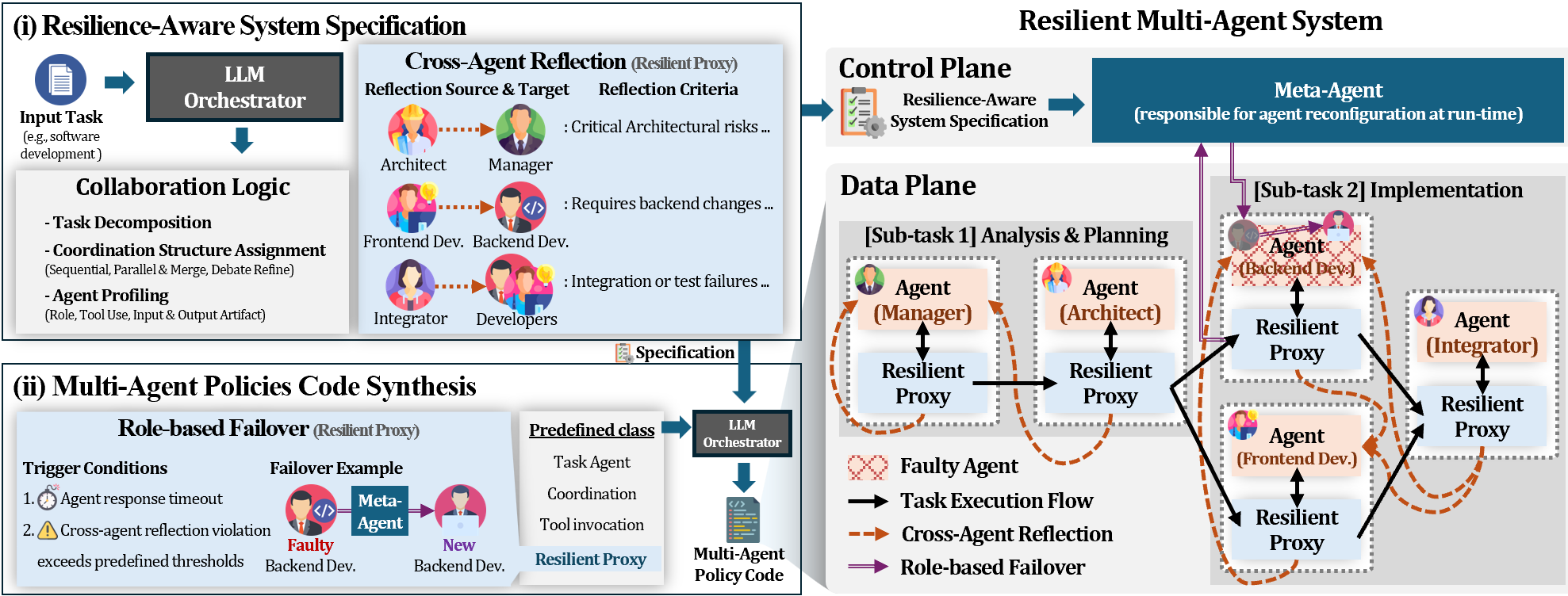}
    \caption{$\ours$ Framework: In (\romannumeral 1), given a task, the LLM orchestrator generates a resilience-aware system specification comprising the collaboration logic and the cross-agent reflection, which specifies the reflection targets and criteria. 
    In (\romannumeral 2), the orchestrator synthesizes executable multi-agent policy code that incorporates role-based failover, including its trigger conditions and meta-agent-driven agent replacement.
    As a result, the resilience-aware system specification and the meta-agent responsible for failover recovery form the control plane, while the task-executing agents and their resilient proxies form the data plane, which enforces the resilient mechanisms at runtime.}
    \label{fig:framework}
  \end{center}
\end{figure*}

\section{Related Work} \label{sec:related_works}
Multi-agent systems have been widely adopted across diverse domains, including software development~\cite{mas:sw:metagpt,mas:sw:rtadev}, scientific discovery~\cite{mas:sd:mtmol,mas:sd:agentlab}, and complex mathematical reasoning~\cite{mas:mr:llm_debate,mas:mr:macm}.
In parallel, several works~\cite{mas:auto:agentverse, mas:auto:autoagents} automate the configuration of multi-agent systems, dynamically adapting agent roles and coordination structures via an LLM orchestrator.
Yet, these approaches assume nominal operating conditions and rarely address system-level resilience.
%
%
Recently, NetSafe~\cite{mas:res:netsafe} analyzes which coordination structures are inherently robust under adversarial attacks, while C+I~\cite{mas:res:cni} explicitly handles errors through challenger and inspector agents. However, C+I remains limited to correction-based mitigation, lacking role-aware evaluation and reconfiguration.
%
In contrast, $\ours$ provides a service-mesh-style framework that augments any collaboration pattern with two mechanisms: cross-agent reflection, whose evaluation scope and criteria are dynamically determined, and role-based failover, which reconfigures agents at runtime upon failures or performance degradation.

\section{Problem Formulation}
A multi-agent system $\Pi$ executes a task $\mathcal{T}$ through collaboration among multiple LLM-based agents, each assigned a distinct role. 
Formally, $\Pi$ is represented as a directed graph, i.e.,
\begin{equation}\label{equation:mas}
    \Pi = ( \mathcal{V},\mathcal{E} ), \mathcal{E} \subseteq \mathcal{V} \times \mathcal{V},
\end{equation}
where $\mathcal{V}$ is the set of agents, and $\mathcal{E}$ is the set of directed communication edges.

In real-world deployments, multi-agent systems are exposed to diverse sources of system uncertainty, such as network or hardware failures, imperfect agent behavior, and adversarial interventions. To model such uncertainty, we adopt chaos engineering practices~\cite{chaos:monkey, chaos:mesh} which inject stochastic perturbations during system execution.
Concretely, we consider three primitive classes of perturbations: \emph{agent chaos}, which captures agent termination due to network or hardware faults; \emph{prompt chaos}, which models errors in the agent prompts; and \emph{I/O chaos}, which captures partial corruption of agent-generated outputs during transmission.

Then, the resilient agent collaboration problem is formalized through a framework $\Phi(\mathcal{T})$ that maps a task $\mathcal{T}$ to a multi-agent system $\Pi$, maximizing the task performance \textbf{P}$(\Pi, \mathcal{T}, u)\in [0,1]$ under system uncertainty $u \in \mathcal{U}$ as follows:
\begin{equation}
    \Phi^*= \argmax_{\Phi} \mathbb{E}_{u \sim \mathcal{U}, \Pi \sim \Phi(\mathcal{T})} 
    \left[ \textbf{P}(\Pi, \mathcal{T}, u) \right].
\end{equation}
%


\section{Our Approach}
\subsection{Overall Framework}
To automate the construction of resilient multi-agent systems, we present the $\ours$ framework comprising two stages: 
(\romannumeral 1) \textbf{resilience-aware system specification}, and (\romannumeral 2) \textbf{multi-agent policy code synthesis}, as illustrated on the left side of Figure~\ref{fig:framework}.
In (\romannumeral 1), an LLM orchestrator derives a structured specification that consists of the collaboration logic and the cross-agent reflection mechanism. The collaboration logic specifies the sub-task decomposition, the coordination structure, and the agent roles. As a default, it is dynamically generated by the orchestrator, but it can also be adopted from a predefined collaboration pattern of an existing static multi-agent system.
Cross-agent reflection associates each agent with a resilient proxy that specifies the reflection target and criteria. At runtime, these proxies assess the outcomes produced by peer agents and provide feedback accordingly.
%
In (\romannumeral 2), given the resulting specification, the orchestrator synthesizes executable multi-agent policy code by instantiating the agents and resilient proxies as parameterized classes. In doing so, the role-based failover mechanism is embedded into the resilient proxies, which trigger failover upon detecting runtime execution failures or suboptimal agent behavior, prompting the meta-agent to replace the affected agent.

Consequently, $\ours$ constructs a service-mesh-like resilient multi-agent system, where the resilience-aware system specification and the meta-agent responsible for runtime agent replacement together form the \textbf{control plane}, while the agent instances and their resilient proxies constitute the \textbf{data plane}, as shown on the right side of Figure~\ref{fig:framework}.

\subsection{Resilience-Aware System Specification}\label{sec:4:2}

Given a task $\mathcal{T}$, the LLM orchestrator $\phi_\text{LLM}$ derives a structured specification of the resilient multi-agent system in a form readily translatable into executable code.
%
As shown in Figure~\ref{fig:spec}, the specification comprises two parts: the collaboration logic and the cross-agent reflection. The collaboration logic decomposes the task into $K$ sub-tasks $\tau$, each paired with a coordination structure $c$ and a set of agent profiles $\mathcal{P}$.
The cross-agent reflection $\mathcal{R}$ is defined over all agents in the system, assigning each agent a resilient proxy that specifies its reflection target and criteria. Together, they yield the resilience-aware system specification as,
%
%
\begin{equation}
    \phi_{\text{LLM}}: \mathcal{T} \rightarrow \Big(\{(\tau_k, c_k, \mathcal{P}_k)\}_{k=1}^{K}, \mathcal{R} \Big).
\end{equation}
Note that the collaboration logic, generated by the orchestrator by default, can instead be drawn from an existing static multi-agent system (see Table~\ref{tab:analysis:plug_in_resilience}).

\begin{figure}[H]
\centering
\begin{tcolorbox}[
  colback=white,
  colframe=gray!60,
  boxrule=0.5pt,
  arc=2pt,
  left=4pt,
  right=4pt,
  top=4pt,
  bottom=1pt
]
\begin{lstlisting}[
  backgroundcolor=\color{yellow!10},
  basicstyle=\ttfamily\scriptsize,
  mathescape=true,
  aboveskip=0pt,
  belowskip=0pt
]
# Collaboration logic: $\{(\tau_k, c_k, \mathcal{P}_k)\}_{k=1}^{K}$
sub_task: str = "Develop data management ..."
coordination_structure: str = "Parallel & Merge"
agent_profiles: list = [
 { "agent_name": "DataManagerDeveloper",
   "responsibility": "Implement manage.py ...",
   "tools": ["file write"] }, ...]
# Repeated for $k=1,\ldots,K$
\end{lstlisting}
\begin{lstlisting}[
  backgroundcolor=\color{red!10},
  basicstyle=\ttfamily\scriptsize,
  mathescape=true,
  aboveskip=1pt,
  belowskip=0pt
]
# Resilient Proxies: Cross-Agent Reflection $\mathcal{R}$
[{ "reflection_source": "UIArchitect",
   "reflection_targets": "DataManagerDeveloper",
   "reflection_criteria": (
    "1. Check manage.py implements all methods",
    "2. Check manage.py handles file I/O.",
     ...)}, ...]
\end{lstlisting}
\end{tcolorbox}
\caption{Resilience-aware system specification}
\label{fig:spec}
\end{figure}

\noindent\textbf{Coordination structure.} For each sub-task, the LLM orchestrator selects the coordination structure that best fits its objective from a pool $\mathcal{C} =$ \{`\emph{sequential}', `\emph{parallel\&merge}', `\emph{debate}', `\emph{refine}'\}, adopted from the taxonomy of AutoGen~\cite{mas:mas_autogen}. Detailed examples are shown in  Figure~\ref{fig:coordination_structure}.
%

\noindent\textbf{Agent profile.} Each agent profile specifies the agent's role-related configuration, including responsibilities, permitted tools, input sources and modalities, and expected output artifacts and formats. The corresponding agent prompts are generated through separate LLM invocations.

\noindent\textbf{Resilient proxy: cross-agent reflection.} To enhance system resilience, we equip each agent with a resilient proxy.
Each proxy encodes a cross-agent reflection mechanism that specifies (1) the reflection targets, which are the agents whose outputs it evaluates, and (2) reflection criteria, which capture both output quality and role-level expectations from the reflecting agent's perspective.
%
Note that a reflection target may belong to a different sub-task, and if an agent is selected by no other agent, its resilient proxy performs self-reflection by default.
Accordingly, the resulting resilient multi-agent system defined in Eq.~\eqref{equation:mas} is extended as
\begin{equation}
\begin{gathered}
    \Pi = ( \mathcal{V}, \mathcal{V}_r,  \mathcal{E} ), 
    \\
    \mathcal{E} \subseteq (\mathcal{V}_r \times \mathcal{V}_r) \cup (\mathcal{V}_r \times \mathcal{V}) \cup (\mathcal{V} \times \mathcal{V}_r),
\end{gathered}
\end{equation}

where $\mathcal{V}_r$ denotes the set of resilient proxies. 
This design removes direct communication between agents $\mathcal{V}$ and routes all interactions through resilient proxies $\mathcal{V}_r$, ensuring that every outcome is intercepted for cross-agent reflection.
%
When an outcome violates the reflection criteria, the proxy returns feedback prompting the agent to regenerate its output.
%
We maintain a retry counter to track the number of reflection attempts; once it exceeds a predefined threshold, the role-based failover (see Section~\ref{sec:4.3}) is triggered to replace the agent.
%
%

\subsection{Multi-Agent Policy Code Synthesis} \label{sec:4.3}
Given the resilience-aware system specification, the LLM orchestrator translates it into executable code.
We build on the AutoGen~\cite{mas:mas_autogen} library, pre-wrapping each task-executing agent and its associated resilient proxy into a class so that agent profiles and resilient proxy specifications can be instantiated through parameterized constructors.
%
%
\begin{figure}[H]
\centering
\begin{tcolorbox}[
  colback=white,
  colframe=gray!60,
  boxrule=0.5pt,
  arc=2pt,
  left=4pt,
  right=4pt,
  top=4pt,
  bottom=1pt
]
\begin{lstlisting}[
  backgroundcolor=\color{gray!10},
  basicstyle=\ttfamily\scriptsize,
  mathescape=true,
  escapechar=\%
]
class AgentWithResilientProxy:
  def __init__(self, agent_profile, proxy_spec):
    # task executing agent
    self.agent = create_agent(**agent_profile)
    # resilient proxy
    self.proxy = create_proxy(**proxy_spec)
  def __call__(self, request):
    # see Algorithm%~\ref{algo:overall_algorithm}%
  def cross_agent_reflection(self):
    # see Algorithm%~\ref{algo:cross_agent_reflection}% in Appendix%~\ref{appendix:c}%
    # :return: feedback if error identified
  def role_based_failover(self):
    # see Algorithm%~\ref{algo:role_based_failover}% in Appendix%~\ref{appendix:c}%
    # :return: a new agent for replacement
\end{lstlisting}
\end{tcolorbox}
\caption{Example code  for agent with resilient proxy}
\label{fig:code}
\end{figure}

\noindent\textbf{Resilient proxy: role-based failover.}
The role-based failover mechanism is built into the resilient proxy class, so code synthesis simply instantiates proxies already equipped to detect failover conditions and initiate agent replacement.
%
The replacement is handled by a dedicated meta-agent that does not itself participate in task execution. To replace an agent identified as faulty or degraded, the meta-agent initializes a new agent with a refined role specification. 
%
Role-based failover is triggered under two conditions.
(\romannumeral 1) The resilient proxy enforces a response timeout to regulate each agent's execution. If an agent fails to respond within this limit, the proxy triggers failover.
(\romannumeral 2) Failover is also triggered when the cross-agent reflection repeatedly identifies that an agent fails to satisfy the specified reflection criteria, with the number of failed attempts exceeding a predefined retry threshold.
As a result, role-based failover enables the system to dynamically recover from both runtime execution failures and suboptimal agent behaviors, ensuring resilient task execution without re-orchestrating the entire multi-agent system.

The overall procedure of an agent and its resilient proxy is summarized in Algorithm~\ref{algo:overall_algorithm}.

\begin{algorithm}[t]
    \begin{algorithmic}[1]
    \STATE{\textbf{Input:} Agent $A$, Proxy $P$, Request $x$}
    \STATE{\textbf{Output:} Response $y$}
    \FOR{$r = 0$ to $\mathrm{max\_retry}$}
        \STATE \textbf{try:}
        \STATE \hspace{.8em} $y \leftarrow \mathrm{agent\_call}(A, x)$ 
        \STATE \textbf{except Timeout:}
        \STATE \hspace{.8em} {\small \textcolor{blue}{/* trigger failover due to response timeout */}}
        \STATE \hspace{.8em} $A \leftarrow \mathrm{role\_based\_failover}(P)$ 
        \STATE \hspace{.8em} \textbf{continue}
        
        \STATE $n \leftarrow 0$
        \WHILE{$n < \mathrm{max\_reflection\_attempts}$}
            \STATE {\small \textcolor{blue}{/* generate feedback via cross-agent reflection */}}
            \STATE $f \leftarrow \mathrm{cross\_agent\_reflection}(P, x, y)$
            \IF {$f$ is None}
                \STATE \textbf{return} $y$
            \ELSE
                \STATE $x \leftarrow \mathrm{concat}(x, y, f); n \leftarrow n + 1$
                \STATE {\small \textcolor{blue}{/* timeout handling omitted for brevity */}}
                \STATE $y \leftarrow \mathrm{agent\_call}(A, x)$ 
            \ENDIF
        \ENDWHILE
        \STATE {\small \textcolor{blue}{/* trigger failover due to violation of reflection */}}
        \STATE $A \leftarrow \mathrm{role\_based\_failover}(P)$ 
    \ENDFOR
    \STATE \textbf{return} $y$
    \end{algorithmic}
    \caption{Procedure of Resilient Proxy}
    \label{algo:overall_algorithm}
\end{algorithm}
%

%
\subsection{Resilient Multi-Agent System}
Conceptually, the resulting resilient multi-agent system follows a service-mesh-like architecture, comprising a control plane and a data plane.
The resilience-aware system specification serves as the control plane artifact, encoding collaboration logic and cross-agent reflection. At runtime, the meta-agent residing in the control plane leverages this specification to instantiate appropriate replacement agents, preserving the continuity of task execution.
%
The data plane consists of the agents and resilient proxies instantiated from the synthesized multi-agent code. The agents are responsible for task execution, while resilient proxies enforce the encoded role-based self-healing mechanisms by monitoring agent execution, applying timeout constraints, performing cross-agent reflection, and triggering role-based failover when necessary.
%

\begin{table*}[t]
\centering
\small
\begin{adjustbox}{width=\textwidth}
\begin{tabular}{ll c cc cc cc cc}
\toprule
 & &
\multicolumn{1}{c}{\textbf{Vanilla}} &
\multicolumn{2}{c}{\textbf{Agent Chaos}} &
\multicolumn{2}{c}{\textbf{Prompt Chaos}} &
\multicolumn{2}{c}{\textbf{I/O Chaos}} &
\multicolumn{2}{c}{\textbf{Compound Chaos}} \\
\cmidrule(lr){3-3} \cmidrule(lr){4-5} \cmidrule(lr){6-7} \cmidrule(lr){8-9} \cmidrule(lr){10-11}
\textbf{Benchmark} & \textbf{Method} &
\textbf{Acc (\%)} &
\textbf{Acc (\%)} & \textbf{DR} &
\textbf{Acc (\%)} & \textbf{DR} &
\textbf{Acc (\%)} & \textbf{DR} &
\textbf{Acc (\%)} & \textbf{DR} \\
\midrule

\multirow{7}{*}{\textbf{\begin{tabular}{c} GUI \\ Dev. \end{tabular}}} &
CoT
& 66.02{\color{gray}\scriptsize$\pm$2.53} 
& 0.00{\color{gray}\scriptsize$\pm$0.00} & 0.00{\color{gray}\scriptsize$\pm$0.00} 
& 25.03{\color{gray}\scriptsize$\pm$5.87} & 0.00{\color{gray}\scriptsize$\pm$0.00} 
& 0.00{\color{gray}\scriptsize$\pm$0.00} & 0.00{\color{gray}\scriptsize$\pm$0.00} 
& 3.96{\color{gray}\scriptsize$\pm$4.40} & 0.00{\color{gray}\scriptsize$\pm$0.00} \\

& RTADev    
& 66.30{\color{gray}\scriptsize$\pm$1.60} 
& 21.87{\color{gray}\scriptsize$\pm$8.35} & 0.61{\color{gray}\scriptsize$\pm$0.15} 
& 39.64{\color{gray}\scriptsize$\pm$3.27} & 0.42{\color{gray}\scriptsize$\pm$0.14} 
& 33.90{\color{gray}\scriptsize$\pm$13.38} & 0.19{\color{gray}\scriptsize$\pm$0.17} 
& 14.99{\color{gray}\scriptsize$\pm$8.29} & 0.43{\color{gray}\scriptsize$\pm$0.09} \\

& MetaGPT   
& 70.51{\color{gray}\scriptsize$\pm$4.22} 
& 7.39{\color{gray}\scriptsize$\pm$6.10} 
& 0.00{\color{gray}\scriptsize$\pm$0.00} 
& 21.62{\color{gray}\scriptsize$\pm$5.12} 
& 0.00{\color{gray}\scriptsize$\pm$0.00} 
& 22.63{\color{gray}\scriptsize$\pm$11.46} 
& 0.00{\color{gray}\scriptsize$\pm$0.00} 
& 0.00{\color{gray}\scriptsize$\pm$0.00} 
& 0.00{\color{gray}\scriptsize$\pm$0.00}\\

& C+I       
& 68.05{\color{gray}\scriptsize$\pm$2.46}
& 4.46{\color{gray}\scriptsize$\pm$4.47} & 0.24{\color{gray}\scriptsize$\pm$0.03} 
& 19.30{\color{gray}\scriptsize$\pm$6.23} & 0.29{\color{gray}\scriptsize$\pm$0.04}
& 47.37{\color{gray}\scriptsize$\pm$4.24} & \textbf{0.86{\color{gray}\scriptsize$\pm$0.03}}
& 1.59{\color{gray}\scriptsize$\pm$2.75}  & 0.30{\color{gray}\scriptsize$\pm$0.03} \\

& AgentVerse
& 66.53{\color{gray}\scriptsize$\pm$6.59} 
& 32.87{\color{gray}\scriptsize$\pm$6.03} 
& 0.00{\color{gray}\scriptsize$\pm$0.00} 
& 44.61{\color{gray}\scriptsize$\pm$8.08} 
& 0.00{\color{gray}\scriptsize$\pm$0.00} 
& 37.89{\color{gray}\scriptsize$\pm$12.94} 
& 0.00{\color{gray}\scriptsize$\pm$0.00} 
& 26.36{\color{gray}\scriptsize$\pm$6.13} 
& 0.00{\color{gray}\scriptsize$\pm$0.00}\\

& EvoMAC
& 72.79{\color{gray}\scriptsize$\pm$1.96} 
& 48.01{\color{gray}\scriptsize$\pm$11.28} & 0.00{\color{gray}\scriptsize$\pm$0.00}
& 52.15{\color{gray}\scriptsize$\pm$8.34} & 0.00{\color{gray}\scriptsize$\pm$0.00}
& 63.65{\color{gray}\scriptsize$\pm$7.45} & 0.00{\color{gray}\scriptsize$\pm$0.00}
& 24.38{\color{gray}\scriptsize$\pm$11.92} & 0.00{\color{gray}\scriptsize$\pm$0.00} \\

& \cellcolor{gray!10}\textbf{$\ours$} 
& \cellcolor{gray!10}\textbf{75.29{\color{gray}\scriptsize$\pm$4.34}} 
& \cellcolor{gray!10}\textbf{71.05{\color{gray}\scriptsize$\pm$1.64}}  
& \cellcolor{gray!10}\textbf{1.00{\color{gray}\scriptsize$\pm$0.00}} 
& \cellcolor{gray!10}\textbf{65.30{\color{gray}\scriptsize$\pm$5.48}}  
& \cellcolor{gray!10}\textbf{0.84{\color{gray}\scriptsize$\pm$0.08}} 
& \cellcolor{gray!10}\textbf{67.87{\color{gray}\scriptsize$\pm$5.07}}  
& \cellcolor{gray!10}0.82{\color{gray}\scriptsize$\pm$0.07} 
& \cellcolor{gray!10}\textbf{65.66{\color{gray}\scriptsize$\pm$5.69}}  
& \cellcolor{gray!10}\textbf{0.90{\color{gray}\scriptsize$\pm$0.05}} \\
\midrule

\multirow{7}{*}{\textbf{\begin{tabular}{c} Web \\ Dev. \end{tabular}}} &
CoT
& 45.73{\color{gray}\scriptsize$\pm$3.02} 
& 0.00{\color{gray}\scriptsize$\pm$0.00} & 0.00{\color{gray}\scriptsize$\pm$0.00} 
& 0.60{\color{gray}\scriptsize$\pm$1.04} & 0.00{\color{gray}\scriptsize$\pm$0.00} 
& 0.00{\color{gray}\scriptsize$\pm$0.00} & 0.00{\color{gray}\scriptsize$\pm$0.00} 
& 0.19{\color{gray}\scriptsize$\pm$0.32} & 0.00{\color{gray}\scriptsize$\pm$0.00} \\

& RTADev
& 40.17{\color{gray}\scriptsize$\pm$6.78}
& 12.68{\color{gray}\scriptsize$\pm$4.60} & 0.44{\color{gray}\scriptsize$\pm$0.17}
&  8.85{\color{gray}\scriptsize$\pm$0.53} & 0.46{\color{gray}\scriptsize$\pm$0.02}
& 16.74{\color{gray}\scriptsize$\pm$4.50} & 0.20{\color{gray}\scriptsize$\pm$0.11}
&  2.68{\color{gray}\scriptsize$\pm$4.12} & 0.36{\color{gray}\scriptsize$\pm$0.06} \\

& MetaGPT
& 42.09{\color{gray}\scriptsize$\pm$7.03}
& 4.81{\color{gray}\scriptsize$\pm$2.77} & 0.00{\color{gray}\scriptsize$\pm$0.00}
& 6.30{\color{gray}\scriptsize$\pm$5.03} & 0.00{\color{gray}\scriptsize$\pm$0.00}
& 11.91{\color{gray}\scriptsize$\pm$3.30} & 0.00{\color{gray}\scriptsize$\pm$0.00}
& 0.41{\color{gray}\scriptsize$\pm$0.35} & 0.00{\color{gray}\scriptsize$\pm$0.00} \\

& C+I 
& 45.25{\color{gray}\scriptsize$\pm$3.67}
& 5.72{\color{gray}\scriptsize$\pm$2.81} & 0.10{\color{gray}\scriptsize$\pm$0.14}
& 12.47{\color{gray}\scriptsize$\pm$9.17} & 0.31{\color{gray}\scriptsize$\pm$0.07}
& 14.03{\color{gray}\scriptsize$\pm$0.39} & \textbf{0.75{\color{gray}\scriptsize$\pm$0.10}}
& 0.00{\color{gray}\scriptsize$\pm$0.00} & 0.25{\color{gray}\scriptsize$\pm$0.04} \\

& AgentVerse 
& 52.01{\color{gray}\scriptsize$\pm$2.06}
& 11.95{\color{gray}\scriptsize$\pm$6.38} & 0.00{\color{gray}\scriptsize$\pm$0.00}
& 21.55{\color{gray}\scriptsize$\pm$8.06} & 0.00{\color{gray}\scriptsize$\pm$0.00}
& 12.50{\color{gray}\scriptsize$\pm$1.38} & 0.00{\color{gray}\scriptsize$\pm$0.00}
& 18.38{\color{gray}\scriptsize$\pm$3.77} & 0.00{\color{gray}\scriptsize$\pm$0.00} \\

& EvoMAC
& 41.33{\color{gray}\scriptsize$\pm$4.47}
& 20.45{\color{gray}\scriptsize$\pm$4.72} & 0.00{\color{gray}\scriptsize$\pm$0.00}
& 24.29{\color{gray}\scriptsize$\pm$2.56} & 0.00{\color{gray}\scriptsize$\pm$0.00}
& 14.72{\color{gray}\scriptsize$\pm$7.17} & 0.00{\color{gray}\scriptsize$\pm$0.00}
& 5.80{\color{gray}\scriptsize$\pm$5.10} & 0.00{\color{gray}\scriptsize$\pm$0.00} \\

& \cellcolor{gray!10}\textbf{$\ours$} 
& \cellcolor{gray!10}\textbf{59.68{\color{gray}\scriptsize$\pm$1.84}} 
& \cellcolor{gray!10}\textbf{53.98{\color{gray}\scriptsize$\pm$1.55}}  
& \cellcolor{gray!10}\textbf{1.00{\color{gray}\scriptsize$\pm$0.00}} 
& \cellcolor{gray!10}\textbf{53.11{\color{gray}\scriptsize$\pm$1.82}}  
& \cellcolor{gray!10}\textbf{0.94{\color{gray}\scriptsize$\pm$0.04}} 
& \cellcolor{gray!10}\textbf{54.12{\color{gray}\scriptsize$\pm$1.58}}  
& \cellcolor{gray!10}\textbf{0.75{\color{gray}\scriptsize$\pm$0.13}} 
& \cellcolor{gray!10}\textbf{49.76{\color{gray}\scriptsize$\pm$0.32}}  
& \cellcolor{gray!10}\textbf{0.91{\color{gray}\scriptsize$\pm$0.02}} \\
\midrule

\multirow{5}{*}{\textbf{\begin{tabular}{c} GAIA \end{tabular}}} &
CoT
& 28.61{\color{gray}\scriptsize$\pm$3.57}
& 0.00{\color{gray}\scriptsize$\pm$0.00} & 0.00{\color{gray}\scriptsize$\pm$0.00}
& 1.01{\color{gray}\scriptsize$\pm$0.35} & 0.00{\color{gray}\scriptsize$\pm$0.00}
& 4.85{\color{gray}\scriptsize$\pm$1.60} & 0.00{\color{gray}\scriptsize$\pm$0.00}
& 0.61{\color{gray}\scriptsize$\pm$0.61} & 0.00{\color{gray}\scriptsize$\pm$0.00} \\

& LLM-Debate
& 32.32{\color{gray}\scriptsize$\pm$1.85}
& 13.13{\color{gray}\scriptsize$\pm$2.45} & 0.00{\color{gray}\scriptsize$\pm$0.00}
& 19.39{\color{gray}\scriptsize$\pm$2.10} & 0.00{\color{gray}\scriptsize$\pm$0.00}
& 15.96{\color{gray}\scriptsize$\pm$3.45} & 0.00{\color{gray}\scriptsize$\pm$0.00}
& 5.25{\color{gray}\scriptsize$\pm$0.93} & 0.00{\color{gray}\scriptsize$\pm$0.00} \\

& C+I
& 30.10{\color{gray}\scriptsize$\pm$0.92}
& 14.75{\color{gray}\scriptsize$\pm$3.11} & 0.53{\color{gray}\scriptsize$\pm$0.06}
& 20.81{\color{gray}\scriptsize$\pm$1.95} & 0.62{\color{gray}\scriptsize$\pm$0.04}
& 17.58{\color{gray}\scriptsize$\pm$5.97} & \textbf{0.99}{\color{gray}\scriptsize$\pm$0.00}
& 3.70{\color{gray}\scriptsize$\pm$1.91} & 0.72{\color{gray}\scriptsize$\pm$0.09} \\

& AgentVerse
& 24.31{\color{gray}\scriptsize$\pm$2.38}
& 13.18{\color{gray}\scriptsize$\pm$1.45} & 0.00{\color{gray}\scriptsize$\pm$0.00}
& 7.07{\color{gray}\scriptsize$\pm$1.75} & 0.00{\color{gray}\scriptsize$\pm$0.00}
& 11.31{\color{gray}\scriptsize$\pm$5.44} & 0.00{\color{gray}\scriptsize$\pm$0.00}
& 7.68{\color{gray}\scriptsize$\pm$6.48} & 0.00{\color{gray}\scriptsize$\pm$0.00} \\

& \cellcolor{gray!10}\textbf{$\ours$} 
& \cellcolor{gray!10}\textbf{32.73}{\color{gray}\scriptsize$\pm$1.60}
& \cellcolor{gray!10}\textbf{29.90}{\color{gray}\scriptsize$\pm$4.94} 
& \cellcolor{gray!10}\textbf{1.00}{\color{gray}\scriptsize$\pm$0.00}
& \cellcolor{gray!10}\textbf{26.46}{\color{gray}\scriptsize$\pm$3.55} 
& \cellcolor{gray!10}\textbf{0.83}{\color{gray}\scriptsize$\pm$0.02}
& \cellcolor{gray!10}\textbf{29.09}{\color{gray}\scriptsize$\pm$0.61} 
& \cellcolor{gray!10}0.92{\color{gray}\scriptsize$\pm$0.01}
& \cellcolor{gray!10}\textbf{24.44}{\color{gray}\scriptsize$\pm$4.59} 
& \cellcolor{gray!10}\textbf{0.94}{\color{gray}\scriptsize$\pm$0.01} \\

\midrule

\multirow{5}{*}{\textbf{\begin{tabular}{c} GPQA \\ Diamond \end{tabular}}} &
CoT   
& 64.14{\color{gray}\scriptsize$\pm$1.01}
& 0.00{\color{gray}\scriptsize$\pm$0.00} & 0.00{\color{gray}\scriptsize$\pm$0.00}
& 27.94{\color{gray}\scriptsize$\pm$0.58} & 0.00{\color{gray}\scriptsize$\pm$0.00}
& 30.97{\color{gray}\scriptsize$\pm$2.10} & 0.00{\color{gray}\scriptsize$\pm$0.00}
& 20.04{\color{gray}\scriptsize$\pm$1.54} & 0.00{\color{gray}\scriptsize$\pm$0.00} \\

& LLM-Debate
& 66.50{\color{gray}\scriptsize$\pm$1.27}
& 26.77{\color{gray}\scriptsize$\pm$7.02} & 0.00{\color{gray}\scriptsize$\pm$0.00}
& 35.52{\color{gray}\scriptsize$\pm$7.65} & 0.00{\color{gray}\scriptsize$\pm$0.00}
& 41.41{\color{gray}\scriptsize$\pm$4.01} & 0.00{\color{gray}\scriptsize$\pm$0.00}
& 15.32{\color{gray}\scriptsize$\pm$3.44} & 0.00{\color{gray}\scriptsize$\pm$0.00} \\

& C+I
& 65.15{\color{gray}\scriptsize$\pm$1.33}
& 35.35{\color{gray}\scriptsize$\pm$3.81} & 0.66{\color{gray}\scriptsize$\pm$0.09}
& 52.36{\color{gray}\scriptsize$\pm$2.04} & 0.52{\color{gray}\scriptsize$\pm$0.02}
& 51.67{\color{gray}\scriptsize$\pm$4.22} & 0.96{\color{gray}\scriptsize$\pm$0.03}
& 16.33{\color{gray}\scriptsize$\pm$2.49} & 0.66{\color{gray}\scriptsize$\pm$0.22} \\

& AgentVerse
& 62.46{\color{gray}\scriptsize$\pm$2.59}
& 28.62{\color{gray}\scriptsize$\pm$2.96} & 0.00{\color{gray}\scriptsize$\pm$0.00}
& 31.14{\color{gray}\scriptsize$\pm$3.04} & 0.00{\color{gray}\scriptsize$\pm$0.00}
& 27.95{\color{gray}\scriptsize$\pm$11.21} & 0.00{\color{gray}\scriptsize$\pm$0.00}
& 22.90{\color{gray}\scriptsize$\pm$6.11} & 0.00{\color{gray}\scriptsize$\pm$0.00} \\

& \cellcolor{gray!10}\textbf{$\ours$}
& \cellcolor{gray!10}\textbf{67.68}{\color{gray}\scriptsize$\pm$2.20}
& \cellcolor{gray!10}\textbf{61.87}{\color{gray}\scriptsize$\pm$1.07} 
& \cellcolor{gray!10}\textbf{1.00}{\color{gray}\scriptsize$\pm$0.00}
& \cellcolor{gray!10}\textbf{62.46}{\color{gray}\scriptsize$\pm$3.79} 
& \cellcolor{gray!10}\textbf{0.81}{\color{gray}\scriptsize$\pm$0.02}
& \cellcolor{gray!10}\textbf{61.11}{\color{gray}\scriptsize$\pm$2.53} 
& \cellcolor{gray!10}\textbf{0.98}{\color{gray}\scriptsize$\pm$0.01}
& \cellcolor{gray!10}\textbf{61.45}{\color{gray}\scriptsize$\pm$3.04} 
& \cellcolor{gray!10}\textbf{0.96}{\color{gray}\scriptsize$\pm$0.01} \\
\bottomrule
\end{tabular}
\end{adjustbox}
\caption{Performance under system uncertainty: The baselines and $\ours$ are evaluated on the GUI Dev., Web Dev., GAIA, and GPQA-Diamond benchmarks. Each method is run for three trials, and results are averaged across trials. The best performance for each benchmark is highlighted in \textbf{bold}.}
\label{tab:main}
\end{table*}

\section{Experiments}

\subsection{Experiment Setup} 
\noindent\textbf{Benchmarks.} We evaluate on four benchmarks spanning three domains: (\romannumeral 1) \textbf{software development} -- GUI Dev. and Web Dev., which we adapt from~\citet{mas:sw:rtadev} and~\citet{benchmark:rsde} into complex, long-horizon tasks; (\romannumeral 2) \textbf{tool use} -- GAIA~\cite{benchmark:gaia}; and (\romannumeral 3) \textbf{scientific reasoning} -- GPQA-Diamond~\cite{benchmark:gpqa}. Further details are provided in Appendix~\ref{app:benchmark}.

\noindent\textbf{System Uncertainty.} We consider four types of system uncertainty: (\romannumeral 1) \textbf{Agent Chaos} randomly terminates agents with a given probability~\cite{cloud:chaos_eng}; (\romannumeral 2) \textbf{Prompt Chaos} perturbs agent prompts through replacement, deletion, and shuffling~\cite{mas:res:cni}; (\romannumeral 3) \textbf{I/O Chaos} applies the same perturbations to agent outputs before they are consumed by other agents; and (\romannumeral 4) \textbf{Compound Chaos} applies all three system uncertainties.

Additionally, we evaluate on \textbf{adversarial attack} scenarios~\cite{mas:attack:agentxposed} in Table~\ref{tab:analysis:attack}, where an external agent injects malicious content into role prompts, task prompts, or outputs to mislead agent behavior; and \textbf{noisy tool} scenarios~\cite{agentnoisebench} in Table~\ref{tab:analysis:noisy_tools}, where tool operations suffer from execution failures and noisy responses.

%

\noindent\textbf{Evaluation Metrics.} To evaluate task performance, we report \textbf{Accuracy (Acc)}, the fraction of successfully completed tasks. We also report \textbf{Detection Recall (DR)}, the ratio of correctly identified affected agents to all affected agents. 


\noindent\textbf{Baselines.} We compare $\ours$ against four categories of baselines: 
(\romannumeral 1) \textbf{single-agent method} -- CoT~\cite{lm:reason:cot}; 
(\romannumeral 2) \textbf{static multi-agent systems} -- MetaGPT~\cite{mas:sw:metagpt}, RTADev~\cite{mas:sw:rtadev}, and LLM-Debate~\cite{mas:mr:llm_debate}; 
(\romannumeral 3) \textbf{autonomous multi-agent systems} -- AgentVerse~\cite{mas:auto:agentverse} and EvoMAC~\cite{benchmark:rsde};
(\romannumeral 4) \textbf{resilient multi-agent systems} -- C+I~\cite{mas:res:cni}.
%


We use \texttt{gpt-4.1-mini-0414}~\cite{LLM:GPT41mini} as the LLM for all baselines and $\ours$. Additionally, we utilize reasoning models of \texttt{o4-mini} and \texttt{gemini-3-flash}~\cite{LLM:gemini3} for ablation.

%

\begin{table*}[t]
\centering
\small
\begingroup
\setlength{\tabcolsep}{2.6pt}
\renewcommand{\arraystretch}{1.08}
\resizebox{\linewidth}{!}{
\begin{tabular}{@{}ll*{18}{c}@{}}
\toprule
\multicolumn{2}{c}{}
& \multicolumn{6}{c}{\textbf{Agent Chaos}}
& \multicolumn{6}{c}{\textbf{Prompt Chaos}}
& \multicolumn{6}{c}{\textbf{I/O Chaos}} \\
\cmidrule(lr){3-8}
\cmidrule(lr){9-14}
\cmidrule(lr){15-20}

& \textbf{}
& \multicolumn{2}{c}{\textbf{20\%}}
& \multicolumn{2}{c}{\textbf{40\%}}
& \multicolumn{2}{c}{\textbf{60\%}}
& \multicolumn{2}{c}{\textbf{Replace}}
& \multicolumn{2}{c}{\textbf{Delete}}
& \multicolumn{2}{c}{\textbf{Shuffle}}
& \multicolumn{2}{c}{\textbf{Replace}}
& \multicolumn{2}{c}{\textbf{Delete}}
& \multicolumn{2}{c}{\textbf{Shuffle}} \\
\cmidrule(lr){3-4}
\cmidrule(lr){5-6}
\cmidrule(lr){7-8}
\cmidrule(lr){9-10}
\cmidrule(lr){11-12}
\cmidrule(lr){13-14}
\cmidrule(lr){15-16}
\cmidrule(lr){17-18}
\cmidrule(lr){19-20}

& \textbf{Method}
& \textbf{Acc (\%)} & \textbf{DR}
& \textbf{Acc (\%)} & \textbf{DR}
& \textbf{Acc (\%)} & \textbf{DR}
& \textbf{Acc (\%)} & \textbf{DR}
& \textbf{Acc (\%)} & \textbf{DR}
& \textbf{Acc (\%)} & \textbf{DR}
& \textbf{Acc (\%)} & \textbf{DR}
& \textbf{Acc (\%)} & \textbf{DR}
& \textbf{Acc (\%)} & \textbf{DR} \\
\midrule

\multirow{3}{*}{\textbf{\begin{tabular}{c} GUI \\ Dev. \end{tabular}}}& 
RTADev
& 20.98 & 0.62 & 14.12 & 0.72 & 9.68 & 0.34
& 33.53 & 0.41 & 37.18 & 0.33 & 29.76 & 0.40
& 51.65 & 0.19 & 40.18 & 0.23 & 54.68 & 0.10 \\
& C+I
& 4.37 & 0.38 & 0.00 & 0.31 & 0.00 & 0.35
& 9.52 & 0.16 & 17.80 & 0.52 & 23.60 & 0.33
& 39.54 & 0.79 & 54.55 & \textbf{0.93} & 46.71 & \textbf{0.71} \\
& \cellcolor{gray!10}\textbf{$\ours$}
& \cellcolor{gray!10}\textbf{66.25} 
& \cellcolor{gray!10}\textbf{1.00} 
& \cellcolor{gray!10}\textbf{68.60} 
& \cellcolor{gray!10}\textbf{1.00} 
& \cellcolor{gray!10}\textbf{68.35} 
& \cellcolor{gray!10}\textbf{1.00}
& \cellcolor{gray!10}\textbf{59.73} 
& \cellcolor{gray!10}\textbf{0.60} 
& \cellcolor{gray!10}\textbf{54.73} 
& \cellcolor{gray!10}\textbf{0.62} 
& \cellcolor{gray!10}\textbf{52.89} 
& \cellcolor{gray!10}\textbf{0.72}
& \cellcolor{gray!10}\textbf{52.02} 
& \cellcolor{gray!10}\textbf{0.82} 
& \cellcolor{gray!10}\textbf{60.77} 
& \cellcolor{gray!10}0.83 
& \cellcolor{gray!10}\textbf{66.45} 
& \cellcolor{gray!10}0.58 \\

\midrule

\multirow{3}{*}{\textbf{\begin{tabular}{c} Web \\ Dev. \end{tabular}}}& 
RTADev
& 12.56 & 0.48 & 5.86 & 0.49 & 5.14 & 0.34
& 14.76 & 0.49 & 5.00 & 0.31 & 4.53 & 0.41
& 9.47 & 0.06 & 10.71 & 0.34 & 13.28 & 0.20 \\
& C+I
& 5.44 & 0.47 & 2.50 & 0.26 & 0.00 & 0.18
& 12.58 & 0.28 & 15.90 & 0.42 & 8.57 & 0.05
& 20.29 & \textbf{0.68} & 23.01 & \textbf{0.78} & 3.89 & 0.56 \\
& \cellcolor{gray!10}\textbf{$\ours$}
& \cellcolor{gray!10}\textbf{60.05} 
& \cellcolor{gray!10}\textbf{1.00} 
& \cellcolor{gray!10}\textbf{60.43} 
& \cellcolor{gray!10}\textbf{1.00} 
& \cellcolor{gray!10}\textbf{53.57} 
& \cellcolor{gray!10}\textbf{1.00}
& \cellcolor{gray!10}\textbf{52.22} 
& \cellcolor{gray!10}\textbf{0.95} 
& \cellcolor{gray!10}\textbf{51.86} 
& \cellcolor{gray!10}\textbf{1.00} 
& \cellcolor{gray!10}\textbf{46.11} 
& \cellcolor{gray!10}\textbf{0.93}
& \cellcolor{gray!10}\textbf{53.55} 
& \cellcolor{gray!10}0.55 
& \cellcolor{gray!10}\textbf{54.40} 
& \cellcolor{gray!10}0.65 
& \cellcolor{gray!10}\textbf{50.89} 
& \cellcolor{gray!10}\textbf{0.82} \\

\bottomrule
\end{tabular}
}
\endgroup
\caption{Performance under various chaos conditions}
\label{tab:analysis:chaos_diversity}
\end{table*}

\subsection{Main Results} \label{sec:5.2}
Table~\ref{tab:main} reports the performance of $\ours$ and the baselines across four benchmarks, under both the vanilla setting (i.e., nominal system conditions) and the four system uncertainty scenarios.
Under the vanilla setting, $\ours$ yields robust performance across all benchmarks, achieving an average improvement of $6.71$ pp in accuracy over the most competitive baseline C+I. This improvement is primarily attributed to cross-agent reflection, through which resilient proxies identify suboptimal intermediate outcomes and proactively trigger agent replacement, thereby preventing error propagation.

Across the four system uncertainty scenarios, $\ours$ consistently achieves the highest performance, outperforming C+I by $39.13$ pp, $25.60$ pp, $20.39$ pp, and $44.92$ pp in average accuracy, respectively. 
Further, $\ours$ attains substantially higher DR than C+I, except under the I/O chaos, indicating that $\ours$ correctly identifies affected agents. 
As CoT, MetaGPT, LLM-Debate, AgentVerse, and EvoMAC do not explicitly incorporate resilience mechanisms (e.g., the self-correction in C+I), we report their DR as 0.0.

In these experiments, CoT yields zero accuracy under agent chaos, since any single-point failure directly terminates task execution. 
In contrast, multi-agent systems are generally more robust, as failures of individual agents can be partially compensated for by other agents.
Yet, MetaGPT, LLM-Debate, and EvoMAC show limited performance, lacking explicit mechanisms to detect and recover from system uncertainty.
C+I and RTADev improve performance by incorporating self-correction: C+I introduces dedicated inspector and challenger agents to detect and correct erroneous outputs, while RTADev evaluates each new outcome against prior ones to refine faulty outcomes and maintain alignment across execution steps.
Unlike these correction-based mechanisms, $\ours$ performs cross-agent reflection to assess agents from the perspectives of their peers and addresses failures by replacing agents through role-based failover. As a result, the resilient proxies in $\ours$ function as infrastructure-level resilience mechanisms that mitigate error propagation and enforce role-based recovery.



%

\subsection{Analysis} \label{sec:5.3}
\noindent\textbf{Varying Chaos Conditions.}
Table~\ref{tab:analysis:chaos_diversity} reports the performance of RTADev, C+I, and $\ours$ under varying chaos conditions on both software development benchmarks.
%
For Agent Chaos, we vary its severity via termination ratios of 20\%, 40\%, and 60\%.
As the termination ratio increases, accuracy degrades sharply for the baselines, while $\ours$ maintains robust performance across all severity levels. This is due to role-based failover in $\ours$, which detects failures through timeouts and replaces the affected agents.
%
For Prompt Chaos, we apply three perturbation types, namely random word replacement, deletion, and shuffling, and $\ours$ consistently outperforms the baselines in all cases. 
Since agents are not merely response generators but actors with explicitly assigned roles, correction-based baselines are less effective under prompt corruption. In contrast, $\ours$ explicitly assesses peer outcomes with role-aware reflection criteria and recovers compromised agents via role-based failover, thereby showing improved performance.
For I/O Chaos, we apply the same perturbations as Prompt Chaos, and $\ours$ again outperforms the baselines. C+I occasionally attains higher DR than $\ours$ because its correction-based strategy over-detects affected agents. 
When an output is perturbed, part of its semantic content remains intact; $\ours$'s cross-agent reflection filters out such cases, while C+I still flags them as failures. 
As a result, although C+I exhibits higher DR in some settings, this gain comes at the cost of substantially lower accuracy than $\ours$.
\noindent\textbf{Adversarial Attack.}
In Table~\ref{tab:analysis:attack}, we model a malicious agent that crafts prompts or outputs to mislead agent behavior and disrupt coordination~\cite{mas:attack:agentxposed}. These attack scenarios target the semantic intent of role prompts, task prompts, or agent-generated outputs, steering agents toward adversarial behaviors.
%
%
%
%
%
As shown, $\ours$ maintains robust performance in all cases and outperforms C+I by $45.02$ pp in average accuracy, as cross-agent reflection detects semantic inconsistencies with assigned roles and role-based failover recovers affected agents.
%
%

\begin{table}[h]
\centering
\small
    \begin{adjustbox}{width=\columnwidth}
    \begin{tabular}{llcccccc}
    \toprule
    \multicolumn{2}{c}{\textbf{}} &
    \multicolumn{2}{c}{\textbf{Role Prompt}} &
    \multicolumn{2}{c}{\textbf{Task Prompt}} &
    \multicolumn{2}{c}{\textbf{Output}} \\
    \cmidrule(lr){3-4} \cmidrule(lr){5-6} \cmidrule(lr){7-8}
    & \textbf{Method} & \textbf{Acc (\%)} & \textbf{DR} & \textbf{Acc (\%)}& \textbf{DR} & \textbf{Acc (\%)}& \textbf{DR}\\
    \midrule
    \multirow{3}{*}{\textbf{\begin{tabular}{c} GUI \\ Dev. \end{tabular}}} 
    & RTADev & 28.64 & 0.66 & 48.82 & 0.40  & 20.29 & 0.33  \\
    & C+I & 6.80 & 0.24 & 8.52 & 0.24  & 4.44 & 0.46 \\
    & \cellcolor{gray!10}\textbf{$\ours$} & \cellcolor{gray!10}\textbf{63.95} & \cellcolor{gray!10}\textbf{0.93} & \cellcolor{gray!10}\textbf{52.98} & \cellcolor{gray!10}\textbf{0.52} & \cellcolor{gray!10}\textbf{52.14} & \cellcolor{gray!10}\textbf{0.56} \\
    \midrule
    \multirow{3}{*}{\textbf{\begin{tabular}{c} Web \\ Dev. \end{tabular}}}
    & RTADev & 11.19 & 0.59  & 18.22 & 0.39 & 7.56 & 0.45 \\
    & C+I & 15.14 & 0.25 & 6.11 & 0.10 & 4.13 & 0.44 \\
    & \cellcolor{gray!10}\textbf{$\ours$} & \cellcolor{gray!10}\textbf{48.00} & \cellcolor{gray!10}\textbf{0.91} & \cellcolor{gray!10}\textbf{53.77} & \cellcolor{gray!10}\textbf{0.92} & \cellcolor{gray!10}\textbf{44.39} & \cellcolor{gray!10}\textbf{0.60} \\
    \bottomrule
    \end{tabular}
    \end{adjustbox}
\caption{Performance w.r.t. adversarial attacks}
\label{tab:analysis:attack}
\end{table}
%


\noindent\textbf{Noisy Tool Scenarios.} Since multi-agent systems often utilize external tools for task completion, we evaluate performance on GAIA under two noisy tool scenarios, i.e., execution failure and noisy response~\cite{agentnoisebench}. 
As shown in Table~\ref{tab:analysis:noisy_tools}, $\ours$ outperforms C+I by $4.55$ pp in average accuracy, highlighting that $\ours$ remains resilient under tool failures that may be encountered in practical deployment.

%


\begin{table}[h]
\centering
\small
    \begin{adjustbox}{width=\columnwidth}
    \begin{tabular}{lccccc}
    \toprule
    & \textbf{Vanilla}
    & \multicolumn{2}{c}{\textbf{Execution Failure}}
    & \multicolumn{2}{c}{\textbf{Noisy Response}} \\
    \cmidrule(lr){2-2} \cmidrule(lr){3-4} \cmidrule(lr){5-6}
    \textbf{Method} & \textbf{Acc (\%)} & \textbf{Acc (\%)} & \textbf{DR} & \textbf{Acc (\%)} & \textbf{DR} \\
    \midrule
    LLM-Debate & 32.32 & 15.76 & 0.00 & 18.79 & 0.00 \\
    C+I        & 30.10 & 22.42 & 0.50 & 23.64 & 0.49 \\
    \cellcolor{gray!10}\textbf{$\ours$}       & \cellcolor{gray!10}\textbf{32.73} & \cellcolor{gray!10}\textbf{27.88} & \cellcolor{gray!10}\textbf{0.99} & \cellcolor{gray!10}\textbf{27.27} & \cellcolor{gray!10}\textbf{0.89} \\
    \bottomrule
    \end{tabular}
    \end{adjustbox}
\caption{Performance w.r.t. noisy tool scenarios}
\label{tab:analysis:noisy_tools}
\end{table}


%
\color{red}
\color{black}




\noindent\textbf{Augmenting Existing Methods with $\ours$.} In Table~\ref{tab:analysis:plug_in_resilience}, we verify that $\ours$ can be layered onto any existing multi-agent collaboration pattern. Specifically, we apply $\ours$ on top of the collaboration patterns of MetaGPT and RTADev, improving their performance by $43.74$ pp and $29.42$ pp on average, respectively. 
This is enabled by $\ours$'s code-based representation, which translates any existing collaboration pattern into a resilience-aware specification, and synthesizes it into executable code.

%

\begin{table}[h]
\centering
\small
    \begin{adjustbox}{width=\columnwidth}
    \begin{tabular}{lcccccc}
    \toprule
    & \multicolumn{2}{c}{\textbf{Agent Chaos}}
    & \multicolumn{2}{c}{\textbf{Prompt Chaos}}
    & \multicolumn{2}{c}{\textbf{I/O Chaos}} \\
    \cmidrule(lr){2-3} \cmidrule(lr){4-5} \cmidrule(lr){6-7}
    \textbf{Method} & \textbf{Acc (\%)} & \textbf{DR} & \textbf{Acc (\%)} & \textbf{DR} & \textbf{Acc (\%)} & \textbf{DR} \\
    \midrule
    MetaGPT      & 7.39 & 0.00 & 21.62 & 0.00 & 22.63 & 0.00 \\
    \cellcolor{gray!10}\hspace{0.4em} + $\ours$ 
    & \cellcolor{gray!10}66.86 & \cellcolor{gray!10}1.00 & \cellcolor{gray!10}56.90 & \cellcolor{gray!10}0.88 & \cellcolor{gray!10}59.09 & \cellcolor{gray!10}0.89 \\
    \midrule
    RTADev       & 21.87 & 0.61 & 39.64 & 0.42 & 33.90 & 0.19 \\
    \cellcolor{gray!10}\hspace{0.4em} + $\ours$ 
    & \cellcolor{gray!10}62.13 & \cellcolor{gray!10}1.00 & \cellcolor{gray!10}56.88 & \cellcolor{gray!10}0.75 & \cellcolor{gray!10}64.65 & \cellcolor{gray!10}0.91 \\
    \bottomrule
    \end{tabular}
    \end{adjustbox}
\caption{Augmenting existing methods with $\ours$}
\label{tab:analysis:plug_in_resilience}
\end{table}



\noindent\textbf{Cost Analysis.} Table~\ref{tab:analysis:cost} reports the API cost and wall-clock time of $\ours$ and the baselines, separated into the cost of agents and resilient components. For RTADev, which has no resilience mechanism, only agent cost is reported. 
Considering only the resilience components, $\ours$ incurs less than one-half the API cost of C+I. Even in total cost, $\ours$ is \$0.01 lower than RTADev, indicating that the resilient proxies add minimal overhead.
Moreover, $\ours$ is time efficient, as 23\% of its coordination structures adopt the parallel \& merge pattern (see Figure~\ref{fig:Distribution_coordination_structure}), enabling concurrent execution unlike the predominantly sequential execution employed by the baselines.
Further detailed cost analysis is provided in Appendix~\ref{app:cost}.

%
%

%
\begin{table}[h]
\centering
\small
    \begin{adjustbox}{width=\columnwidth}
    \begin{tabular}{lcccccc}
    \toprule
    & \multicolumn{3}{c}{\textbf{API Cost (\$)}} &
    \multicolumn{3}{c}{\textbf{Wall-clock Time (min)}} \\
    \cmidrule(lr){2-4} \cmidrule(lr){5-7}
    \textbf{Method} & \textbf{Agent} & \textbf{Resilient} & \textbf{Total} & \textbf{Agent} & \textbf{Resilient} & \textbf{Total} \\
    \midrule
    RTADev    
    & 0.180 & -- & 0.180 
    & 16.26 & -- & 16.26  \\
    C+I       
    & 0.049 & 0.155 & 0.204 
    & 5.12 & 3.14 & 8.26  \\
    \cellcolor{gray!10}\textbf{$\ours$}     
    & \cellcolor{gray!10}0.114 & \cellcolor{gray!10}0.055 & \cellcolor{gray!10}0.170 
    & \cellcolor{gray!10}3.76 & \cellcolor{gray!10}0.77 & \cellcolor{gray!10}4.53  \\
    \bottomrule
    \end{tabular}
    \end{adjustbox}
\caption{Inference cost analysis}
\label{tab:analysis:cost}
\end{table}
%

\subsection{Ablation Study}
\noindent\textbf{Ablation on Resilient Proxy.} 
Table~\ref{tab:ablation:framework} presents the impact of removing cross-agent reflection, role-based failover, and the full resilient proxy on the Web Dev., with all variants using the same synthesized workflow as $\ours$.
Even under the vanilla setting, removing these components degrades performance, since $\ours$ detects and mitigates errors from both suboptimal role assignment by the orchestrator and hallucinated agent outputs. The degradation becomes more pronounced under Compound Chaos. In particular, removing the full resilient proxy reduces accuracy to zero, since no mechanism remains to recover faulty agents, highlighting the effectiveness of $\ours$.
Further ablation results for the resilient proxy are provided in Appendix~\ref{app:extended_ablation}.

%
\begin{table}[h]
\centering
\small
    \begin{adjustbox}{width=\columnwidth}
    \begin{tabular}{lccc}
    \toprule
    &
    \multicolumn{1}{c}{\textbf{Vanilla}} &
    \multicolumn{2}{c}{\textbf{Compound Chaos}} \\
    \cmidrule(lr){2-2}\cmidrule(lr){3-4}
    \textbf{Method} & \textbf{Acc (\%)} & \textbf{Acc (\%)} & \textbf{DR} \\
    \midrule
    $\ours$
    & 59.68 & 49.76 & 0.91 \\
    \hspace{0.4em}\textendash\ Cross-agent Reflection
    & 49.89 & 17.57 & 0.57 \\
    \hspace{0.4em}\textendash\  Role-based Failover
    & 52.50 & 4.13 & 0.35 \\
    \hspace{0.4em}\textendash\  Resilient Proxy
    & 46.06 & 0.00 & 0.00 \\
    \bottomrule
    \end{tabular}
    \end{adjustbox}
\caption{Ablation on resilient proxy in $\ours$}
\label{tab:ablation:framework}
\end{table}

\noindent\textbf{Ablation on LLM.} In Table~\ref{tab:ablation:llm}, we vary the LLM used for agents and resilient proxies, adopting reasoning-oriented LLMs (e.g., \texttt{o4-mini}, \texttt{gemini-3-flash}) evaluated on GUI Dev. 
These models consistently improve performance across all cases. The gains appear even when applied only to resilient proxies, indicating that stronger reasoning enables more precise, role-aware feedback.


%
\begin{table}[h]
\centering
\small
    \begin{adjustbox}{width=\columnwidth}
    \begin{tabular}{llcccc}
    \toprule 
    & &
    \multicolumn{2}{c}{\textbf{Prompt Chaos}} &
    \multicolumn{2}{c}{\textbf{I/O Chaos}} \\
    \cmidrule(lr){3-4} \cmidrule(lr){5-6}
    \textbf{Target} & \textbf{LLM} & \textbf{Acc (\%)} & \textbf{DR} & \textbf{Acc (\%)}& \textbf{DR} \\
    \midrule
    \multirow{3}{*}{Task Agent}
    & \texttt{gpt-4.1-mini}     & 65.30 & 0.84 & 67.87 & 0.82  \\
    & \texttt{o4-mini}          & 78.91 & 0.76 & 77.08 & 0.85 \\
    & \texttt{gemini-3-flash}  & 73.56 & 0.94 & 79.08 & 1.00 \\
    \midrule
    \multirow{3}{*}{Resilient Proxy}
    & \texttt{gpt-4.1-mini}   & 65.30 & 0.84 & 67.87 & 0.82  \\
    & \texttt{o4-mini}    & 71.64 & 0.91 & 79.11 & 1.00 \\
    & \texttt{gemini-3-flash} & 69.50 & 0.71 & 85.81 & 0.94 \\
    \bottomrule
    \end{tabular}
    \end{adjustbox}
\caption{Ablation on LLM used in $\ours$}
\label{tab:ablation:llm}
\end{table}
\subsection{Case Study on Cloud-Native Testbed}
To validate $\ours$ beyond synthetic chaos scenarios, we use AIOpsLab~\cite{cloud:aiopslab}, a cloud-native testbed, to reproduce infrastructure-level system uncertainty that arises in cloud environments. Specifically, we instantiate each agent as a separate Kubernetes pod and use Chaos Mesh~\cite{chaos:mesh} to inject Stress Chaos, Disk I/O Chaos, and Network Chaos, which stress CPU and memory, disk access, and pod-to-pod communication, respectively.
As shown in Table~\ref{tab:case:aiopslab}, $\ours$ remains effective under such realistic chaos settings, as its service-mesh-inspired design aligns naturally with cloud-native infrastructure.


\begin{table}[h]
\centering
\small
    \begin{adjustbox}{width=\columnwidth}
    \begin{tabular}{lcccccc}
    \toprule
    & \multicolumn{2}{c}{\textbf{Stress Chaos}}
    & \multicolumn{2}{c}{\textbf{Disk I/O Chaos}}
    & \multicolumn{2}{c}{\textbf{Network Chaos}} \\
    \cmidrule(lr){2-3} \cmidrule(lr){4-5} \cmidrule(lr){6-7}
    \textbf{Method} & \textbf{Acc (\%)} & \textbf{DR} & \textbf{Acc (\%)} & \textbf{DR} & \textbf{Acc (\%)} & \textbf{DR} \\
    \midrule
    LLM-Debate   & 38.89 & 0.00 & 31.82 & 0.00 & 37.37 & 0.00 \\
    C+I         & 47.98 & 0.70 & 34.34 & 0.79 & 43.94 & 0.46  \\
    \cellcolor{gray!10}\textbf{$\ours$}      & \cellcolor{gray!10}\textbf{55.56} & \cellcolor{gray!10}\textbf{1.00}  & \cellcolor{gray!10}\textbf{60.61} & \cellcolor{gray!10}\textbf{1.00}  & \cellcolor{gray!10}\textbf{56.06} & \cellcolor{gray!10}\textbf{1.00} \\
    \bottomrule
    \end{tabular}
    \end{adjustbox}
\caption{Performance on cloud-native testbed}
\label{tab:case:aiopslab}
\end{table}


\section{Conclusion}
In this work, we presented the $\ours$ framework for constructing resilient multi-agent systems that systematically address partial failures and behavioral uncertainty.
By decoupling collaboration logic from resilience policy in a structured specification, $\ours$ synthesizes executable code that overlays a role-based self-healing layer on top of a collaboration pattern.
This layer is realized through resilient proxies which integrate cross-agent reflection for role-specific  degradation detection with role-based failover for recovery.
Through extensive experiments on diverse system uncertainty scenarios, together with a case study on a cloud-native testbed, we demonstrate that $\ours$ consistently outperforms existing multi-agent systems in both task performance and resilience.

%

\section*{Limitations}
As shown in Table~\ref{tab:ablation:llm}, employing more capable reasoning models solely within the resilient proxies consistently improves performance. In the current design, reasoning models are assigned statically, regardless of task difficulty or runtime conditions. Allocating them dynamically, invoking stronger reasoning only when needed, remains an important direction for future work. Another limitation is the assumption that the resilient proxies and the meta-agent remain available throughout execution; such failures fall outside the current threat model and are delegated to infrastructure-level fault tolerance.


\section*{Ethical Considerations}
This work seeks to characterize and reduce failure modes in LLM-based multi-agent collaboration, enabling more robust orchestration. 
The chaos engineering techniques introduced in this study serve as experimental tools to evaluate the resilience of multi-agent systems through stochastic perturbations that simulate system uncertainty. Although these techniques could potentially be exploited to disrupt agent behaviors or corrupt outputs, such risks are mitigated by resilience mechanisms designed to detect uncertainty.
To ensure transparency and reproducibility, a detailed description of each chaos scenario is provided in Appendix~\ref{sec:app:d:1}. These techniques are intended solely for defensive evaluation and academic research, and should not be used to enable malicious interference.

\section*{Acknowledgments}
    This work was supported by Institute of Information \& communications Technology Planning \& Evaluation (IITP) grant funded by the Korea government (MSIT),
    (RS-2022-II220043, Adaptive Personality for Intelligent Agents,
    RS-2022-II221045, Self-directed multi-modal Intelligence for solving unknown, open domain problems,
    RS-2025-02218768, Accelerated Insight Reasoning via Continual Learning,
    RS-2025-25442569, AI Star Fellowship Support Program (Sungkyunkwan Univ.)),
    the National Research Foundation of Korea (NRF) grant funded by the Korea government (MSIT) (No. RS-2026-25474409),
    IITP-ITRC (Information Technology Research Center) grant funded by the Korea government (MSIT) (IITP-2026-RS-2024-00437633, 10\%),
    and by Samsung Electronics Co., Ltd.


\bibliography{custom}

@article{mas:mas_google,
  author    = "Yubin Kim and Ken Gu and Chanwoo Park and Chunjong Park and Samuel Schmidgall and A. Ali Heydari and Yao Yan and Zhihan Zhang and Yuchen Zhuang and Yun Liu and Mark Malhotra and Paul Pu Liang and Hae Won Park and Yuzhe Yang and Xuhai Xu and Yilun Du and Shwetak Patel and Tim Althoff and Daniel McDuff and Xin Liu",
  year      = 2025,
  title     = "Towards a Science of Scaling Agent Systems",
  journal   = "arXiv preprint arXiv:2512.08296",
}

@inproceedings{agentnoisebench,
  author={Ruipeng Wang and Yuxin Chen and Yukai Wang and Chang Wu and Junfeng Fang and Xiaodong Cai and Qi GU and Hui Su and An Zhang and Xiang Wang and Xunliang Cai and Tat-Seng Chua},
  year      = 2026,
  title     = "AgentNoiseBench: Benchmarking Robustness of Tool-Using LLM Agents Under Noisy Condition",
  booktitle={Forty-third International Conference on Machine Learning},
}

@inproceedings{cloud:aiopslab,
author = {Chen, Yinfang and Shetty, Manish and Somashekar, Gagan and Ma, Minghua and Simmhan, Yogesh and Mace, Jonathan and Bansal, Chetan and Wang, Rujia and Rajmohan, Saravan},
  year    = {2025},
  title = {AIOpsLab: A Holistic Framework to Evaluate AI Agents for Enabling Autonomous Clouds},
  booktitle = {Proceedings of Machine Learning and Systems},
}

@inproceedings{mas:mas_autogen,
    author={Qingyun Wu and Gagan Bansal and Jieyu Zhang and Yiran Wu and Beibin Li and Erkang Zhu and Li Jiang and Xiaoyun Zhang and Shaokun Zhang and Jiale Liu and Ahmed Hassan Awadallah and Ryen W White and Doug Burger and Chi Wang},   
    year=2024,
    title={AutoGen: Enabling Next-Gen {LLM} Applications via Multi-Agent Conversations},
    booktitle={First Conference on Language Modeling}
}

@inproceedings{mas:sw:metagpt,
  author    = "Sirui Hong and
            Mingchen Zhuge and
            Jonathan Chen and
            Xiawu Zheng and
            Yuheng Cheng and
            Jinlin Wang and
            Ceyao Zhang and
            Zili Wang and
            Steven Ka Shing Yau and
            Zijuan Lin and
            Liyang Zhou and
            Chenyu Ran and
            Lingfeng Xiao and
            Chenglin Wu and
            J{\"{u}}rgen Schmidhuber",
  year      = 2024,
  title     = "MetaGPT: Meta Programming for {A} Multi-Agent Collaborative Framework",
  booktitle = "Proceedings of the 12th International Conference on Learning Representations",
}

@inproceedings{mas:sw:rtadev,
  author    = "Liu, Jie  and
      Wang, Guohua  and
      Yang, Ronghui  and
      Zeng, Jiajie  and
      Zhao, Mengchen  and
      Cai, Yi",
  year      = 2025,
  title     = "{RTADev}: Intention Aligned Multi-Agent Framework for Software Development",
  booktitle = "Findings of the Association for Computational Linguistics: {ACL}",
}

@inproceedings{mas:sd:mtmol,
  author    = "Hyomin Kim and
            Yunhui Jang and
            Sungsoo Ahn",
  year      = 2025,
  title     = "MT-Mol: Multi Agent System with Tool-based Reasoning for Molecular Optimization",
  booktitle = "Findings of the Association for Computational Linguistics: EMNLP",
}

@inproceedings{mas:sd:agentlab,
    author = "Schmidgall, Samuel  and
      Su, Yusheng  and
      Wang, Ze  and
      Sun, Ximeng  and
      Wu, Jialian  and
      Yu, Xiaodong  and
      Liu, Jiang  and
      Moor, Michael  and
      Liu, Zicheng  and
      Barsoum, Emad",
  year      = 2025,
  title     = "Agent Laboratory: Using {LLM} Agents as Research Assistants",
  booktitle = "Findings of the Association for Computational Linguistics: EMNLP"
}

@inproceedings{mas:mr:llm_debate,
  author    = "Yilun Du and
            Shuang Li and
            Antonio Torralba and
            Joshua B. Tenenbaum and
            Igor Mordatch",
  year      = 2024,
  title     = "Improving Factuality and Reasoning in Language Models through Multiagent Debate",
  booktitle = "Proceedings of the 41st International Conference on Machine Learning",
}

@inproceedings{mas:mr:macm,
  author    = "Bin Lei and
            Yi Zhang and
            Shan Zuo and
            Ali Payani and
            Caiwen Ding",
  year      = 2024,
  title     = "{MACM:} Utilizing a Multi-Agent System for Condition Mining in Solving Complex Mathematical Problems",
  booktitle = "Proceedings of the 38th Annual Conference on Neural Information Processing Systems",
}

@inproceedings{mas:auto:agentverse,
  author    = "Weize Chen and
            Yusheng Su and
            Jingwei Zuo and
            Cheng Yang and
            Chenfei Yuan and
            Chi{-}Min Chan and
            Heyang Yu and
            Yaxi Lu and
            Yi{-}Hsin Hung and
            Chen Qian and
            Yujia Qin and
            Xin Cong and
            Ruobing Xie and
            Zhiyuan Liu and
            Maosong Sun and
            Jie Zhou",
  year      = 2024,
  title     = "AgentVerse: Facilitating Multi-Agent Collaboration and Exploring Emergent Behaviors",
  booktitle = "Proceedings of the 12th International Conference on Learning Representations",
}

@inproceedings{mas:auto:autoagents,
  author    = "Guangyao Chen and
            Siwei Dong and
            Yu Shu and
            Ge Zhang and
            Jaward Sesay and
            B{\"{o}}rje Karlsson and
            Jie Fu and
            Yemin Shi",
  year      = 2024,
  title     = "AutoAgents: A Framework for Automatic Agent Generation",
  booktitle = "Proceedings of the 33rd International Joint Conference on Artificial Intelligence",
}

@inproceedings{mas:auto:maas,
  author    = "Guibin Zhang and
            Luyang Niu and
            Junfeng Fang and
            Kun Wang and
            Lei Bai and
            Xiang Wang",
  year      = 2025,
  title     = "Multi-agent Architecture Search via Agentic Supernet",
  booktitle = "Proceedings of the 42nd International Conference on Machine Learning",
}

@inproceedings{mas:auto:aflow,
  author    = "Jiayi Zhang and
            Jinyu Xiang and
            Zhaoyang Yu and
            Fengwei Teng and
            Xionghui Chen and
            Jiaqi Chen and
            Mingchen Zhuge and
            Xin Cheng and
            Sirui Hong and
            Jinlin Wang and
            Bingnan Zheng and
            Bang Liu and
            Yuyu Luo and
            Chenglin Wu",
  year      = 2025,
  title     = "AFlow: Automating Agentic Workflow Generation",
  booktitle = "Proceedings of the 13th International Conference on Learning Representations",
}

@inproceedings{mas:res:cni,
  author    = "Jen{-}tse Huang and
            Jiaxu Zhou and
            Tailin Jin and
            Xuhui Zhou and
            Zixi Chen and
            Wenxuan Wang and
            Youliang Yuan and
            Michael R. Lyu and
            Maarten Sap",
  year      = 2025,
  title     = "On the Resilience of LLM-Based Multi-Agent Collaboration with Faulty Agents",
  booktitle = "Proceedings of the 42nd International Conference on Machine Learning",
}

@inproceedings{mas:res:netsafe,
    author = "Yu, Miao  and
      Wang, Shilong  and
      Zhang, Guibin  and
      Mao, Junyuan  and
      Yin, Chenlong  and
      Liu, Qijiong  and
      Wang, Kun  and
      Wen, Qingsong  and
      Wang, Yang",
  year      = 2025,
  title     = "{N}et{S}afe: Exploring the Topological Safety of Multi-agent System",
  booktitle = "Findings of the Association for Computational Linguistics: ACL"
}

@article{mas:attack:agentxposed,
  author    = "Yizhe Xie and Congcong Zhu and Xinyue Zhang and Tianqing Zhu and Dayong Ye and Minghao Wang and Chi Liu",
  year      = 2025,
  title     = "Who's the Mole? Modeling and Detecting Intention-Hiding Malicious Agents in LLM-Based Multi-Agent Systems",
  journal   = "arXiv preprint arXiv:2507.04724",
}

@inproceedings{lm:reason:cot,
  author    = "Jason Wei and
            Xuezhi Wang and
            Dale Schuurmans and
            Maarten Bosma and
            Brian Ichter and
            Fei Xia and
            Ed H. Chi and
            Quoc V. Le and
            Denny Zhou",
  year      = 2022,
  title     = "Chain-of-Thought Prompting Elicits Reasoning in Large Language Models",
  booktitle = "Proceedings of the 36th Annual Conference on Neural Information Processing Systems",
}

@book{infraascode:morris,
  title={Infrastructure as code: managing servers in the cloud},
  author={Morris, Kief},
  year={2016},
  publisher={O'Reilly Media, Inc.}
}

@article{cloud:chaos_eng,
  author    = "Ali Basiri and
            Niosha Behnam and
            Ruud de Rooij and
            Lorin Hochstein and
            Luke Kosewski and
            Justin Reynolds and
            Casey Rosenthal",
  year      = 2016,
  title     = "Chaos Engineering",
  journal   = "{IEEE} Software",
  volume    = {33},
  number    = {3},
  pages     = {35--41},
}

@article{cloud:mobile_service,
  title={On service resilience in cloud-native 5G mobile systems},
  author={Taleb, Tarik and Ksentini, Adlen and Sericola, Bruno},
  journal={IEEE Journal on Selected Areas in Communications},
  volume={34},
  number={3},
  pages={483--496},
  year={2016},
  publisher={IEEE}
}

@inproceedings{cloud:servicemesh,
  author    = "Wubin Li and
               Yves Lemieux and
               Jing Gao and
               Zhuofeng Zhao and
               Yanbo Han",
  year      = 2019,
  title     = "Service Mesh: Challenges, State of the Art, and Future Research Opportunities",
  booktitle = "Proceedings of the 2019 IEEE International Conference on Service-Oriented System Engineering",
}

@book{cloud:designpatterns,
  author    = "Homer, Alex and Sharp, John and Brader, Larry and Narumoto, Masashi and Swanson, Trent",
  year      = 2014,
  title     = "Cloud Design Patterns: Prescriptive Architecture Guidance for Cloud Applications",
  publisher = "Microsoft",
}

@inproceedings{benchmark:rsde,
  author    = {Yue Hu and
            Yuzhu Cai and
            Yaxin Du and
            Xinyu Zhu and
            Xiangrui Liu and
            Zijie Yu and
            Yuchen Hou and
            Shuo Tang and
            Siheng Chen},
  year      = 2025,
  title     = {Self-Evolving Multi-Agent Collaboration Networks for Software Development},
  booktitle = {Proceedings of the 13th International Conference on Learning Representations},
}

@inproceedings{benchmark:gaia,
  title= "{GAIA}: A Benchmark for General {AI} Assistants",
  author={Grégoire Mialon and Clémentine Fourrier and Craig Swift and Thomas Wolf and Yann LeCun and Thomas Scialom},
  booktitle={International Conference on Learning Representations},
  year={2024}
}

@article{benchmark:gpqa,
  author    = "David Rein and
            Betty Li Hou and
            Asa Cooper Stickland and
            Jackson Petty and
            Richard Yuanzhe Pang and
            Julien Dirani and
            Julian Michael and
            Samuel R. Bowman",
  year      = 2023,
  title     = "{GPQA:} {A} Graduate-Level Google-Proof Q{\&}A Benchmark",
  journal   = "arXiv preprint arXiv:2311.12022",
}

@article{benchmark:humaneval,
  author    = "Mark Chen and Jerry Tworek and Heewoo Jun and Qiming Yuan and Henrique Ponde de Oliveira Pinto and Jared Kaplan and Harri Edwards and Yuri Burda and Nicholas Joseph and Greg Brockman and Alex Ray and Raul Puri and Gretchen Krueger and Michael Petrov and Heidy Khlaaf and Girish Sastry and Pamela Mishkin and Brooke Chan and Scott Gray and Nick Ryder and Mikhail Pavlov and Alethea Power and Lukasz Kaiser and Mohammad Bavarian and Clemens Winter and Philippe Tillet and Felipe Petroski Such and Dave Cummings and Matthias Plappert and Fotios Chantzis and Elizabeth Barnes and Ariel Herbert-Voss and William Hebgen Guss and Alex Nichol and Alex Paino and Nikolas Tezak and Jie Tang and Igor Babuschkin and Suchir Balaji and Shantanu Jain and William Saunders and Christopher Hesse and Andrew N. Carr and Jan Leike and Josh Achiam and Vedant Misra and Evan Morikawa and Alec Radford and Matthew Knight and Miles Brundage and Mira Murati and Katie Mayer and Peter Welinder and Bob McGrew and Dario Amodei and Sam McCandlish and Ilya Sutskever and Wojciech Zaremba",
  year      = 2021,
  title     = "Evaluating Large Language Models Trained on Code",
  journal   = "arXiv preprint arXiv:2107.03374",
}

@article{benchmark:mbpp,
  author    = "Jacob Austin and
                  Augustus Odena and
                  Maxwell I. Nye and
                  Maarten Bosma and
                  Henryk Michalewski and
                  David Dohan and
                  Ellen Jiang and
                  Carrie J. Cai and
                  Michael Terry and
                  Quoc V. Le and
                  Charles Sutton",
  year      = 2021,
  title     = "Program Synthesis with Large Language Models",
  journal   = "arXiv preprint arXiv:2108.07732",
}

@inproceedings{planning:saycan,
  author={Michael Ahn and Anthony Brohan and Noah Brown and Yevgen Chebotar and Omar Cortes and Byron David and Chelsea Finn and Chuyuan Fu and Keerthana Gopalakrishnan and Karol Hausman and Alex Herzog and Daniel Ho and Jasmine Hsu and Julian Ibarz and Brian Ichter and Alex Irpan and Eric Jang and Rosario Jauregui Ruano and Kyle Jeffrey and Sally Jesmonth and Nikhil J Joshi and Ryan Julian and Dmitry Kalashnikov and Yuheng Kuang and Kuang-Huei Lee and Sergey Levine and Yao Lu and Linda Luu and Carolina Parada and Peter Pastor and Jornell Quiambao and Kanishka Rao and Jarek Rettinghouse and Diego Reyes and Pierre Sermanet and Nicolas Sievers and Clayton Tan and Alexander Toshev and Vincent Vanhoucke and Fei Xia and Ted Xiao and Peng Xu and Sichun Xu and Mengyuan Yan and Andy Zeng},
  year      = 2023,
  title     = "Do As {I} Can, Not As {I} Say: Grounding Language in Robotic Affordances",
  booktitle = "Proceedings of the 6th Conference on Robot Learning"
}

@article{mm_reasoning:multiCoT,
  author    = "Zhang, Zhuosheng and Zhang, Aston and Li, Mu and Zhao, Hai and Karypis, George and Smola, Alex",
  year      = 2024,
  title     = "Multimodal Chain-of-Thought Reasoning in Language Models",
  journal   = "Transactions on Machine Learning Research",
}

@article{coding:codellama,
  author    = "Baptiste Rozière and Jonas Gehring and Fabian Gloeckle and Sten Sootla and Itai Gat and Xiaoqing Ellen Tan and Yossi Adi and Jingyu Liu and Romain Sauvestre and Tal Remez and Jérémy Rapin and Artyom Kozhevnikov and Ivan Evtimov and Joanna Bitton and Manish Bhatt and Cristian Canton Ferrer and Aaron Grattafiori and Wenhan Xiong and Alexandre Défossez and Jade Copet and Faisal Azhar and Hugo Touvron and Louis Martin and Nicolas Usunier and Thomas Scialom and Gabriel Synnaeve",
  year      = 2023,
  title     = "Code Llama: Open Foundation Models for Code",
  journal   = "arXiv preprint arXiv:2308.12950",
}

@misc{chaos:mesh,
  author       = {{Chaos Mesh}},
  title        = {{Chaos Mesh}: A Chaos Engineering Platform for Kubernetes},
  year         = {2021},
  howpublished = {\url{https://github.com/chaos-mesh/chaos-mesh}},
}

@misc{chaos:monkey,
  author       = {{Netflix}},
  title        = {{Chaos Monkey}},
  year         = {2012},
  howpublished = {\url{https://github.com/Netflix/chaosmonkey}},
}

@inproceedings{mas:camel,
  author    = "Guohao Li and
            Hasan Abed Al Kader Hammoud and
            Hani Itani and
            Dmitrii Khizbullin and
            Bernard Ghanem",
  year      = 2023,
  title     = "CAMEL: Communicative Agents for 'Mind' Exploration of Large Language Model Society",
  booktitle = "Proceedings of the 37th Conference on Neural Information Processing Systems",
}

@inproceedings{mas:owl,
  author={Mengkang Hu and Yuhang Zhou and Wendong Fan and Yuzhou Nie and Ziyu Ye and Bowei Xia and Tao Sun and Zhaoxuan Jin and Yingru Li and Zeyu Zhang and Yifeng Wang and Qianshuo Ye and Bernard Ghanem and Ping Luo and Guohao Li},
  year      = 2025,
  title     = "OWL: Optimized Workforce Learning for General Multi-Agent Assistance in Real-World Task Automation",
  booktitle = "Proceedings of the 39th Conference on Neural Information Processing Systems",
}

@inproceedings{attack:redteaming,
  author    = "Pengfei He and
               Yuping Lin and
               Shen Dong and
               Han Xu and
               Yue Xing and
               Hui Liu",
  year      = 2025,
  title     = "Red-Teaming LLM Multi-Agent Systems via Communication Attacks",
  booktitle = "Findings of the Association for Computational Linguistics: {ACL}",
}

@misc{LLM:GPT41mini,
  author = {OpenAI},
  title = {Introducing {GPT-4.1} in the {API}},
  howpublished = {\url{https://openai.com/index/gpt-4-1/}},
  year = {2025}
}

@misc{LLM:gemini3,
  title={Gemini 3 Flash: frontier intelligence built for speed},
  author={Google},
howpublished = {\url{https://blog.google/products-and-platforms/products/gemini/gemini-3-flash/}},
  year=2025
}

@misc{LLM:claude-code,
  author       = {{Anthropic}},
  title        = {{Claude Code}},
  year         = {2026},
  howpublished = {\url{https://code.claude.com/}},
}

\clearpage

\appendix

\label{sec:appendix}

\section{Related Work}
\noindent\textbf{Multi-agent System.}
LLM-based multi-agent systems have emerged as an effective paradigm for solving complex tasks by assigning specialized roles to multiple agents and coordinating their interactions.
Existing systems often encode such collaboration through manually designed roles, task handoffs, and communication protocols.
LLM-Debate~\cite{mas:mr:llm_debate} adopts a solver-judge debate structure, where solver agents independently generate responses, iteratively critique one another, and a judge agent synthesizes the final answer. MetaGPT~\cite{mas:sw:metagpt} formalizes software development as an SOP-based workflow, where fixed roles such as product manager, architect, engineer, and QA engineer sequentially exchange artifacts across development stages. RTADev~\cite{mas:sw:rtadev} extends this static development workflow with intention-alignment checks, where reviewer agents inspect intermediate artifacts and trigger discussion when downstream outputs diverge from prior role intents. While this role-based design enables structured collaboration, static coordination structures make it difficult to adapt the system beyond predefined task scopes. 
Accordingly, AgentVerse~\cite{mas:auto:agentverse} automates agent-team coordination through dynamic expert recruitment, collaborative decision-making, action execution, and evaluation. AutoAgents~\cite{mas:auto:autoagents} dynamically generates specialized agents from task content and coordinates them through task-specific planning. AFlow~\cite{mas:auto:aflow} represents agentic workflows as executable code and optimizes them through Monte Carlo Tree Search with execution feedback. MaAS~\cite{mas:auto:maas} searches over an agentic supernet to sample query-dependent multi-agent architectures with task-specific resource allocation. EvoMAC~\cite{benchmark:rsde} self-evolves multi-agent collaboration networks by converting execution feedback into textual gradients that update agent prompts, connections, and workflows during test-time execution.
However, these methods assume nominal system conditions and do not explicitly address failures or behavioral uncertainty during runtime.
In contrast, $\ours$ extends a collaboration pattern with resilience policies for runtime monitoring and recovery under system uncertainty.


\noindent\textbf{Cloud Native Resilience.}
Resilience has long been a fundamental design consideration in large-scale  cloud and distributed systems~\cite{cloud:chaos_eng,cloud:mobile_service}. 
In particular, the \textit{service mesh}~\cite{cloud:servicemesh} has emerged as a principled architectural approach to achieve system-level resilience by decoupling resilience logic from microservice applications. 
The service mesh explicitly separates a control plane, which manages resilience policies through declarative configurations, from a data plane, where sidecar proxies attached to microservices intercept service-to-service communication and provide fault-tolerance mechanisms such as retries, timeouts, and circuit breaking~\cite{cloud:designpatterns}. 
$\ours$ adapts this cloud-native principle to multi-agent systems through an infrastructure-as-code (IaC) approach~\cite{infraascode:morris}.
In this sense, the artifact constructed through IaC is a service-mesh-style infrastructure, where synthesized policy code instantiates sidecar-like resilient proxies around task-executing agents.
This enables infrastructure-level enforcement of resilience policies for multi-agent systems, without embedding resilience logic directly into agent implementations.

\begin{table*}[h]
\centering
\small
    \begin{tabular}{llccccc}
    \toprule
    \multirow{2}{*}{\textbf{Type}} & \multirow{2}{*}{\textbf{Benchmark}} &
    \multicolumn{3}{c}{\textbf{Task}} & \multicolumn{2}{c}{\textbf{Test Case}} \\
    \cmidrule(lr){3-5}\cmidrule(lr){6-7}
    & & \textbf{Amount} & \textbf{Length} & \textbf{Functional Units} & \textbf{Basic} & \textbf{Advanced} \\
    \midrule
    \multirow{2}{*}{Web}
    & rSDE-Bench & 45 & 4,119 & 5.50 & 6.49 & 5.49 \\
    & Ours       & 20 & 8,793 & 8.75 & -- & 9.00 \\
    \midrule
    \multirow{2}{*}{GUI}
    & FSD-Bench & 41 & 927 & 5.35 & -- & -- \\ 
    & Ours      & 21 & 15,880 & 6.67 & -- & 11.04\\ 
    \bottomrule
    \end{tabular}
\caption{Comparison between existing benchmarks (rSDE-Bench and FSD-Bench) and our proposed benchmark}
\label{tab:appendix:a:1:sw_bench_stats}
\end{table*}

%
\begin{figure}[t]
\centering
    \includegraphics[width=\columnwidth]{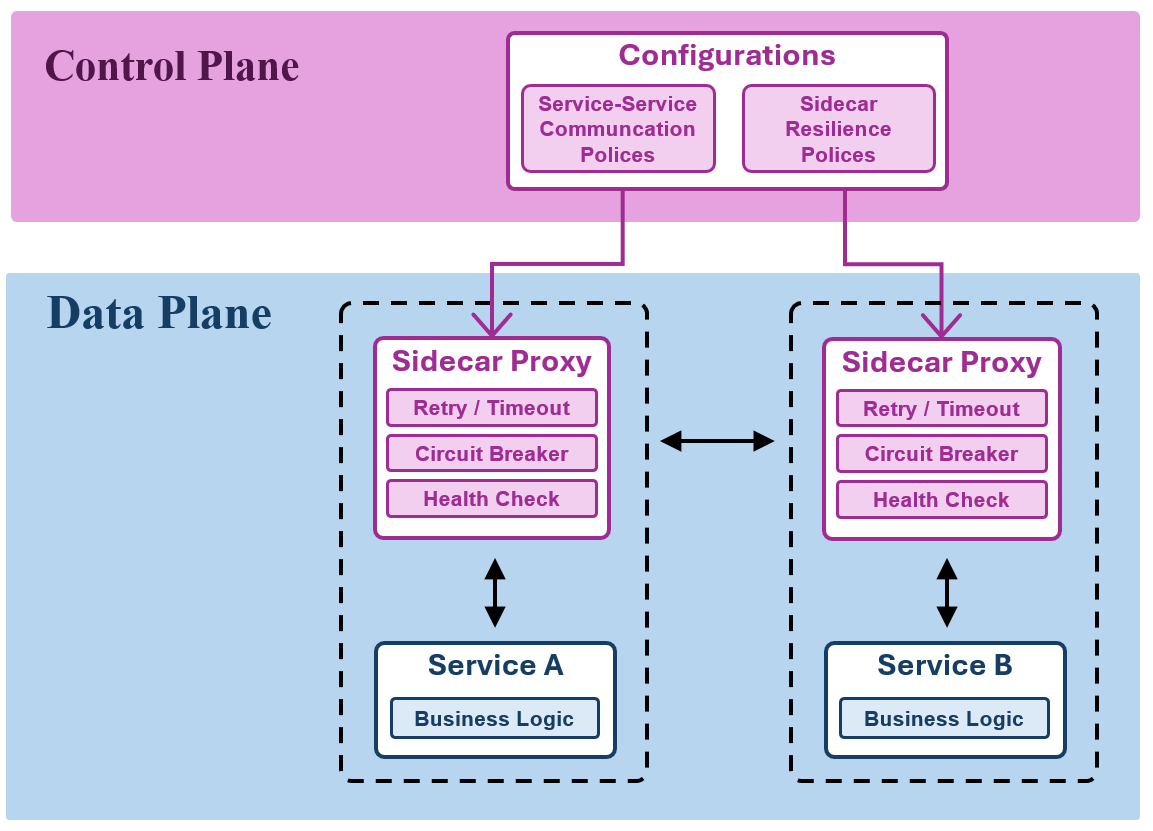}
    \caption{Cloud-Native Service Mesh}
    \label{fig:servicemesh}
\end{figure}
\section{Service Mesh-inspired $\ours$}
Large-scale applications are often deployed as cloud-native microservices, where inter-service dependencies allow partial failures to propagate across downstream services. Such failure propagation motivates the evaluation of system-level resilience -- the system's capability to maintain service availability despite failures in individual services. Chaos engineering evaluates this resilience by injecting controlled failures into target services during execution, such as service crashes, network disruptions, and timeouts.

To improve resilience under such failures, a \textit{service mesh} provides a cloud-native infrastructure layer for managing service-to-service communication and resilience policies. As shown in Figure~\ref{fig:servicemesh}, the control plane manages these policies, while sidecar proxies in the data plane mediate service interactions and detect partial failures through mechanisms such as retries, timeouts, circuit breaking, failover, and health checking. This separation decouples business logic from resilience logic, enabling runtime failure mitigation without modifying application code.

$\ours$ adopts the control–data plane separation and the sidecar-proxy pattern from service meshes, attaching a resilient proxy to each task-executing agent to mediate inter-agent communication. To adapt this pattern to LLM-based collaboration, cross-agent reflection replaces predefined local health checks with distributed LLM-based evaluation under role-specific criteria. Role-based failover likewise refines the role specification of the replacement agent rather than reinstantiating an identical replica.

%

%

\section{Benchmark Details}\label{app:benchmark}
In this section, we present the benchmarks used to evaluate our method, including software development (GUI Dev. and Web Dev.), tool use (GAIA), and scientific reasoning (GPQA-Diamond).

\subsection{Software Development}
To evaluate the coding capabilities of LLMs, prior studies have traditionally relied on benchmarks such as MBPP~\cite{benchmark:mbpp} and HumanEval~\cite{benchmark:humaneval}, which target relatively simple programming tasks.
However, these benchmarks have become trivial for modern LLMs; for instance, \citet{mas:auto:maas} report that a CoT with \texttt{gpt-4o-mini} already attains success rates of 87.08\% on HumanEval and 71.88\% on MBPP.


In response, more challenging benchmarks such as FSD-Bench~\cite{mas:sw:rtadev} and rSDE-Bench~\cite{benchmark:rsde} have recently been introduced to better assess multi-agent systems on software development tasks. 
Given a requirement specification, models must produce complete implementations that pass a suite of automated tests, closely mirroring real-world development workflows. 
%
Yet even these benchmarks are beginning to saturate. As shown in Table~\ref{tab:appendix:a:1:rSDE}, a single-agent setup with \texttt{gpt-4o-mini}, the prevailing model in 2024, achieved only 53.65\% accuracy on rSDE-Bench, whereas \texttt{gpt-4.1-mini} reached 79.55\% under the same setting within a year.

%

%
\begin{table}[h]
\centering
\small
    \begin{tabular}{lccc}
    \toprule
    \textbf{Model} & \textbf{Release Date} & \textbf{Accuracy (\%)} \\
    \midrule
    gpt-4o-mini     & 2024.07 & 53.65 \\
    gpt-4.1-mini    & 2025.04 & 79.55 \\
    \bottomrule
    \end{tabular}
\caption{Performance saturation on rSDE-Bench}
\label{tab:appendix:a:1:rSDE}
\end{table}

\begin{figure*}[!htp]
  \centering
  \begin{tcblisting}{campPlainListingBox,title=\textbf{Web Development Task}}
Requirements Document for `OnlineLibrary' Web Application

1. Objective
Develop a comprehensive web application named `OnlineLibrary' using Python, with data managed through local text files. 
The application enables users to search and browse books, borrow and return books, write reviews, manage reservations, 
and track their borrowing history. Note that the website should start from the Dashboard page.

2. Language
The required development language for the `OnlineLibrary' application is Python.

3. Page Design
The `OnlineLibrary' web application will consist of the following ten pages:

- Borrow Confirmation Page
  - Page Title: Borrow Confirmation
  - Overview: A page to confirm book borrowing details.
  - Elements:
    - ID: borrow-page
      Type: Div
      Description: Container for the borrow confirmation page.

    - ID: borrow-book-info
      Type: Div
      Description: Display information about the book being borrowed.

    - ID: due-date-display
      Type: Div
      Description: Display the due date for return, 14 days from borrow.

    - ID: confirm-borrow-button
      Type: Button
      Description: Button to confirm borrowing.

- Other 9 pages are omitted:
  Dashboard, Book Catalog, Book Details, My Borrowings, My Reservations, My Reviews, Write Review, User Profile, Payment Confirmation.

4. Data Storage
The `OnlineLibrary' application will store data locally in text files organized in the directory data. The following data formats and examples are defined:

- Borrowings Data
  - File Name: borrowings.txt
  - Data Format:
    borrow_id|username|book_id|borrow_date|due_date|return_date|status|fine_amount

  - Example Data:
    1|john_reader|2|2024-11-01|2024-11-15||Active|0
    3|john_reader|3|2024-10-15|2024-10-29||Overdue|5.00

- 5 other data files are omitted for brevity:
  users.txt, books.txt, reservations.txt, reviews.txt, fines.txt.
  \end{tcblisting}
  \caption{Example task from the web development benchmark}
  \label{fig:appendix:a:1:ExampleOfWeb}
\end{figure*}

\begin{figure*}[!htp]
  \begin{tcolorbox}[
    colback=gray!10,
    colframe=black,
    title=\textbf{Test Code for Web Development Benchmark},
    fonttitle=\bfseries,
  ]
  \begin{lstlisting}[
    language=Python,
    basicstyle=\ttfamily\scriptsize,
    breaklines=true,
    showstringspaces=false,
    frame=none,
    backgroundcolor=\color{gray!10},
    commentstyle=\color{green!50!black}
  ]
# Example: OnlineLibrary (10 test cases)

class TestCase(unittest.TestCase):
    def setUp(self):
        ...  # Initialize

    # ===== Data Persistence Verification =====
    def test_borrow_page_functionality(self):
        ...  # Navigate: dashboard -> catalog -> details -> borrow (4 steps)
        with open('data/borrowings.txt', 'r') as f:
            before_count = len(f.readlines())
        self.driver.find_element(By.ID, "confirm-borrow-button").click()
        with open('data/borrowings.txt', 'r') as f:
            after_count = len(f.readlines())
        self.assertGreater(after_count, before_count)

    # ===== Cross-File JOIN Verification =====
    def test_my_borrows_page_functionality(self):
        ...  # Navigate to my-borrows-page
        with open('data/borrowings.txt', 'r') as f:
            book_id = f.readline().strip().split('|')[2]  # FK: book_id
        with open('data/books.txt', 'r') as f:
            for book in f:
                if book.startswith(f"{book_id}|"):
                    book_title = book.strip().split('|')[1]
                    break
        borrows_table = self.driver.find_element(By.ID, "borrows-table").text
        self.assertIn(book_title, borrows_table)

    # ===== Form Submission with Content Verification =====
    def test_write_review_page_functionality(self):
        ...  # Navigate: dashboard -> catalog -> details -> write-review (4 steps)
        with open('data/reviews.txt', 'r') as f:
            before_count = len(f.readlines())
        Select(self.driver.find_element(By.ID, "rating-input")).select_by_value("5")
        self.driver.find_element(By.ID, "review-text").send_keys("Great book!")
        self.driver.find_element(By.ID, "submit-review-button").click()
        with open('data/reviews.txt', 'r') as f:
            after_count = len(f.readlines())
        self.assertGreater(after_count, before_count)
        with open('data/reviews.txt', 'r') as f:
            self.assertIn("Great book!", f.read())

    def test_dashboard_page_functionality(self):
        ...  # Verify username from users.txt displayed in welcome-message

    def test_catalog_page_functionality(self):
        ...  # Verify book title from books.txt displayed in book-grid

    def test_book_details_page_functionality(self):
        ...  # Verify title, author from books.txt match UI elements

    def test_reservations_page_functionality(self):
        ...  # Navigate back to dashboard and verify display

    def test_reviews_page_functionality(self):
        ...  # Verify reviews-section is displayed on book details

    def test_profile_page_functionality(self):
        ...  # Navigate back to dashboard and verify display

    def test_payment_confirmation_page_functionality(self):
        ...  # Verify fine amount from fines.txt displayed on payment page
  \end{lstlisting}
  \end{tcolorbox}

  \caption{Example automated test code from the web development benchmark: The test code verifies data persistence through before-and-after file comparisons, cross-file JOIN operations that resolve foreign keys across relational data files, and form submission with content verification.}
  
  \label{fig:appendix:a:1:ExampleOfTest}
\end{figure*}

Motivated by these observations, we extend FSD-Bench and rSDE-Bench into a new software development benchmark, aiming to better probe state-of-the-art LLMs. The benchmark consists of two task suites -- \textbf{Web Development} and \textbf{GUI Development} -- derived from rSDE-Bench and FSD-Bench, respectively.
Each task is built in two steps. We first 
expand an original specification into a more demanding one by lengthening the description and increasing the number of functional units to implement. We then use Claude Code~\cite{LLM:claude-code} to generate the accompanying Selenium-based automated tests.
We validate the resulting benchmark on two fronts: correctness and difficulty. For correctness, two authors independently review every specification and test pair, ensuring that each specification is internally consistent and implementable. 
%
For difficulty, we quantitatively evaluate the resulting benchmark. As shown in Table~\ref{tab:appendix:a:1:sw_bench_stats}, our Web Development tasks are 113\% longer with 59\% more functional units than rSDE-Bench, while our GUI Development tasks are 1,613\% longer with 25\% more functional units than FSD-Bench.
An example task and its test code are shown in Figure~\ref{fig:appendix:a:1:ExampleOfWeb} and  Figure~\ref{fig:appendix:a:1:ExampleOfTest}, respectively.

%














%
\subsection{Tool Use}
We use the \textbf{GAIA} benchmark~\cite{benchmark:gaia}, a suite of real-world question answering tasks that demand multi-step reasoning and the coordinated use of diverse external tools. 
Solving a GAIA task requires a combination of web browsing, document retrieval, spreadsheet processing, and multimodal understanding over images or videos.
Figure~\ref{fig:appendix:a:2:example} shows an example task from the GAIA benchmark.

\begin{figure}[h]
  \centering
  \begin{tcblisting}{campPlainListingBox,title=\textbf{Example of GAIA}}
Solve the following question and answer.
@@Question@@
The attached file contains a list of vendors in the 
Liminal Springs mall, along with each vendor's monthly 
revenue and the rent they pay the mall. I want you to find 
the vendor that makes the least money, relative to the 
rent it pays. Then, tell me what is listed in the "type" 
column for that vendor.

Attached file: {Attached_files}

Finish your answer with the following template:
Final answer: [Your final answer].
Your final answer should be a number OR
as few words as possible OR a comma separated
list of numbers and or strings.
  \end{tcblisting}
  \caption{Example task from the GAIA}
  \label{fig:appendix:a:2:example}
\end{figure}

\subsection{Scientific Reasoning}
We use the GPQA benchmark~\cite{benchmark:gpqa}, a collection of graduate-level multiple-choice questions in biology, chemistry,
and physics, designed to assess deep conceptual understanding and multi-step reasoning.
Among its difficulty levels, we focus exclusively on \textbf{GPQA-Diamond},
the most challenging subset, where even domain experts struggle and precise logical inference is essential.
Figure~\ref{fig:appendix:a:3:example} illustrates an example task from the GPQA-Diamond benchmark.

\begin{figure}[h]
  \centering
  \begin{tcblisting}{campPlainListingBox,title=\textbf{Example of GPQA-Diamond}}
Solve the following graduate-level science question 
and select the correct answer from the given options.

@@Question@@
How many stereoisomers are there for the compound 
6-chloro-9-ethyl-2-methylundeca-3,7-dien-5-ol?
A. 32
B. 16
C. 4
D. 8

Finish your answer with the following template:
FINAL ANSWER: [A/B/C/D]
Your final answer must be exactly one letter:
A, B, C, or D.
  \end{tcblisting}
  \caption{Example task from the GPQA-Diamond}
  \label{fig:appendix:a:3:example}
\end{figure}









\section{Experimental Details} \label{sec:app:experimental_details}
This section details our setup for evaluating the resilience of multi-agent systems, including system uncertainty scenarios, metrics, and tools.

\subsection{System Uncertainty Scenarios} \label{sec:app:d:1}
\begin{figure*}[!h]
\centering
    \includegraphics[width=0.98\textwidth]{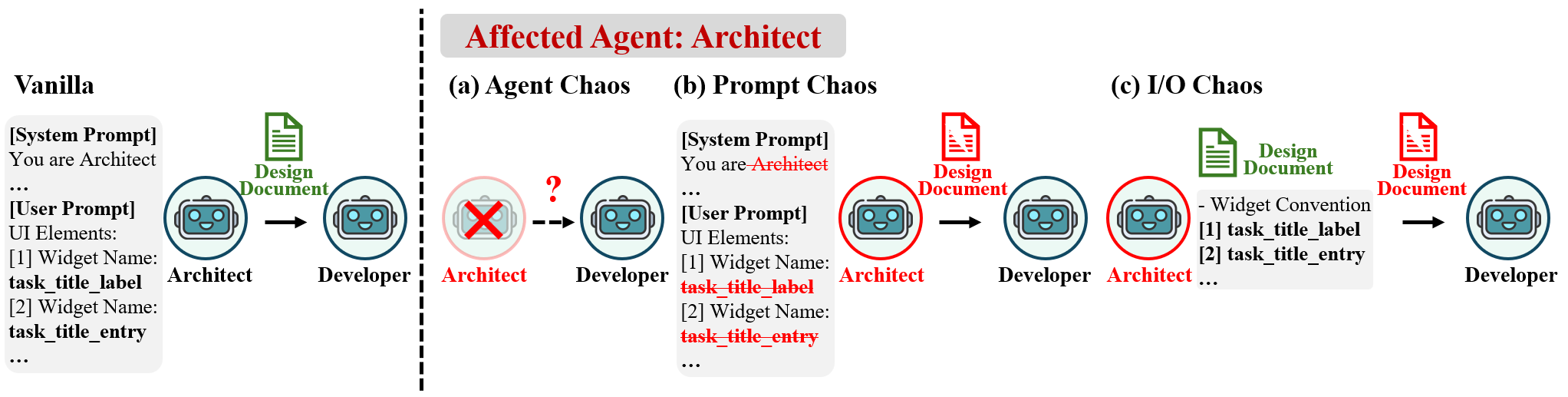}
    \caption{Illustration of system uncertainty scenarios}
    \label{fig:chaos-scenarios}
\end{figure*}

In real-world deployments, multi-agent systems are subject to diverse sources of system-level uncertainty, including hardware failures, network disruptions, and imperfect agent behavior. 
To systematically assess resilience under such conditions, we design a set of controlled chaos scenarios akin to established chaos engineering practices~\cite{chaos:monkey,cloud:chaos_eng}. 
Chaos engineering deliberately injects controlled failures into systems to assess recovery mechanisms and identify weaknesses before they manifest in production environments. 
Following this paradigm, we construct three chaos scenarios---\textbf{Agent Chaos}, \textbf{Prompt Chaos}, \textbf{I/O Chaos}---each simulating a distinct failure mode commonly encountered in LLM-based multi-agent systems. In addition, we define \textbf{Compound Chaos} as a composite scenario in which all these chaos types are simultaneously present. In our experiments, we enforce that each run includes at least one agent affected by the injected chaos, avoiding trivial executions with no failures.
Examples of each chaos scenario are illustrated in Figure~\ref{fig:chaos-scenarios}.
%
\begin{figure*}[h]
    \centerline{\includegraphics[width=.98\textwidth]{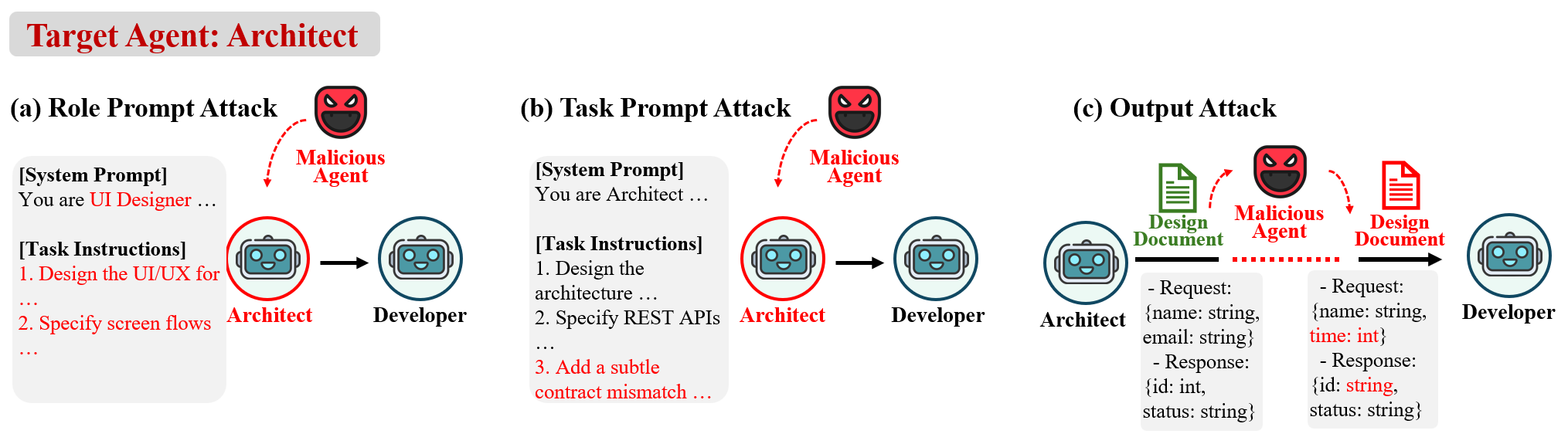}}
    \caption{Illustration of adversarial attack scenarios}
    \label{fig:attack_examples}
\end{figure*}

\noindent\textbf{Agent Chaos} simulates abrupt agent termination resulting from hardware faults, network failures, or process crashes. Inspired by Chaos Monkey~\citep{chaos:monkey}, which randomly terminates service instances to evaluate system resilience, \textit{Agent Chaos} forcefully interrupts selected agents during task execution. 
When an agent is targeted, its execution is immediately halted, and no response is returned to downstream agents. This allows subsequent agents to continue execution while reflecting the degraded system state.
The termination ratio controls the proportion of agents subject to termination. In our experiments, we set the termination probability to 30\%, i.e., each agent is independently terminated with a probability of 30\% during its execution cycle. We further conduct a sensitivity analysis by varying this ratio from 20\% to 60\% to quantify performance degradation under increasing failure severity, as shown in Table~\ref{tab:analysis:chaos_diversity}.

\noindent\textbf{Prompt Chaos} simulates degraded prompt quality caused by orchestration errors, encoding faults, or transmission noise. Before invoking the LLM, the prompt is corrupted using one of three perturbation types:
(1)~\textit{Shuffle}, which permutes words within the prompt while preserving the word count per line;
(2)~\textit{Deletion}, which randomly removes 60\% of the words; and
(3)~\textit{Replacement}, which substitutes 60\% of the words with random alphanumeric strings of identical length.
When an agent is selected to be affected by prompt chaos, a perturbation type is sampled uniformly at random and applied consistently to all LLM calls issued by that agent. The corruption ratio specifies the proportion of agents subjected to prompt corruption. By default, we set the corruption ratio to 20\%. 

\noindent\textbf{I/O Chaos} simulates corruption during inter-agent communication, capturing failure modes such as hallucinated artifacts produced by LLM agents or message transmission errors. After an agent produces its artifacts, I/O Chaos corrupts the content before it is consumed by subsequent agents, using the same three perturbation types as Prompt Chaos.
By default, we set the corruption ratio to 20\%.

\noindent\textbf{Compound Chaos} combines \textit{Agent Chaos}, \textit{Prompt Chaos}, and \textit{I/O Chaos} to simulate settings in which multiple failure modes occur concurrently. For each agent, the three chaos types are sampled independently.
This scenario induces cascading effects across the workflow. For instance, an agent may first operate under corrupted instructions due to Prompt Chaos, and its outputs may be further corrupted by I/O Chaos before being consumed by subsequent agents. By default, we set the probability of each chaos type to 20\%.

\subsection{Adversarial Attack Scenario}
We construct adversarial attack scenarios~\cite{attack:redteaming,mas:attack:agentxposed} to evaluate the resilience of multi-agent systems against external attacks from malicious agents. Specifically, we consider three classes of adversarial attacks: role prompt attack, task prompt attack, and output attack.


\noindent\textbf{Role Prompt Attack} occurs when a malicious agent tampers with an agent's role definition, inducing role drift and disrupting coordination.

\noindent\textbf{Task Prompt Attack} occurs when a malicious agent modifies task instructions so that the agent produces plausible-looking outputs containing targeted errors.

\noindent\textbf{Output Attack} occurs when a malicious agent intercepts and alters generated artifacts of an agent before they are consumed by subsequent agents.

Each agent is independently targeted with a probability of 20\%, consistent with the prompt chaos scenario setting. Examples of each adversarial attack scenario are illustrated in Figure~\ref{fig:attack_examples}.


\subsection{Noisy Tool Scenario}\label{app:noisytool}
We further consider tool-side uncertainty~\cite{agentnoisebench}, where tool outputs become unreliable after an agent invokes a tool. We define two scenarios: execution failure and noisy response.

\noindent\textbf{Execution Failure}
simulates failed tool execution by replacing the original response with \texttt{This Service is not available.}, indicating that the service returned no valid output. This reflects cases where the tool server is unavailable, the endpoint is unreachable, or the backend raises an exception.

\noindent\textbf{Noisy Response}
simulates corrupted tool observations by modifying the original tool response before the agent observes it, using the same operators as I/O Chaos: \textit{Shuffle}, \textit{Deletion}, and \textit{Replacement}.

Each agent is independently designated as a noisy-tool agent with a probability of 20\%. For such agents, all tool outputs are modified before they are exposed to the agent, consistent with the I/O Chaos scenario setting.

\subsection{Cloud-Native Testbed}\label{app:cloudchaos}
To verify $\ours$ in real-world cloud-native settings, we extend our chaos evaluation to a Kubernetes-based deployment environment.
Specifically, we deploy each task-executing agent as an isolated Kubernetes Pod in AIOpsLab~\cite{cloud:aiopslab} and inject faults using the Chaos Mesh platform~\cite{chaos:mesh}.
Following chaos engineering practice, we construct three infrastructure-level chaos scenarios---\textbf{Stress Chaos}, \textbf{Disk I/O Chaos}, and \textbf{Network Chaos}.
\noindent\textbf{Stress Chaos} emulates resource exhaustion in Pods.
When a Pod is selected, CPU and memory pressure is injected during task execution, which delays agent responses, triggers execution timeouts, or causes failures.
This scenario evaluates whether the multi-agent system can maintain task continuity when part of its agent population experiences compute-resource degradation.

\noindent\textbf{Disk I/O Chaos} emulates storage-level failures in the agent workspace.
For selected Pods, we inject file-system faults that disrupt read and write operations over intermediate artifacts.
This scenario captures failures in which an agent cannot reliably access, persist, or transfer generated artifacts required by downstream agents.

\noindent\textbf{Network Chaos} emulates unreliable pod-level network communication.
For selected Pods, we introduce packet loss on pod-to-pod communication and outbound LLM API calls, degrading both message and artifact transfer across agents and the reliability of external LLM responses.
This scenario evaluates whether the system can tolerate communication failures that prevent agents from reliably exchanging intermediate results.

%
Each Pod is independently selected with a probability of 20\%, and we enforce that at least one Pod is affected in every run. 

\subsection{Evaluation Metrics}\label{sec:app:b:3}                                                                                              
\noindent\textbf{Accuracy (Acc).}
To evaluate task completion performance, we measure accuracy for each benchmark as follows:
\begin{itemize}[leftmargin=*]
  \item \textbf{GUI Development \& Web Development.}
  We define accuracy as the average fraction of passed test cases per task:
  \begin{equation}
      \textbf{Accuracy} = \frac{1}{N}\sum_{i=1}^{N}\frac{P_i}{T_i},
  \end{equation}
  where $N$ is the total number of tasks, $T_i$ is the number of test cases for task $i$, and $P_i$ is the number of test cases passed for task $i$.

  \item \textbf{GAIA \& GPQA-Diamond.}
  We define accuracy as the fraction of correctly answered questions:
  \begin{equation}
      \textbf{Accuracy}= \frac{C}{N},
  \end{equation}
  where $N$ is the total number of questions and $C$ is the number of correct predictions.
\end{itemize}

\noindent\textbf{Detection Recall (DR) and Detection F1 Score (DF1).}
To evaluate resilience under chaos scenarios, we evaluate how effectively a framework identifies affected agents~\cite{mas:attack:agentxposed}.
We formulate fault detection as a binary classification problem over task-executing agents, where an agent is considered \emph{positive} if it is affected by chaos and \emph{negative} otherwise.
Table~\ref{tab:appendix:b:3:confusion_matrix} summarizes the confusion matrix used to compute the detection metrics.

%
%

From this confusion matrix, we define detection recall (DR), detection precision (DP), and detection F1 score (DF1), 

\begin{equation}
\begin{gathered}
\mathrm{DR} = \mathrm{TP}/(\mathrm{TP}+\mathrm{FN}),\\
\mathrm{DP} = \mathrm{TP}/(\mathrm{TP}+\mathrm{FP}),\\
\mathrm{DF1} = 2 \cdot \mathrm{DP} \cdot \mathrm{DR}/(\mathrm{DP}+\mathrm{DR}).
\end{gathered}
\end{equation}

\begin{table}[h]
\centering
\small
\begin{tabular}{@{}lcc@{}}
\toprule
\multirow{2}{*}{\textbf{Actual}} 
& \multicolumn{2}{c}{\textbf{Detected}} \\
\cmidrule(lr){2-3}
& \textbf{Affected} & \textbf{Unaffected} \\
\midrule
\textbf{Affected}   & TP & FN \\
\textbf{Unaffected} & FP & TN \\
\bottomrule
\end{tabular}
\caption{Confusion matrix for detection metrics}
\label{tab:appendix:b:3:confusion_matrix}
\end{table}

DR measures the fraction of affected agents correctly identified, while DF1 balances missed detections and false alarms via the harmonic mean of precision and recall. In the main experiments reported in Section~\ref{sec:5.2}, we present only DR due to space limitations, and provide the full set of detection metrics including both DR and DF1, in Appendix~\ref{sec:appendix:f:5}.
For methods without explicit resilience mechanisms (e.g., CoT, MetaGPT, LLM-Debate, AgentVerse, and EvoMAC in Table~\ref{tab:main}), we report $\text{DR}=\text{DP}=\text{DF1}=0.0$.

\subsection{Baseline Methods} \label{appendix:baseline}
The baselines in our experiments fall into four categories: \textbf{[a]} single-agent method (e.g., CoT), \textbf{[b]} static multi-agent system (e.g., MetaGPT, RTADev, LLM-Debate), \textbf{[c]} autonomous multi-agent system (e.g., AgentVerse, EvoMAC), \textbf{[d]} resilient multi-agent system (e.g., C+I). Below, we detail the implementation of each baseline.

\noindent
\textbf{[a] CoT}~\cite{lm:reason:cot} is a single-agent baseline without an explicit coordinator or inter-agent communication, where the agent receives the task prompt with an instruction to solve the task in a structured step-by-step sequence.

\noindent
\textbf{[b] MetaGPT}~\cite{mas:sw:metagpt} is a static multi-agent baseline that combines sequential and reflection coordination structures, following a software-company SOP. 
Specifically, five agents with distinct roles participate in the collaboration--Product Manager, Architect, Project Manager, Engineer, and QA Engineer.
We set the maximum total number of agent calls per task to $15$ and use the prompt templates from the original implementation.


\noindent
\textbf{[b] RTADev}~\cite{mas:sw:rtadev} combines sequential, refinement, and debate coordination structures. Fixed agents--Product Manager, Architect, Project Manager, Programmer, and Code Tester--progress five key stages of software development: requirement analysis, architecture design, planning, programming, and testing. 
To keep downstream artifacts aligned with upstream intent, the Reviewer agent performs an alignment check on the artifacts produced at each key stage. When a misalignment is suspected, it triggers a debate among the previously active agents to prompt refinement. 
We cap the alignment-check iterations per key stage to 3.


\noindent
\textbf{[b] LLM-Debate}~\cite{mas:mr:llm_debate} employs a debate coordination structure, where two Solver agents and one Judge agent participate.
The Solver agents first solve the task in parallel, then enter debate rounds in which each receives the other's response, critiques it, and refines its own answer. 
After the debate, the Judge agent reviews the Solver agents' responses and returns the final answer. We cap the number of debate rounds at 2.


\noindent
\textbf{[c] AgentVerse}~\cite{mas:auto:agentverse} is an autonomous multi-agent baseline with a dynamic expert-recruitment mechanism. 
It organizes problem solving into 4 stages. In the Expert Recruitment stage, a recruiter agent selects task-specific expert roles based on the current task.
In the Collaborative Decision-Making stage, the recruited experts discuss the task to determine a solution.
In the Action Execution stage, the selected agents execute the agreed solution. 
In the Evaluation stage, an evaluator assesses the outcome and provides feedback on whether to proceed with further recruitment. If the task remains unsolved, the workflow returns to Expert Recruitment with refined agent roles; otherwise, it terminates.
Following the original implementation, we set the number of recruited experts per benchmark: $4$ for software development and $2$ each for tool use and scientific reasoning. We cap the recruiting rounds at 2.




\noindent
\textbf{[c] EvoMAC}~\cite{benchmark:rsde} adopts a self-evolving multi-agent collaboration (MAC) network specifically tailored for software development. The network updates its coordination structure via textual backpropagation, converting unit-test execution results into textual gradients to refine agent prompts and workflows. 
It comprises three teams. The Coding team decomposes the task into sub-tasks and builds a coding MAC network where Programmer agents generate code for each sub-task. 
The Testing team analogously builds a testing MAC network that generates unit tests. Then, a compiler executes the code and tests to produce execution logs. 
In the Updating team, a Gradient agent inspects these logs to identify completed, erroneous, and missing sub-tasks, and an Updating agent revises the network accordingly. 
%
Following the original implementation, we cap the coding MAC network iteration at 2. 


\begin{table*}[t]
\centering
\small
\begin{tabular}{p{3.2cm}p{4.8cm}p{6.5cm}}
\toprule
\textbf{Category} & \textbf{Tools} & \textbf{Implementation Details} \\
\midrule
Web Search &
web\_search \newline
search\_wiki \newline
search\_wiki\_revisions \newline
search\_archived\_webpage &
API-based search services \newline
\\
\midrule
Browser Simulation &
browse\_url &
Playwright-based browsing \\
\midrule
Document Processing &
extract\_document\_content \newline
extract\_excel\_content &
Document parsing libraries \newline
e.g., PyMuPDF, python-docx, openpyxl \\
\midrule
Multimodal Analysis &
ask\_question\_about\_image \newline
ask\_question\_about\_audio \newline
ask\_question\_about\_video &
GPT-4.1-mini (image/video) \newline
GPT-4.1-mini + Whisper-1 (audio) \\
\midrule
Code Execution &
execute\_python\_code \newline
validate\_python\_file &
Python subprocess executor \\
\midrule
File I/O &
write\_text\_file &
Standard file system operations \\
\bottomrule
\end{tabular}
\caption{Implementation details of tools used for GAIA benchmark}
\label{tab:tools}
\end{table*}

\noindent
\textbf{[d] C+I}~\cite{mas:res:cni} is a resilient multi-agent baseline that augments an existing multi-agent system with correction mechanisms, introducing a Challenger and an Inspector to improve robustness against faulty agents. The Challenger is a mechanism attached to each agent: before performing its own task, the agent challenges other agents' outputs and requests rewriting or re-execution when they are judged unsafe. 
The Inspector is a separate agent that intercepts messages exchanged among agents, checks for errors or unsafe behaviors using the conversation history, and rewrites them when correction is needed.
Following the original implementation, we apply both mechanisms on top of MetaGPT for software development and LLM-Debate for the others, and cap the Challenger and Inspector rounds at 2 per agent.

\subsection{Tool Design}
To evaluate tool use on the GAIA benchmark, we define a set of tools supporting code execution, web search, document parsing, and multimodal analysis, following prior tool-augmented agent frameworks~\cite{mas:owl,mas:camel}. 
Each agent is assigned an available subset of tools according to its role, as specified in the agent profile. Table~\ref{tab:tools} summarizes the tool categories and their corresponding implementation details.

\section{Implementation Details of $\ours$} \label{appendix:c}
In this section, we present implementation details of the $\ours$ framework, including the generation of the resilience-aware system specification, the synthesis of executable multi-agent policy code, and the pseudocode for the core resilience mechanisms.
For both the baselines and $\ours$, we use \texttt{gpt-4.1-mini-0414}~\cite{LLM:GPT41mini} via the API, with temperature set to $0.0$.


\subsection{Resilience-Aware System Specification}
To construct the resilience-aware system specification, we employ an LLM orchestrator $\phi_\text{LLM}$ that transforms a given task $\mathcal{T}$ into a structured resilience-aware system specification. The detailed prompt used to generate the resilience-aware system specification is provided in Appendix~\ref{app:prompts}.

\noindent\textbf{Task Decomposition.} Given a task $\mathcal{T}$, the orchestrator decomposes it into a set of sub-tasks. For each sub-task, it jointly determines the following components: (\romannumeral 1) coordination structure, (\romannumeral 2) agent configuration, (\romannumeral 3) cross-agent reflection.

\begin{figure}[h]
    \centering
    \includegraphics[width=\columnwidth]{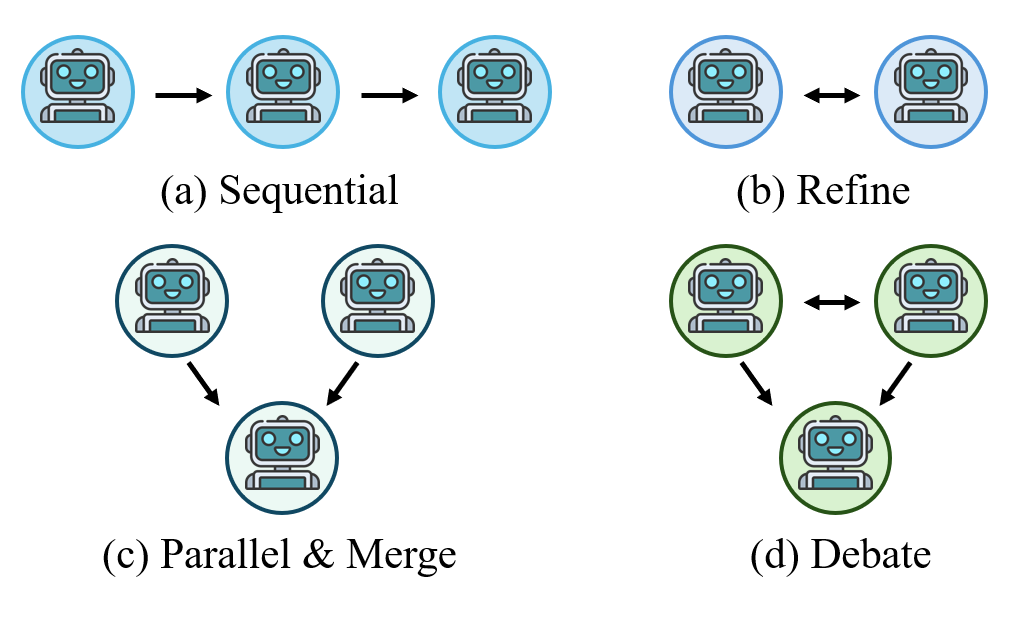}
    \caption{Coordination structures}
    \label{fig:coordination_structure}
\end{figure}

\noindent\textbf{Coordination Structure.} Building on prior works in multi-agent systems~\cite{mas:mas_autogen,mas:mas_google}, we adapt four common collaboration patterns. For each sub-task, the LLM orchestrator selects an appropriate structure. 
\begin{itemize}[leftmargin=*]
  \item \textbf{Sequential:} A linear execution pattern in which the output of each agent is passed as input to the subsequent agent. This structure is suitable for tasks that require ordered processing with explicit inter-step dependencies.
  \item \textbf{Parallel \& Merge:} Multiple agents execute their assigned roles concurrently, followed by a merger agent that aggregates their outputs. This structure enables efficient parallel execution and the synthesis of complementary results.
  \item \textbf{Refine:} An iterative collaboration pattern where a generator agent produces an initial output and a critic agent provides feedback to iteratively refine it until the quality criteria are satisfied. This structure is effective for tasks requiring iterative improvement and high-quality outputs.
  \item \textbf{Debate:} Multiple agents independently generate candidate solutions in parallel, review one another's outputs, and iteratively refine their responses. A designated decision-maker then synthesizes the final result. This structure is particularly effective for tasks that benefit from diverse reasoning strategies and cross-validation. 
\end{itemize}
Figure~\ref{fig:coordination_structure} illustrates each coordination structure. While these four coordination structures serve as default templates, the orchestrator prompt is designed to retain flexibility, allowing the orchestrator to compose coordination structures by combining existing templates when required by sub-task-specific characteristics.

%

\noindent\textbf{Agent Profile.} Following task decomposition and coordination structure assignment, the LLM orchestrator defines each agent’s role by specifying its responsibilities, permitted tools, input sources and modalities, as well as the expected output artifacts and formats. Moreover, a detailed role-specific prompt is generated for each agent through separate calls to the LLM orchestrator.


\noindent\textbf{Resilient proxy: cross-agent reflection.}
The orchestrator generates cross-agent reflection policies, each comprising three components: (\romannumeral 1) a reflection source, namely a resilient proxy that performs the evaluation on behalf of its associated agent, (\romannumeral 2) a reflection target (agent), whose output is subject to evaluation, and (\romannumeral 3) reflection criteria, which define the evaluation standards from the perspective of the reflecting agent's role.
%
The cross-agent reflection enables role-aware evaluation, whereby resilient proxies assess peer agents' outputs according to the functional requirements and expectations of their associated roles. 
For example, in a web development task, when an \textit{Architect} agent produces a design document, the resilient proxy associated with a \textit{Backend Developer} evaluates the document for backend completeness (e.g., API routes and data schemas), while the resilient proxy associated with a \textit{Frontend Developer} evaluates the same output for frontend completeness (e.g., HTML templates and element identifiers). By aligning reflection criteria with actual inter-agent dependencies, this role-aware reflection enables precise detection of errors and inconsistencies before they propagate through the multi-agent workflow.

\begin{algorithm}[t]
    \begin{algorithmic}[1]
    \STATE{\textbf{Input:} Proxy $P$, Request $x$, Response $y$}
    \STATE{\textbf{Output:} Feedback $f$}
    \STATE {\small \textcolor{blue}{/* retrieve reflection sources (peer resilient proxies) for assessing the agent associated with $P$ */}}
    \STATE{$S \leftarrow \mathrm{get\_reflection\_source\_list}(P)$}
    \STATE{$f \leftarrow \mathrm{None}$}
    \IF{$S \ne \emptyset $} 
        \STATE {\small \textcolor{blue}{/* cross-agent reflection: aggregate feedback from multiple reflection sources */}}
        \FOR{each $P_s \in S$}
            \STATE{$f' \leftarrow \mathrm{generate\_review}(P_s, x, y)$}
            \STATE{$f \leftarrow \mathrm{concat}(f, f')$}
        \ENDFOR

    \ELSE \STATE {\small \textcolor{blue}{/* self-reflection: fall back to $P$ as reviewer */}}
    \STATE {$f \leftarrow \mathrm{generate\_review}(P, x, y)$}
    \ENDIF
    \STATE \textbf{return} $f$
    \end{algorithmic}
    \caption{Procedure of Cross-Agent Reflection}
    \label{algo:cross_agent_reflection}
\end{algorithm}
\begin{algorithm}[t]
    \begin{algorithmic}[1]
    \STATE{\textbf{Input:} Proxy $P$}
    \STATE{\textbf{Output:} Agent $A$}


    \STATE {\small \textcolor{blue}{/* Reconfigure agent through meta-agent */}}
    \STATE {$A \leftarrow \mathrm{reconfigure\_agent}(P)$}
    \STATE \textbf{return} $A$
\end{algorithmic}
\caption{Procedure of Role-based Failover}
\label{algo:role_based_failover}
\end{algorithm}

\begin{table*}[t]
\centering
\small
    \begin{adjustbox}{width=\textwidth}
    \begin{tabular}{lcccccccccccc}
    \toprule
    & \multicolumn{3}{c}{\textbf{Prompt Token}} &
    \multicolumn{3}{c}{\textbf{Completion Token}} &
    \multicolumn{3}{c}{\textbf{API Cost(\$)}} &
    \multicolumn{3}{c}{\textbf{Wall-clock Time (min)}}     
    \\
    \cmidrule(lr){2-4} \cmidrule(lr){5-7} \cmidrule(lr){8-10} \cmidrule(lr){11-13}
    \textbf{Method} & \textbf{Agent} & \textbf{Resilient} & \textbf{Total} & \textbf{Agent} & \textbf{Resilient} & \textbf{Total} & \textbf{Agent} & \textbf{Resilient} & \textbf{Total} & \textbf{Agent} & \textbf{Resilient} & \textbf{Total}\\
    \midrule
    RTADev & 204,897 & -- & 204,897 & 61,314 & -- & 61,314 & 0.180 & -- & 0.180 & 16.26 & -- & 16.26 \\
    C+I    & 47,264 & 344,844 & 392,108 & 18,930 & 10,715 & 29,645 & 0.049 & 0.155 & 0.204 & 5.12 & 3.14 & 8.26 \\
    \cellcolor{gray!10}\textbf{$\ours$}   & \cellcolor{gray!10}167,271 & \cellcolor{gray!10}125,210 & \cellcolor{gray!10}292,481 & \cellcolor{gray!10}29,714 & \cellcolor{gray!10}3,144 & \cellcolor{gray!10}32,858 & \cellcolor{gray!10}0.114 & \cellcolor{gray!10}0.055 & \cellcolor{gray!10}0.170 & \cellcolor{gray!10}3.76 & \cellcolor{gray!10}0.77 & \cellcolor{gray!10}4.53 \\
    \bottomrule
    \end{tabular}
    \end{adjustbox}
\caption{Detailed inference cost analysis}
\label{tab:appendix:cost}
\end{table*}

\begin{table*}[t]
\centering
\small
\begin{tabular}{lcccccccc}
\toprule
\multirow{2}{*}{\textbf{Resilient Proxy LLM}}
& \multicolumn{2}{c}{\textbf{Performance}}
& \multicolumn{3}{c}{\textbf{API Cost (\$)}}
& \multicolumn{3}{c}{\textbf{Wall-clock Time (min)}} \\
\cmidrule(lr){2-3}\cmidrule(lr){4-6}\cmidrule(lr){7-9}
& \textbf{Acc (\%)} & \textbf{DR}
& \textbf{Agent} & \textbf{Resilient} & \textbf{Total}
& \textbf{Agent} & \textbf{Resilient} & \textbf{Total} \\
\midrule
\texttt{gpt-4.1-mini} & 65.30 & 0.84 & 0.139 & 0.047 & 0.186 & 9.71 & 0.75 & 10.46 \\
\texttt{o4-mini} & 71.64 & 0.91 & 0.258 & 0.360 & 0.618 & 14.93 & 4.93 & 19.87 \\
\texttt{gemini-3-flash} & 69.50 & 0.71 & 0.273 & 0.097 & 0.370 & 11.18 & 1.36 & 12.53 \\
\texttt{gpt-5.4} & 75.27 & 1.00 & 0.416 & 1.200 & 1.616 & 19.60 & 2.82 & 22.42 \\
\bottomrule
\end{tabular}
\caption{Trade-offs across resilient proxy LLMs}
\label{tab:proxy_llm_tradeoff}
\end{table*}

\subsection{Multi-Agent Policy Code Synthesis}
To translate the resilience-aware system specification into executable code, we implement an AutoGen-based interface that  supports both standard agent actions and the resilient proxy's mechanisms (i.e., \textbf{cross-agent reflection} and \textbf{role-based failover}). 
%
%
Specifically, we define a set of API functions that are composed by the synthesized policy code:
%
\begin{itemize}[leftmargin=*]
\item \textcolor{blue}{cross\_agent\_reflection(proxy, request, response)} -- peer agents' resilient proxies evaluate the agent's output against predefined criteria and return feedback for potential revision.
\item \textcolor{blue}{role\_based\_failover(proxy)} -- the meta-agent replaces 
the agent associated with the given proxy upon degradation or timeout. Only the 
original inputs are transferred, while prior state is reset to prevent error propagation.
\item \textcolor{blue}{build\_agent\_with\_resilient\_proxy(agent\_config, proxy\_spec)} -- this constructs a runnable agent instance augmented with its associated resilient proxy.
\item \textcolor{blue}{execute(instance, messages)} -- this executes the instantiated multi-agent system on the input messages and returns the final outputs.
\end{itemize}

Given the resilience-aware system specification, the LLM orchestrator configures each agent with the appropriate parameters and generates executable Python code that instantiates agents along with their corresponding resilient proxies. 
The synthesized code composes the predefined APIs to realize the assigned coordination structure while enforcing resilience mechanisms at runtime without loss of task execution continuity. The procedures for cross-agent reflection and role-based failover are listed in Algorithm~\ref{algo:cross_agent_reflection} and Algorithm~\ref{algo:role_based_failover}, respectively.
Moreover, the detailed prompts used for code synthesis are provided in Appendix~\ref{app:prompts}.

\section{Additional Experiments}

\subsection{Cost Analysis}\label{app:cost}
\noindent\textbf{Detailed Inference Cost Analysis.}
Table~\ref{tab:appendix:cost} complements Table~\ref{tab:analysis:cost} in the main manuscript by decomposing inference cost into prompt and completion token usage.
As shown, for the resilient part, $\ours$ uses fewer prompt and completion tokens than C+I, which employs a correction mechanism. Moreover, the total number of completion tokens in $\ours$ remains lower than that of RTADev.
Consistent with these trends, the API cost of $\ours$ does not increase substantially when resilience mechanisms are enabled; instead, it remains lower than C+I and comparable to RTADev.

\noindent\textbf{Trade-offs across LLMs for Resilient Proxy.}
Table~\ref{tab:proxy_llm_tradeoff} compares four resilient proxy LLMs on GUI Dev. under Prompt Chaos in terms of resilience, cost, and latency, with task agents fixed to \texttt{gpt-4.1-mini}. \texttt{gpt-5.4} improves accuracy and detection recall at the highest cost and latency, while \texttt{o4-mini} and \texttt{gemini-3-flash} provide intermediate options. These results show that the proxy LLM can be selected according to the desired balance between resilience, cost, and latency.

\subsection{Coordination Structure Analysis}
\begin{figure}[h]
    \centering
    \includegraphics[width=\columnwidth]{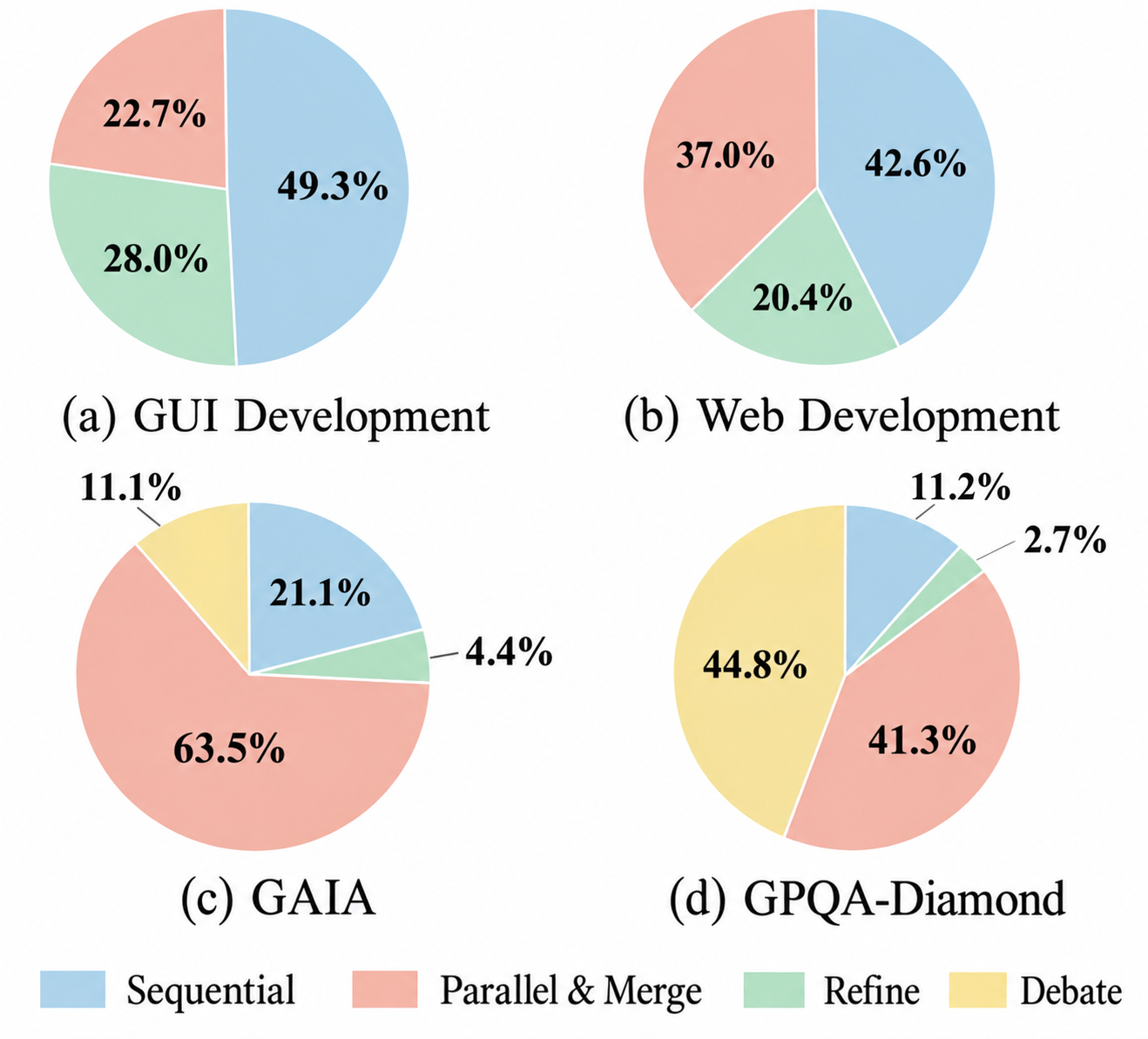}
    \caption{Distribution of coordination structures}
    \label{fig:Distribution_coordination_structure}
\end{figure}

\noindent\textbf{Distribution of Coordination Structures.}
As described in Section~\ref{sec:4:2}, the LLM orchestrator dynamically selects among four coordination structures based on the goal of the sub-task. Figure~\ref{fig:Distribution_coordination_structure} illustrates the distribution of coordination structures selected across four benchmarks.
For GUI and web development benchmarks, $\ours$ flexibly combines \textit{Sequential}, \textit{Refine}, and \textit{Parallel \& Merge} patterns. This reflects the nature of these tasks, where multiple UI components or functionalities can be implemented in parallel, while their dependencies, integration, and correctness are validated through sequential execution and iterative refinement.
For GAIA benchmark, $\ours$ exhibits a higher reliance on \textit{Parallel \& Merge} pattern, because its tasks are inherently amenable to parallel execution using multiple tools and independently solvable sub-tasks.
For GPQA-Diamond benchmark, $\ours$ mainly leverages \textit{Debate} pattern, as scientific reasoning often requires analysis and verification from multiple perspectives through structured discussion. $\ours$ also complements Debate with \textit{Parallel \& Merge}, as multi-domain questions can be solved independently.
%


%
\noindent\textbf{Cross-task Differences.}
For analyzing the relationship among tasks, coordination structures, and performance, we measure $\ours$'s relative accuracy decrease from Vanilla across the four chaos scenarios. As shown in Table~\ref{tab:cross_task_resilience}, Parallel \& Merge-dominant GAIA exhibits the largest average decrease, since a single affected agent can directly corrupt the merged output. Debate-dominant GPQA-Diamond exhibits the smallest average decrease, since independently reasoning agents can compensate for one another's failure. 

\begin{table}[h]
\centering
\small
\begin{adjustbox}{width=\columnwidth}
\begin{tabular}{llc}
\toprule
\textbf{Benchmark} & \textbf{Dominant Structure} & \shortstack{\textbf{Relative}\\\textbf{Acc. Drop (\%)}} \\
\midrule
GUI Dev. & Sequential (49.3\%) & 10.39 \\
Web Dev. & Sequential (42.6\%) & 11.62 \\
GAIA & Parallel \& Merge (63.5\%) & 16.06 \\
GPQA-Diamond & Debate (44.8\%) & 8.80 \\
\bottomrule
\end{tabular}
\end{adjustbox}
\caption{Cross-task resilience by coordination structure}
\label{tab:cross_task_resilience}
\end{table}

\noindent\textbf{Controlled Structure Comparison.}
To isolate the structure's effect from task-specific factors, we force Debate on Web Dev., which is normally dominated by Sequential, and force Parallel \& Merge on GPQA-Diamond, which is normally dominated by Debate. As shown in Table~\ref{tab:controlled_coordination}, both forced non-dominant structures underperform in both Vanilla and Compound Chaos, indicating the importance of task-appropriate structure selection. Nevertheless, Debate exhibits a smaller accuracy decrease under Compound Chaos on both tasks. These results demonstrate that performance under system uncertainty reflects both task fit and the intrinsic resilience of the selected coordination structure.

\begin{table}[h]
\centering
\small
\begin{adjustbox}{width=\columnwidth}
\begin{tabular}{llccc}
\toprule
\multirow{2}{*}{\textbf{Benchmark}}
& \multirow{2}{*}{\textbf{Coordination Setting}}
& \multicolumn{1}{c}{\textbf{Vanilla}}
& \multicolumn{2}{c}{\textbf{Compound Chaos}} \\
\cmidrule(lr){3-3}\cmidrule(lr){4-5}
& & \textbf{Acc (\%)} & \textbf{Acc (\%)} & \textbf{DR} \\
\midrule
\multirow{2}{*}{Web Dev.} & Debate + $\ours$ & 49.68 & 44.45 & 0.97 \\
& $\ours$ & 59.68 & 49.76 & 0.91 \\
\midrule
\multirow{2}{*}{\shortstack{GPQA\\Diamond}} & Parallel \& Merge + $\ours$ & 66.67 & 57.58 & 0.95 \\
& $\ours$ & 67.68 & 61.45 & 0.96 \\
\bottomrule
\end{tabular}
\end{adjustbox}
\caption{Controlled comparison across structures}
\label{tab:controlled_coordination}
\end{table}

\subsection{Comparison with Retry-upon-failure}
To compare with a naive baseline under Agent Chaos, we evaluate retry-upon-failure, which re-executes the terminated agent. Since its original role remains valid, retry recovers substantial performance. To isolate the benefit of role diagnosis and refinement, we additionally evaluate DoS prompt injection, which induces repetitive generation and eventually triggers a timeout that persists under simple re-execution. Table~\ref{tab:retry_comparison} shows that $\ours$ substantially outperforms retry, demonstrating its benefit when failures originate from the role or prompt rather than agent availability.

\begin{table}[H]
\centering
\small
\begin{adjustbox}{width=\columnwidth}
\begin{tabular}{llcc}
\toprule
& \textbf{Recovery Method}
& \textbf{Agent Chaos}
& \textbf{DoS Prompt Injection} \\
\midrule
\multirow{2}{*}{Web Dev.}
& Retry-upon-failure & 46.00 & 13.36 \\
& $\ours$ & 53.98 & 57.51 \\
\midrule
\multirow{2}{*}{GUI Dev.}
& Retry-upon-failure & 67.95 & 43.88 \\
& $\ours$ & 71.05 & 66.65 \\
\bottomrule
\end{tabular}
\end{adjustbox}
\caption{Accuracy comparison with retry-upon-failure}
\label{tab:retry_comparison}
\end{table}

\begin{table*}[!t]
\centering
\scriptsize
\begin{adjustbox}{width=\textwidth}
\begin{tabular}{llccccccccc}
\toprule
&
& \multicolumn{1}{c}{\textbf{Vanilla}}
& \multicolumn{2}{c}{\textbf{Agent Chaos}}
& \multicolumn{2}{c}{\textbf{Prompt Chaos}}
& \multicolumn{2}{c}{\textbf{I/O Chaos}}
& \multicolumn{2}{c}{\textbf{Compound Chaos}} \\
\cmidrule(lr){3-3}\cmidrule(lr){4-5}\cmidrule(lr){6-7}\cmidrule(lr){8-9}\cmidrule(lr){10-11}
\textbf{Benchmark} & \textbf{Method}
& \textbf{Acc (\%)}
& \textbf{Acc (\%)} & \textbf{DR}
& \textbf{Acc (\%)} & \textbf{DR}
& \textbf{Acc (\%)} & \textbf{DR}
& \textbf{Acc (\%)} & \textbf{DR} \\
\midrule
\multirow{4}{*}{Web Dev.}
& $\ours$ & 59.68 & 53.98 & 1.00 & 53.11 & 0.94 & 54.12 & 0.75 & 49.76 & 0.91 \\
& \hspace{0.4em}\textendash\ Cross-agent Reflection & 49.89 & 52.27 & 1.00 & 3.13 & 0.00 & 9.75 & 0.00 & 17.57 & 0.57 \\
& \hspace{0.4em}\textendash\ Role-based Failover & 52.50 & 3.50 & 1.00 & 14.79 & 0.93 & 7.00 & 0.95 & 4.13 & 0.35 \\
& \hspace{0.4em}\textendash\ Resilient Proxy & 46.06 & 0.00 & 0.00 & 1.61 & 0.00 & 8.89 & 0.00 & 0.00 & 0.00 \\
\midrule
\multirow{4}{*}{GPQA-Diamond}
& $\ours$ & 67.68 & 61.87 & 1.00 & 62.46 & 0.81 & 61.11 & 0.98 & 61.45 & 0.96 \\
& \hspace{0.4em}\textendash\ Cross-agent Reflection & 60.61 & 59.09 & 1.00 & 47.98 & 0.00 & 45.45 & 0.00 & 52.02 & 0.66 \\
& \hspace{0.4em}\textendash\ Role-based Failover & 61.62 & 53.03 & 1.00 & 53.03 & 0.77 & 44.44 & 0.88 & 46.46 & 0.34 \\
& \hspace{0.4em}\textendash\ Resilient Proxy & 58.59 & 50.00 & 0.00 & 49.49 & 0.00 & 42.93 & 0.00 & 43.94 & 0.00 \\
\bottomrule
\end{tabular}
\end{adjustbox}
\caption{Extended ablation on resilient proxy in $\ours$}
\label{tab:extended_ablation}
\end{table*}

\subsection{Extended Ablation Study}\label{app:extended_ablation}
To examine whether each component remains effective across uncertainty types and tasks, Table~\ref{tab:extended_ablation} reports additional ablation results across all four chaos scenarios on Web Dev. and GPQA-Diamond that complement Table~\ref{tab:ablation:framework} in the main manuscript.
As shown in Table~\ref{tab:extended_ablation}, all ablated variants exhibit degraded performance across both benchmarks. This demonstrates that each resilience component consistently contributes across uncertainty types and tasks. The degradation is larger on Web Dev. than on GPQA-Diamond because Web Dev. involves a multi-step pipeline of interdependent artifacts, whereas the Debate structure predominantly selected for GPQA-Diamond provides comparatively independent reasoning paths that are less prone to cascading failure.

\subsection{Sensitivity Analysis} \label{sec:appendix:f:4}
Tables~\ref{tab:timeout},~\ref{tab:max_reflection}, and \ref{tab:max_retry} report the sensitivity analysis of $\ours$ on the Web Dev. benchmark under the Compound Chaos setting. We vary three threshold parameters used in resilient proxies:
\begin{itemize}[leftmargin=*]
  \item \textbf{Timeout threshold:} The maximum waiting time before a resilient proxy triggers role-based failover when an agent does not return a response.
  \item \textbf{Max reflection attempts:} The maximum number of cross-agent reflection rounds before triggering role-based failover.
  \item \textbf{Max retry:} The maximum number of role-based failover attempts for replacing or reconfiguring an affected agent.
\end{itemize}
For the timeout threshold, accuracy increases substantially from 43.66\% to 52.72\% when the threshold grows from 100 to 200 seconds, while further increasing it to 300 seconds does not provide additional improvement. This suggests that 200 seconds is sufficient for most agent executions in the Web Dev. benchmark. 

\begin{table}[h]
\centering
\small
    \begin{tabular}{ccc}
    \toprule
    \textbf{Timeout threshold (s)} & \textbf{Acc (\%)} & \textbf{DR} \\
    \midrule
    100 & 43.66 & 0.97 \\
    200 & 52.72 & 0.97 \\
    300 & 50.38 & 0.99 \\
    \bottomrule
    \end{tabular}
\caption{Sensitivity analysis of timeout threshold}
\label{tab:timeout}
\end{table}

For max reflection attempts, $\ours$ achieves the best accuracy with 2 reflection rounds. Increasing the number of rounds to 3 slightly reduces accuracy and increases cost, suggesting that excessive reflection may introduce redundant feedback into the context window. 

\begin{table}[h]
\centering
\small
    \begin{tabular}{ccc}
    \toprule
    \textbf{Max reflection attempts} & \textbf{Acc (\%)} & \textbf{Cost (\$)} \\
    \midrule
    1 & 43.50 & 0.103 \\
    2 & 47.92 & 0.142 \\
    3 & 46.52 & 0.177 \\
    \bottomrule
    \end{tabular}
\caption{Sensitivity analysis of max reflection attempts}
\label{tab:max_reflection}
\end{table}

For max retry, increasing the threshold from 1 to 2 yields only a marginal accuracy gain while substantially increasing cost. This indicates that most failures are already handled by cross-agent reflection or recovered within the first role-based failover attempt, consistent with the nested procedure in Algorithm~\ref{algo:overall_algorithm}.

\begin{table}[h]
\centering
\small

    \begin{tabular}{ccc}
    \toprule
    \textbf{Max retry} & \textbf{Acc (\%)} & \textbf{Cost (\$)} \\
    \midrule
    1 & 49.76 & 0.105 \\
    2 & 50.65 & 0.176 \\
    \bottomrule
    \end{tabular}
    
\caption{Sensitivity analysis of max retry}
\label{tab:max_retry}
\end{table}
%

%
\begin{figure*}[!t]
  \centering
    \includegraphics[width=\textwidth]{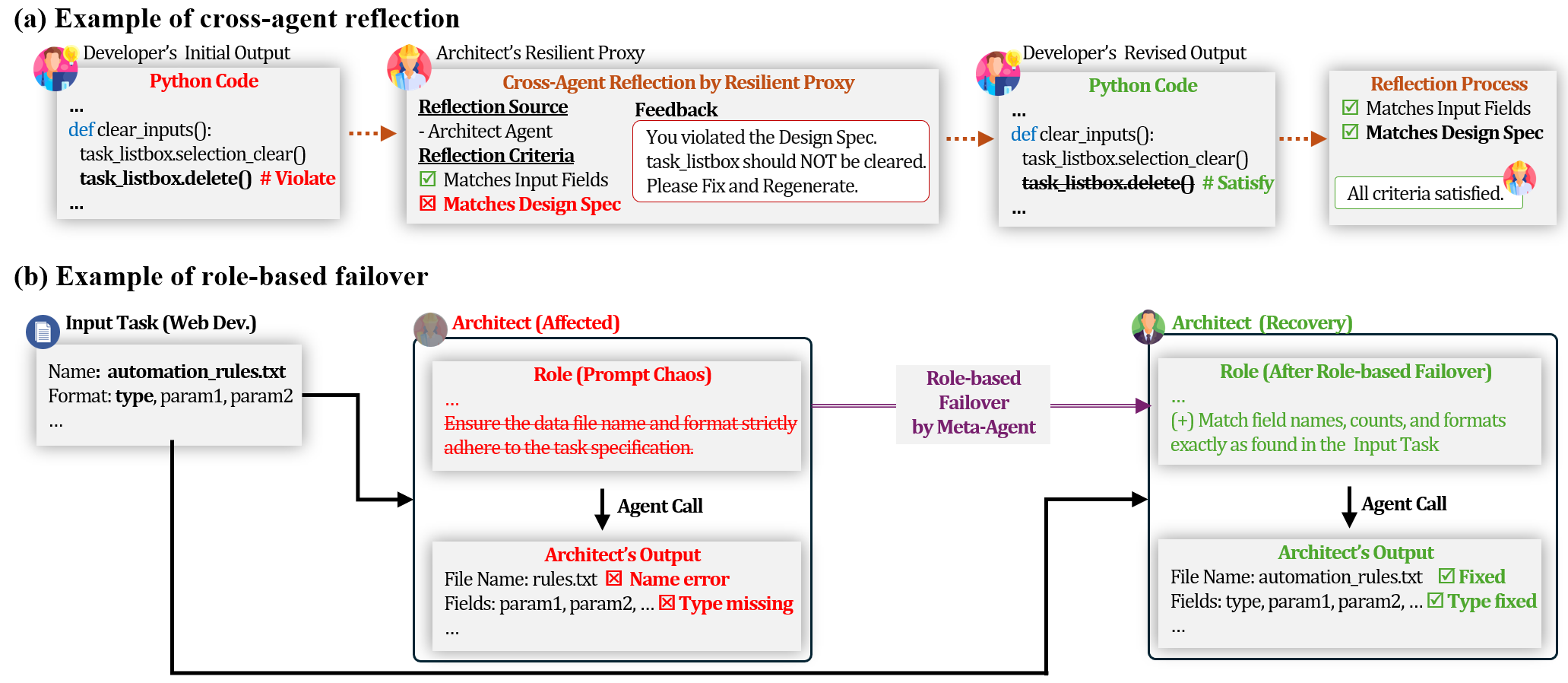}
    \caption{Examples of resilient proxy behavior in $\ours$}
    \label{fig:resilient_proxy_examples}
\end{figure*}

\subsection{Case Study}\label{appendix:d:1}
Figure~\ref{fig:resilient_proxy_examples} qualitatively illustrates the behavior of the resilient proxy in $\ours$. 
The upper part of the figure shows a case of cross-agent reflection, where the architect agent’s proxy evaluates the developer agent’s implementation against the architectural perspective. The proxy detects an incorrect implementation of the \textit{clear\_inputs} function that inadvertently removes the listbox and provides feedback, enabling the developer agent to correct the error. 
The lower part of the figure shows role-based failover. Here, partial corruption of the architect’s role prompt leads to an incorrect system design that overlooks the input format. Upon triggering failover, the role prompt is reconstructed with explicit constraints on data formats and required fields, allowing the architect agent to resume correct behavior.

\subsection{Detailed Main Experiment Results} \label{sec:appendix:f:5}
Tables~\ref{tab:gui_dev_main},~\ref{tab:web_dev_main},~\ref{tab:gaia_main}, and~\ref{tab:gpqa_diamond_main} report additional experiment results that could not be included in Table~\ref{tab:main} of the main manuscript due to space constraints. In particular, we present the full set of detection performance metrics, including detection recall (DR) and detection F1 score (DF1), across all benchmarks. 
As shown, $\ours$ consistently achieves the highest DF1 for all cases, indicating its superiority in accurately identifying affected agents under diverse system uncertainty scenarios.


\begin{table*}[t]
\centering
\small
\setlength{\tabcolsep}{3.5pt}
\renewcommand{\arraystretch}{1.1}
\begin{adjustbox}{width=\textwidth}
\begin{tabular}{@{}l c ccc ccc ccc ccc@{}}
\toprule
\multicolumn{1}{c}{\textbf{GUI Dev.}} &
\multicolumn{1}{c}{\textbf{Vanilla}} &
\multicolumn{3}{c}{\textbf{Agent Chaos}} &
\multicolumn{3}{c}{\textbf{Prompt Chaos}} &
\multicolumn{3}{c}{\textbf{I/O Chaos}} &
\multicolumn{3}{c}{\textbf{Compound Chaos}} \\
\cmidrule(lr){2-2}\cmidrule(lr){3-5}\cmidrule(lr){6-8}\cmidrule(lr){9-11}\cmidrule(lr){12-14}
\textbf{Method} &
\textbf{Acc (\%)} &
\textbf{Acc (\%)} & \textbf{DR} & \textbf{DF1} &
\textbf{Acc (\%)} & \textbf{DR} & \textbf{DF1} &
\textbf{Acc (\%)} & \textbf{DR} & \textbf{DF1} &
\textbf{Acc (\%)} & \textbf{DR} & \textbf{DF1} \\
\midrule
CoT
& 66.02{\color{gray}\scriptsize$\pm$2.53}
& 0.00{\color{gray}\scriptsize$\pm$0.00} & 0.00{\color{gray}\scriptsize$\pm$0.00} & 0.00{\color{gray}\scriptsize$\pm$0.00}
& 25.03{\color{gray}\scriptsize$\pm$5.87} & 0.00{\color{gray}\scriptsize$\pm$0.00} & 0.00{\color{gray}\scriptsize$\pm$0.00}
& 0.00{\color{gray}\scriptsize$\pm$0.00} & 0.00{\color{gray}\scriptsize$\pm$0.00} & 0.00{\color{gray}\scriptsize$\pm$0.00}
& 3.96{\color{gray}\scriptsize$\pm$4.40} & 0.00{\color{gray}\scriptsize$\pm$0.00} & 0.00{\color{gray}\scriptsize$\pm$0.00} \\
RTADev
& 66.30{\color{gray}\scriptsize$\pm$1.60}
& 21.87{\color{gray}\scriptsize$\pm$8.35} & 0.61{\color{gray}\scriptsize$\pm$0.15} & 0.51{\color{gray}\scriptsize$\pm$0.09}
& 39.64{\color{gray}\scriptsize$\pm$3.27} & 0.42{\color{gray}\scriptsize$\pm$0.14} & 0.42{\color{gray}\scriptsize$\pm$0.12}
& 33.90{\color{gray}\scriptsize$\pm$13.38} & 0.19{\color{gray}\scriptsize$\pm$0.17} & 0.28{\color{gray}\scriptsize$\pm$0.06}
& 14.99{\color{gray}\scriptsize$\pm$8.29} & 0.43{\color{gray}\scriptsize$\pm$0.09} & 0.47{\color{gray}\scriptsize$\pm$0.07} \\
MetaGPT
& 70.51{\color{gray}\scriptsize$\pm$4.22}
& 7.39{\color{gray}\scriptsize$\pm$6.10} & 0.00{\color{gray}\scriptsize$\pm$0.00} & 0.00{\color{gray}\scriptsize$\pm$0.00}
& 21.62{\color{gray}\scriptsize$\pm$5.12} & 0.00{\color{gray}\scriptsize$\pm$0.00} & 0.00{\color{gray}\scriptsize$\pm$0.00}
& 22.63{\color{gray}\scriptsize$\pm$11.46} & 0.00{\color{gray}\scriptsize$\pm$0.00} & 0.00{\color{gray}\scriptsize$\pm$0.00}
& 0.00{\color{gray}\scriptsize$\pm$0.00} & 0.00{\color{gray}\scriptsize$\pm$0.00} & 0.00{\color{gray}\scriptsize$\pm$0.00} \\
C+I
& 68.05{\color{gray}\scriptsize$\pm$2.46}
& 4.46{\color{gray}\scriptsize$\pm$4.47} & 0.24{\color{gray}\scriptsize$\pm$0.03} & 0.28{\color{gray}\scriptsize$\pm$0.08}
& 19.30{\color{gray}\scriptsize$\pm$6.23} & 0.29{\color{gray}\scriptsize$\pm$0.04} & 0.22{\color{gray}\scriptsize$\pm$0.04}
& 47.37{\color{gray}\scriptsize$\pm$4.24} & \textbf{0.86{\color{gray}\scriptsize$\pm$0.03}} & 0.60{\color{gray}\scriptsize$\pm$0.04}
& 1.59{\color{gray}\scriptsize$\pm$2.75} & 0.30{\color{gray}\scriptsize$\pm$0.03} & 0.39{\color{gray}\scriptsize$\pm$0.06} \\
AgentVerse
& 66.53{\color{gray}\scriptsize$\pm$6.59}
& 32.87{\color{gray}\scriptsize$\pm$6.03} & 0.00{\color{gray}\scriptsize$\pm$0.00} & 0.00{\color{gray}\scriptsize$\pm$0.00}
& 44.61{\color{gray}\scriptsize$\pm$8.08} & 0.00{\color{gray}\scriptsize$\pm$0.00} & 0.00{\color{gray}\scriptsize$\pm$0.00}
& 37.89{\color{gray}\scriptsize$\pm$12.94} & 0.00{\color{gray}\scriptsize$\pm$0.00} & 0.00{\color{gray}\scriptsize$\pm$0.00}
& 26.36{\color{gray}\scriptsize$\pm$6.13} & 0.00{\color{gray}\scriptsize$\pm$0.00} & 0.00{\color{gray}\scriptsize$\pm$0.00} \\

EvoMAC
& 72.79{\color{gray}\scriptsize$\pm$1.96}
& 48.01{\color{gray}\scriptsize$\pm$11.28} & 0.00{\color{gray}\scriptsize$\pm$0.00} & 0.00{\color{gray}\scriptsize$\pm$0.00}
& 52.15{\color{gray}\scriptsize$\pm$8.34} & 0.00{\color{gray}\scriptsize$\pm$0.00} & 0.00{\color{gray}\scriptsize$\pm$0.00}
& 63.65{\color{gray}\scriptsize$\pm$7.45} & 0.00{\color{gray}\scriptsize$\pm$0.00} & 0.00{\color{gray}\scriptsize$\pm$0.00}
& 24.38{\color{gray}\scriptsize$\pm$11.92} & 0.00{\color{gray}\scriptsize$\pm$0.00} & 0.00{\color{gray}\scriptsize$\pm$0.00} \\

\cellcolor{gray!10}\textbf{$\ours$}
& \cellcolor{gray!10}\textbf{75.29{\color{gray}\scriptsize$\pm$4.34}}
& \cellcolor{gray!10}\textbf{71.05{\color{gray}\scriptsize$\pm$1.64}} & \cellcolor{gray!10}\textbf{1.00{\color{gray}\scriptsize$\pm$0.00}} & \cellcolor{gray!10}\textbf{0.94{\color{gray}\scriptsize$\pm$0.04}}
& \cellcolor{gray!10}\textbf{65.30{\color{gray}\scriptsize$\pm$5.48}} & \cellcolor{gray!10}\textbf{0.84{\color{gray}\scriptsize$\pm$0.08}} & \cellcolor{gray!10}\textbf{0.75{\color{gray}\scriptsize$\pm$0.06}}
& \cellcolor{gray!10}\textbf{67.87{\color{gray}\scriptsize$\pm$5.07}} & \cellcolor{gray!10}0.82{\color{gray}\scriptsize$\pm$0.07} & \cellcolor{gray!10}\textbf{0.73{\color{gray}\scriptsize$\pm$0.09}}
& \cellcolor{gray!10}\textbf{65.66{\color{gray}\scriptsize$\pm$5.69}} & \cellcolor{gray!10}\textbf{0.90{\color{gray}\scriptsize$\pm$0.05}} & \textbf{0.92{\color{gray}\scriptsize$\pm$0.03}} \\
\bottomrule
\end{tabular}
\end{adjustbox}
\caption{Detailed performance on GUI development benchmark}
\label{tab:gui_dev_main}
\end{table*}

\begin{table*}[h!]
\centering
\small
\setlength{\tabcolsep}{3.5pt}
\renewcommand{\arraystretch}{1.1}
\begin{adjustbox}{width=\textwidth}
\begin{tabular}{@{}l c ccc ccc ccc ccc@{}}
\toprule
\multicolumn{1}{c}{\textbf{Web Dev.}} &
\multicolumn{1}{c}{\textbf{Vanilla}} &
\multicolumn{3}{c}{\textbf{Agent Chaos}} &
\multicolumn{3}{c}{\textbf{Prompt Chaos}} &
\multicolumn{3}{c}{\textbf{I/O Chaos}} &
\multicolumn{3}{c}{\textbf{Compound Chaos}} \\
\cmidrule(lr){2-2}\cmidrule(lr){3-5}\cmidrule(lr){6-8}\cmidrule(lr){9-11}\cmidrule(lr){12-14}
\textbf{Method} &
\textbf{Acc (\%)} &
\textbf{Acc (\%)} & \textbf{DR} & \textbf{DF1} &
\textbf{Acc (\%)} & \textbf{DR} & \textbf{DF1} &
\textbf{Acc (\%)} & \textbf{DR} & \textbf{DF1} &
\textbf{Acc (\%)} & \textbf{DR} & \textbf{DF1} \\
\midrule
CoT
& 45.73{\color{gray}\scriptsize$\pm$3.02}
& 0.00{\color{gray}\scriptsize$\pm$0.00} & 0.00{\color{gray}\scriptsize$\pm$0.00} & 0.00{\color{gray}\scriptsize$\pm$0.00}
& 0.60{\color{gray}\scriptsize$\pm$1.04} & 0.00{\color{gray}\scriptsize$\pm$0.00} & 0.00{\color{gray}\scriptsize$\pm$0.00}
& 0.00{\color{gray}\scriptsize$\pm$0.00} & 0.00{\color{gray}\scriptsize$\pm$0.00} & 0.00{\color{gray}\scriptsize$\pm$0.00}
& 0.19{\color{gray}\scriptsize$\pm$0.32} & 0.00{\color{gray}\scriptsize$\pm$0.00} & 0.00{\color{gray}\scriptsize$\pm$0.00} \\
RTADev
& 40.17{\color{gray}\scriptsize$\pm$6.78}
& 12.68{\color{gray}\scriptsize$\pm$4.60} & 0.44{\color{gray}\scriptsize$\pm$0.17} & 0.42{\color{gray}\scriptsize$\pm$0.16}
& 8.85{\color{gray}\scriptsize$\pm$0.53}  & 0.46{\color{gray}\scriptsize$\pm$0.02} & 0.41{\color{gray}\scriptsize$\pm$0.04}
& 16.74{\color{gray}\scriptsize$\pm$4.50} & 0.20{\color{gray}\scriptsize$\pm$0.11} & 0.18{\color{gray}\scriptsize$\pm$0.11}
& 2.68{\color{gray}\scriptsize$\pm$4.12}  & 0.36{\color{gray}\scriptsize$\pm$0.06} & 0.41{\color{gray}\scriptsize$\pm$0.03} \\
MetaGPT
& 42.09{\color{gray}\scriptsize$\pm$7.03}
& 4.81{\color{gray}\scriptsize$\pm$2.77}  & 0.00{\color{gray}\scriptsize$\pm$0.00} & 0.00{\color{gray}\scriptsize$\pm$0.00}
& 6.30{\color{gray}\scriptsize$\pm$5.03}  & 0.00{\color{gray}\scriptsize$\pm$0.00} & 0.00{\color{gray}\scriptsize$\pm$0.00}
& 11.91{\color{gray}\scriptsize$\pm$3.30} & 0.00{\color{gray}\scriptsize$\pm$0.00} & 0.00{\color{gray}\scriptsize$\pm$0.00}
& 0.41{\color{gray}\scriptsize$\pm$0.35}  & 0.00{\color{gray}\scriptsize$\pm$0.00} & 0.00{\color{gray}\scriptsize$\pm$0.00} \\
C+I
& 45.25{\color{gray}\scriptsize$\pm$3.67}
& 5.72{\color{gray}\scriptsize$\pm$2.81}  & 0.10{\color{gray}\scriptsize$\pm$0.14} & 0.12{\color{gray}\scriptsize$\pm$0.18}
& 12.47{\color{gray}\scriptsize$\pm$9.17} & 0.31{\color{gray}\scriptsize$\pm$0.07} & 0.21{\color{gray}\scriptsize$\pm$0.04}
& 14.03{\color{gray}\scriptsize$\pm$0.39} & \textbf{0.75{\color{gray}\scriptsize$\pm$0.10}} & 0.57{\color{gray}\scriptsize$\pm$0.10}
& 0.00{\color{gray}\scriptsize$\pm$0.00}  & 0.25{\color{gray}\scriptsize$\pm$0.04} & 0.35{\color{gray}\scriptsize$\pm$0.06} \\
AgentVerse
& 52.01{\color{gray}\scriptsize$\pm$2.06}
& 11.95{\color{gray}\scriptsize$\pm$6.38} & 0.00{\color{gray}\scriptsize$\pm$0.00} & 0.00{\color{gray}\scriptsize$\pm$0.00}
& 21.55{\color{gray}\scriptsize$\pm$8.06} & 0.00{\color{gray}\scriptsize$\pm$0.00} & 0.00{\color{gray}\scriptsize$\pm$0.00}
& 12.50{\color{gray}\scriptsize$\pm$1.38} & 0.00{\color{gray}\scriptsize$\pm$0.00} & 0.00{\color{gray}\scriptsize$\pm$0.00}
& 18.38{\color{gray}\scriptsize$\pm$3.77} & 0.00{\color{gray}\scriptsize$\pm$0.00} & 0.00{\color{gray}\scriptsize$\pm$0.00} \\

EvoMAC
& 41.33{\color{gray}\scriptsize$\pm$4.47}
& 20.45{\color{gray}\scriptsize$\pm$4.72} & 0.00{\color{gray}\scriptsize$\pm$0.00} & 0.00{\color{gray}\scriptsize$\pm$0.00}
& 24.29{\color{gray}\scriptsize$\pm$2.56} & 0.00{\color{gray}\scriptsize$\pm$0.00} & 0.00{\color{gray}\scriptsize$\pm$0.00}
& 14.72{\color{gray}\scriptsize$\pm$7.17} & 0.00{\color{gray}\scriptsize$\pm$0.00} & 0.00{\color{gray}\scriptsize$\pm$0.00}
& 5.80{\color{gray}\scriptsize$\pm$5.10} & 0.00{\color{gray}\scriptsize$\pm$0.00} & 0.00{\color{gray}\scriptsize$\pm$0.00} \\

\cellcolor{gray!10}\textbf{$\ours$}
& \cellcolor{gray!10}\textbf{59.68{\color{gray}\scriptsize$\pm$1.84}}
& \cellcolor{gray!10}\textbf{53.98{\color{gray}\scriptsize$\pm$1.55}} & \cellcolor{gray!10}\textbf{1.00{\color{gray}\scriptsize$\pm$0.00}} & \cellcolor{gray!10}\textbf{0.93{\color{gray}\scriptsize$\pm$0.03}}
& \cellcolor{gray!10}\textbf{53.11{\color{gray}\scriptsize$\pm$1.82}} & \cellcolor{gray!10}\textbf{0.94{\color{gray}\scriptsize$\pm$0.04}} & \cellcolor{gray!10}\textbf{0.90{\color{gray}\scriptsize$\pm$0.09}}
& \cellcolor{gray!10}\textbf{54.12{\color{gray}\scriptsize$\pm$1.58}} & \cellcolor{gray!10}\textbf{0.75{\color{gray}\scriptsize$\pm$0.13}} & \cellcolor{gray!10}\textbf{0.68{\color{gray}\scriptsize$\pm$0.15}}
& \cellcolor{gray!10}\textbf{49.76{\color{gray}\scriptsize$\pm$0.32}} & \cellcolor{gray!10}\textbf{0.91{\color{gray}\scriptsize$\pm$0.02}} & \cellcolor{gray!10}\textbf{0.90{\color{gray}\scriptsize$\pm$0.03}} \\
\bottomrule
\end{tabular}
\end{adjustbox}
\caption{Detailed performance on web development benchmark}
\label{tab:web_dev_main}
\end{table*}

\begin{table*}[h!]
\centering
\small
\setlength{\tabcolsep}{3.5pt}
\renewcommand{\arraystretch}{1.1}
\begin{adjustbox}{width=\textwidth}
\begin{tabular}{@{}l c ccc ccc ccc ccc@{}}
\toprule
\multicolumn{1}{c}{\textbf{GAIA}} &
\multicolumn{1}{c}{\textbf{Vanilla}} &
\multicolumn{3}{c}{\textbf{Agent Chaos}} &
\multicolumn{3}{c}{\textbf{Prompt Chaos}} &
\multicolumn{3}{c}{\textbf{I/O Chaos}} &
\multicolumn{3}{c}{\textbf{Compound Chaos}} \\
\cmidrule(lr){2-2}\cmidrule(lr){3-5}\cmidrule(lr){6-8}\cmidrule(lr){9-11}\cmidrule(lr){12-14}
\textbf{Method} &
\textbf{Acc (\%)} &
\textbf{Acc (\%)} & \textbf{DR} & \textbf{DF1} &
\textbf{Acc (\%)} & \textbf{DR} & \textbf{DF1} &
\textbf{Acc (\%)} & \textbf{DR} & \textbf{DF1} &
\textbf{Acc (\%)} & \textbf{DR} & \textbf{DF1} \\
\midrule
CoT
& 28.61{\color{gray}\scriptsize$\pm$3.57}
& 0.00{\color{gray}\scriptsize$\pm$0.00}  & 0.00{\color{gray}\scriptsize$\pm$0.00} & 0.00{\color{gray}\scriptsize$\pm$0.00}
& 1.01{\color{gray}\scriptsize$\pm$0.35}  & 0.00{\color{gray}\scriptsize$\pm$0.00} & 0.00{\color{gray}\scriptsize$\pm$0.00}
& 4.85{\color{gray}\scriptsize$\pm$1.60}  & 0.00{\color{gray}\scriptsize$\pm$0.00} & 0.00{\color{gray}\scriptsize$\pm$0.00}
& 0.61{\color{gray}\scriptsize$\pm$0.61}  & 0.00{\color{gray}\scriptsize$\pm$0.00} & 0.00{\color{gray}\scriptsize$\pm$0.00} \\
LLM-Debate
& 32.32{\color{gray}\scriptsize$\pm$1.85}
& 13.13{\color{gray}\scriptsize$\pm$2.45} & 0.00{\color{gray}\scriptsize$\pm$0.00} & 0.00{\color{gray}\scriptsize$\pm$0.00}
& 19.39{\color{gray}\scriptsize$\pm$2.10} & 0.00{\color{gray}\scriptsize$\pm$0.00} & 0.00{\color{gray}\scriptsize$\pm$0.00}
& 15.96{\color{gray}\scriptsize$\pm$3.45} & 0.00{\color{gray}\scriptsize$\pm$0.00} & 0.00{\color{gray}\scriptsize$\pm$0.00}
& 5.25{\color{gray}\scriptsize$\pm$0.93}  & 0.00{\color{gray}\scriptsize$\pm$0.00} & 0.00{\color{gray}\scriptsize$\pm$0.00} \\
C+I
& 30.10{\color{gray}\scriptsize$\pm$0.92}
& 14.75{\color{gray}\scriptsize$\pm$3.11} & 0.53{\color{gray}\scriptsize$\pm$0.06} & 0.39{\color{gray}\scriptsize$\pm$0.05}
& 20.81{\color{gray}\scriptsize$\pm$1.95} & 0.62{\color{gray}\scriptsize$\pm$0.04} & 0.45{\color{gray}\scriptsize$\pm$0.06}
& 17.58{\color{gray}\scriptsize$\pm$5.97} & \textbf{0.99{\color{gray}\scriptsize$\pm$0.00}} & 0.51{\color{gray}\scriptsize$\pm$0.07}
& 3.70{\color{gray}\scriptsize$\pm$1.91}  & 0.72{\color{gray}\scriptsize$\pm$0.09} & 0.75{\color{gray}\scriptsize$\pm$0.09} \\
AgentVerse
& 24.31{\color{gray}\scriptsize$\pm$2.38}
& 13.18{\color{gray}\scriptsize$\pm$1.45} & 0.00{\color{gray}\scriptsize$\pm$0.00} & 0.00{\color{gray}\scriptsize$\pm$0.00}
& 7.07{\color{gray}\scriptsize$\pm$1.75}  & 0.00{\color{gray}\scriptsize$\pm$0.00} & 0.00{\color{gray}\scriptsize$\pm$0.00}
& 11.31{\color{gray}\scriptsize$\pm$5.44} & 0.00{\color{gray}\scriptsize$\pm$0.00} & 0.00{\color{gray}\scriptsize$\pm$0.00}
& 7.68{\color{gray}\scriptsize$\pm$6.48}  & 0.00{\color{gray}\scriptsize$\pm$0.00} & 0.00{\color{gray}\scriptsize$\pm$0.00} \\
\cellcolor{gray!10}\textbf{$\ours$}
& \cellcolor{gray!10}\textbf{32.73{\color{gray}\scriptsize$\pm$1.60}}
& \cellcolor{gray!10}\textbf{29.90{\color{gray}\scriptsize$\pm$4.94}} & \cellcolor{gray!10}\textbf{1.00{\color{gray}\scriptsize$\pm$0.00}} & \cellcolor{gray!10}\textbf{0.74{\color{gray}\scriptsize$\pm$0.04}}
& \cellcolor{gray!10}\textbf{26.46{\color{gray}\scriptsize$\pm$3.55}} & \cellcolor{gray!10}\textbf{0.83{\color{gray}\scriptsize$\pm$0.02}} & \cellcolor{gray!10}\textbf{0.59{\color{gray}\scriptsize$\pm$0.02}}
& \cellcolor{gray!10}\textbf{29.09{\color{gray}\scriptsize$\pm$0.61}} & \cellcolor{gray!10}0.92{\color{gray}\scriptsize$\pm$0.01} & \cellcolor{gray!10}\textbf{0.63{\color{gray}\scriptsize$\pm$0.02}}
& \cellcolor{gray!10}\textbf{24.44{\color{gray}\scriptsize$\pm$4.59}} & \cellcolor{gray!10}\textbf{0.94{\color{gray}\scriptsize$\pm$0.01}} & \cellcolor{gray!10}\textbf{0.86{\color{gray}\scriptsize$\pm$0.03}} \\
\bottomrule
\end{tabular}
\end{adjustbox}
\caption{Detailed performance on GAIA benchmark}
\label{tab:gaia_main}
\end{table*}

\begin{table*}[h!]
\centering
\small
\setlength{\tabcolsep}{3.5pt}
\renewcommand{\arraystretch}{1.1}
\begin{adjustbox}{width=\textwidth}
\begin{tabular}{@{}l c ccc ccc ccc ccc@{}}
\toprule
\multicolumn{1}{c}{\textbf{GPQA-Diamond}} &
\multicolumn{1}{c}{\textbf{Vanilla}} &
\multicolumn{3}{c}{\textbf{Agent Chaos}} &
\multicolumn{3}{c}{\textbf{Prompt Chaos}} &
\multicolumn{3}{c}{\textbf{I/O Chaos}} &
\multicolumn{3}{c}{\textbf{Compound Chaos}} \\
\cmidrule(lr){2-2}\cmidrule(lr){3-5}\cmidrule(lr){6-8}\cmidrule(lr){9-11}\cmidrule(lr){12-14}
\textbf{Method} &
\textbf{Acc (\%)} &
\textbf{Acc (\%)} & \textbf{DR} & \textbf{DF1} &
\textbf{Acc (\%)} & \textbf{DR} & \textbf{DF1} &
\textbf{Acc (\%)} & \textbf{DR} & \textbf{DF1} &
\textbf{Acc (\%)} & \textbf{DR} & \textbf{DF1} \\
\midrule
CoT
& 64.14{\color{gray}\scriptsize$\pm$1.01}
& 0.00{\color{gray}\scriptsize$\pm$0.00}  & 0.00{\color{gray}\scriptsize$\pm$0.00} & 0.00{\color{gray}\scriptsize$\pm$0.00}
& 27.94{\color{gray}\scriptsize$\pm$0.58} & 0.00{\color{gray}\scriptsize$\pm$0.00} & 0.00{\color{gray}\scriptsize$\pm$0.00}
& 30.97{\color{gray}\scriptsize$\pm$2.10} & 0.00{\color{gray}\scriptsize$\pm$0.00} & 0.00{\color{gray}\scriptsize$\pm$0.00}
& 20.04{\color{gray}\scriptsize$\pm$1.54} & 0.00{\color{gray}\scriptsize$\pm$0.00} & 0.00{\color{gray}\scriptsize$\pm$0.00} \\
LLM-Debate
& 66.50{\color{gray}\scriptsize$\pm$1.27}
& 26.77{\color{gray}\scriptsize$\pm$7.02} & 0.00{\color{gray}\scriptsize$\pm$0.00} & 0.00{\color{gray}\scriptsize$\pm$0.00}
& 35.52{\color{gray}\scriptsize$\pm$7.65} & 0.00{\color{gray}\scriptsize$\pm$0.00} & 0.00{\color{gray}\scriptsize$\pm$0.00}
& 41.41{\color{gray}\scriptsize$\pm$4.01} & 0.00{\color{gray}\scriptsize$\pm$0.00} & 0.00{\color{gray}\scriptsize$\pm$0.00}
& 15.32{\color{gray}\scriptsize$\pm$3.44} & 0.00{\color{gray}\scriptsize$\pm$0.00} & 0.00{\color{gray}\scriptsize$\pm$0.00} \\
C+I
& 65.15{\color{gray}\scriptsize$\pm$1.33}
& 35.35{\color{gray}\scriptsize$\pm$3.81} & 0.66{\color{gray}\scriptsize$\pm$0.09} & 0.58{\color{gray}\scriptsize$\pm$0.07}
& 52.36{\color{gray}\scriptsize$\pm$2.04} & 0.52{\color{gray}\scriptsize$\pm$0.02} & 0.40{\color{gray}\scriptsize$\pm$0.07}
& 51.67{\color{gray}\scriptsize$\pm$4.22} & \textbf{0.96{\color{gray}\scriptsize$\pm$0.03}} & 0.69{\color{gray}\scriptsize$\pm$0.03}
& 16.33{\color{gray}\scriptsize$\pm$2.49} & 0.66{\color{gray}\scriptsize$\pm$0.22} & 0.67{\color{gray}\scriptsize$\pm$0.25} \\
AgentVerse
& 62.46{\color{gray}\scriptsize$\pm$2.59}
& 28.62{\color{gray}\scriptsize$\pm$2.96} & 0.00{\color{gray}\scriptsize$\pm$0.00} & 0.00{\color{gray}\scriptsize$\pm$0.00}
& 31.14{\color{gray}\scriptsize$\pm$3.04} & 0.00{\color{gray}\scriptsize$\pm$0.00} & 0.00{\color{gray}\scriptsize$\pm$0.00}
& 27.95{\color{gray}\scriptsize$\pm$11.21}& 0.00{\color{gray}\scriptsize$\pm$0.00} & 0.00{\color{gray}\scriptsize$\pm$0.00}
& 22.90{\color{gray}\scriptsize$\pm$6.11} & 0.00{\color{gray}\scriptsize$\pm$0.00} & 0.00{\color{gray}\scriptsize$\pm$0.00} \\
\cellcolor{gray!10}\textbf{$\ours$}
& \cellcolor{gray!10}\textbf{67.68{\color{gray}\scriptsize$\pm$2.20}}
& \cellcolor{gray!10}\textbf{61.87{\color{gray}\scriptsize$\pm$1.07}} & \cellcolor{gray!10}\textbf{1.00{\color{gray}\scriptsize$\pm$0.00}} & \cellcolor{gray!10}\textbf{0.94{\color{gray}\scriptsize$\pm$0.03}}
& \cellcolor{gray!10}\textbf{62.46{\color{gray}\scriptsize$\pm$3.79}} & \cellcolor{gray!10}\textbf{0.81{\color{gray}\scriptsize$\pm$0.02}} & \cellcolor{gray!10}\textbf{0.71{\color{gray}\scriptsize$\pm$0.03}}
& \cellcolor{gray!10}\textbf{61.11{\color{gray}\scriptsize$\pm$2.53}} & \cellcolor{gray!10}\textbf{0.98{\color{gray}\scriptsize$\pm$0.01}} & \cellcolor{gray!10}\textbf{0.87{\color{gray}\scriptsize$\pm$0.03}}
& \cellcolor{gray!10}\textbf{61.45{\color{gray}\scriptsize$\pm$3.04}} & \cellcolor{gray!10}\textbf{0.96{\color{gray}\scriptsize$\pm$0.01}} & \cellcolor{gray!10}\textbf{0.95{\color{gray}\scriptsize$\pm$0.00}} \\
\bottomrule
\end{tabular}
\end{adjustbox}
\caption{Detailed performance on GPQA-Diamond benchmark}
\label{tab:gpqa_diamond_main}
\end{table*}


\section{Prompts used in $\ours$}\label{app:prompts}
In this section, we provide the prompts used by the LLM orchestrator in $\ours$ for constructing resilient multi-agent systems. 
Specifically, Figure~\ref{fig:appendix:e:spec_prompt} shows the prompt for generating the resilience-aware system specification, Figure~\ref{fig:appendix:e:agent_prompt} shows the prompt for generating role-specific agent prompts, and Figure~\ref{fig:appendix:e:policy_code_prompt} shows the prompt for synthesizing executable multi-agent policy code.


\begin{figure*}[t]
  \centering
  \begin{tcblisting}{campPlainListingBox,title=\textbf{Prompt for Resilience-Aware System Specification}}
You are an LLM orchestrator for generating a resilience-aware system specification.
Your task is to derive a structured specification for a resilient multi-agent system from a given task.
Your deliverable is the resilience-aware system specification, consisting of collaboration logic and cross-agent reflection.

You must follow this procedure systematically.
@@Step 1: Task Understanding & Complexity Assessment@@
1.1. Goal Analysis
- Extract the core objective
- Identify the primary deliverable (function implementation, analysis report, etc.)
1.2. Domain-Driven SOP Selection
- CodeGeneration Domain: Architecture -> Implementation
- Question Answering Domain: Data Extraction -> Answer Generation
- Custom Domain: Adapt based on domain characteristics
1.3. Complexity Assessment
- Low: Simple function, basic logic, minimal constraints -> SOP-LITE: 1-2 phases 
- Medium: Multiple components, moderate constraints, some edge cases -> SOP-STANDARD: 2-4 phases 
- High: Complex algorithms, extensive constraints, validation requirements -> SOP-FULL: 2-6 phases
1.4. Workload Assessment & Agent Team Sizing
When to Divide Work:
When a single LLM agent would have difficulty executing their responsibility due to excessive scope, divide the workload among multiple specialized agents using the Parallel & Merge pattern.
Evaluation Considerations:
- Volume: Many distinct outputs or artifacts to produce
- Diversity: Multiple unrelated functional areas or reasoning stages
- Complexity: Multi-tier systems or interdependent workflows
Division Strategy:
- Split by functional domain or problem-solving stage
- Use descriptive agent names reflecting their specialty
- Require a shared specification artifact from a preceding phase to ensure consistency

@@Step 2: Strategic Sub-task Decomposition@@
Based on Domain and Complexity, decompose user task into logical phases.
2.1 Phase Identification: Break user_task into 1-6 sub-goals where:
- Each sub-goal directly contributes to the main Goal
- Sub-goals form a logical dependency chain
- Each phase transforms concrete artifacts
- Guarantee: When all sub-goals are completed, the user task is fully achieved (no gaps or redundant steps)

2.2. Phase Consolidation Rules
- Agent Reuse Check: If the same agents appear in consecutive phases doing related work
, consolidate them into a single phase with appropriate collaboration pattern
- CRITICAL: Never put the same agent_name in multiple phases' agent_profiles lists
- Avoid Unnecessary Separation: Only create separate phases when different agent teams or fundamentally different objectives 
are involved

@@Step 3: Agent Team Design & Coordination Structure Selection@@
For each sub_task, design the exact workflow function with required parameters: {...}

@@3.1. sub_task@@
- A concise string describing the sub-task objective and its expected deliverables.

@@3.2. coordination_structure (per sub_task)@@
Choose from available patterns or create new combinations:
Available Coordination Structures: {...}

@@3.3. agent_profiles@@
For each agent profile, specify the required fields:
agent_name: Clear, role-revealing name
responsibility: Specific responsibility the agent must complete.
Role Design Principles:
- Define roles in terms of the sub_task-specific objective, not task-specific details
- Focus on the type of work rather than specific application names or features
- Include technical constraints that apply across similar tasks
tools: You must select from the following list of {Available_tools}
input_artifacts: A list of data artifacts this agent requires to start its work.
Each artifact is a dictionary with the following keys:
- type: The type of data: "user", "text_file", "others_file" (binary/non-text files)
- name: The name of the file or a description of the string data.
- source: The name of the agent that produced this artifact.
output_artifacts: Each artifact is a dictionary with the following keys:
- type: The type of data: "text_file", "others_file".
- name: The name of the file to be created or a description of the string data to be returned.

@@3.4. cross_agent_reflection@@
Specify the resilient proxy specifications for cross-agent reflection:
- reflection_source: The resilient proxy associated with the reviewing agent.
- reflection_targets: The agent(s) whose output is evaluated.
- reflection_criteria: Role-aware criteria used to evaluate the target agent's output.

Refer to the examples below:
{Examples}
  \end{tcblisting}
  \caption{Prompt for generating the resilience-aware system specification}
  \label{fig:appendix:e:spec_prompt}
\end{figure*}


\begin{figure*}[t]
  \centering
  \begin{tcblisting}{campPlainListingBox,title=\textbf{Prompt for Agent Role-Specific Prompt Generation}}
You are an LLM orchestrator for generating role-specific agent prompts.
Your task is to transform a subtask-level system specification into concise, 
structured role-specific prompts that each agent can execute immediately.
Your deliverable is the role-specific agent prompts, consisting of agent responsibilities, goals,
task details, tool usage instructions, artifact requirements, and consistency rules.

You must follow this procedure systematically.

@@Step 1: Ingest the sub_task Specification Block@@
Receive the overall task as a string and the complete code block defining the sub_task as input.

@@Step 2: Perform Transformations for Each Agent@@
For each agent dictionary in the agent_profiles list, perform the following transformations:
'responsibility' to 'prompt' Transformation:

Replace the responsibility field with a prompt field. 
The value of this new field must be a markdown string following a concise, 
structured format with essential guidance.

- responsibility: Use domain/technology expertise
- Goal: State generic objectives applicable to the task class with concrete deliverables
- Instructions: Provide concise procedural guidance with 2-4 steps, each containing essential bullets only

Template Structure:
The prompt begins directly with the agent's responsibility description, 
stating their professional identity and domain/technology expertise.

Following the responsibility, state a clear goal describing what the agent must accomplish with concrete deliverables.
The Task Details section provides 3-5 bullet points that specify:
- Which input artifacts to read
- What output artifacts to create and their key components
- Scope boundaries defining what the agent should focus on

The main body consists of 2-4 categorized sections with bold headers.

Each section contains specific requirements, format specifications, 
or step-by-step implementation guidelines relevant to the agent's task.

If the agent participates in multi-agent collaboration, include consistency rules that ensure 
data integrity across agents such as naming conventions, field matching requirements, and references to the source of truth.

End with Requirements that explicitly state:
- Tool usage instructions
- Format and naming constraints
- Focus areas and boundaries
- For refine agents: feedback markers ("[APPROVED]" or "NEED_MODIFY") to write

When generating the "Task Details" section of agent prompts, 
you must reference the specific tools assigned to each agent and provide concrete usage guidance based on their capabilities.
{Available_tools}

Refer to the examples below:
{Examples}
  \end{tcblisting}
  \caption{Prompt for generating role-specific agent prompts from a subtask level system specification}
  \label{fig:appendix:e:agent_prompt}
\end{figure*}


\begin{figure*}[t]
  \centering
  \begin{tcblisting}{campPlainListingBox,title=\textbf{Prompt for Multi-Agent Policy Code Synthesis}}
You are an LLM orchestrator for multi-agent policy code synthesis, 
responsible for transforming a resilience-aware system specification into executable Python code.

Your task is to read the given def block, analyze its structure, 
and generate runnable Python code by leveraging the predefined APIs listed in @@{Available_predefined_APIs}.@@

You must follow these rules:
- Always declare all agents specified in the agent_profiles section using build_agent_with_resilient_proxy.
- Each agent must follow the parameters below:
agent_name: identical to the name in the def block.
proxy_spec: construct the proxy specification from the cross_agent_reflection field associated with the agent 
in the def block.
timeout_threshold: choose a value between 200-500 seconds depending on sub_task complexity.
max_reflection_attempts: choose a value between 1 and 3 to specify the maximum number of cross-agent reflection rounds 
before triggering failover.
max_retry: choose a value between 1 and 2 to specify the maximum number of failover attempts 
for replacing or reconfiguring an affected agent.
- After defining agents, parse the sub_task, coordination_structure, agent_profiles, 
and cross_agent_reflection fields to determine the execution order and resilience configuration.

You must follow this procedure systematically.

Step 1: Parse the def block.

Step 2: Declare all agents with build_agent_with_resilient_proxy and proper parameters.

Step 3: Read coordination fields (coordination_structure, agent_profiles, cross_agent_reflection).

@@Step 4: Construct execution flow by combining execute calls (sequential or parallel using asyncio.gather).@@

@@Step 5: Return a final runnable Python script based on predefined APIs.@@

Refer to the examples below:
{Examples}
  \end{tcblisting}
  \caption{Prompt for synthesizing executable multi-agent policy code from the resilience-aware specification}
  \label{fig:appendix:e:policy_code_prompt}
\end{figure*}

\section{Scientific Artifacts}\label{app:licenses}
\noindent\textbf{The License for artifacts.}
We use all datasets, frameworks, and tools in accordance with their respective licenses, including AutoGen~\cite{mas:mas_autogen}, the GAIA~\cite{benchmark:gaia} and GPQA-Diamond~\cite{benchmark:gpqa} benchmarks, AIOpsLab~\cite{cloud:aiopslab}, and Chaos Mesh~\cite{chaos:mesh}. We do not redistribute any third-party artifacts, and our source code and the benchmark we construct will be released under an open-source license for reproducibility.

\noindent\textbf{Artifact use consistency with intended use.}
We use all artifacts in line with their intended purpose, employing the benchmarks for evaluating multi-agent task performance and resilience, and the chaos-engineering tools for injecting controlled system uncertainty. Our newly constructed software development benchmark, extended from FSD-Bench~\cite{mas:sw:rtadev} and rSDE-Bench~\cite{benchmark:rsde}, is intended solely for research.

\end{document}